%% file: main.tex
\documentclass[sigconf,nonacm]{aamas} 

\usepackage{subcaption}
\usepackage{caption}
\usepackage{adjustbox}
\usepackage{color}
\usepackage{soul}

\usepackage{float}

\usepackage{balance} 

\setcopyright{ifaamas}
\acmConference[AAMAS '22]{Proc.\@ of the 21st International Conference
on Autonomous Agents and Multiagent Systems (AAMAS 2022)}{May 9--13, 2022}
{Auckland, New Zealand}{P.~Faliszewski, V.~Mascardi, C.~Pelachaud,
M.E.~Taylor (eds.)}
\copyrightyear{2022}
\acmYear{2022}
\acmDOI{}
\acmPrice{}
\acmISBN{}

\acmSubmissionID{109}

\title{Pre-trained Language Models as Prior Knowledge for Playing Text-based Games}

\author{
  Ishika Singh, Gargi Singh, Ashutosh Modi\\
  Department of Computer Science and Engineering\\
  Indian Institute of Technology Kanpur (IITK), India \\
  \texttt{\{ishikas,sgargi\}@iitk.ac.in},\  \texttt{ashutoshm@cse.iitk.ac.in}
}

\begin{abstract}
Recently, text world games have been proposed to enable artificial agents to understand and reason about real-world scenarios. These text-based games are challenging for artificial agents, as it requires an understanding of and interaction using natural language in a partially observable environment. Agents observe the environment via textual descriptions designed to be challenging enough for even human players. Past approaches have not paid enough attention to the language understanding capability of the proposed agents. Typically, these approaches train from scratch, an agent that learns both textual representations and the gameplay online during training using a temporal loss function. Given the sample-inefficiency of RL approaches, it is inefficient to learn rich enough textual representations to be able to understand and reason using the textual observation in such a complicated game environment setting.  In this paper, we improve the semantic understanding of the agent by proposing a simple \textit{RL with LM} framework where we use transformer-based language models with Deep RL models. We perform a detailed study of our framework to demonstrate how our model outperforms all existing agents on the popular game, \texttt{Zork1}, to achieve a score of 44.7, which is 1.6 higher than the state-of-the-art model. Overall, our proposed approach outperforms 4 games out of the 14 text-based games, while performing comparable to the state-of-the-art models on the remaining games. 
\end{abstract}

\newcommand{\BibTeX}{\rm B\kern-.05em{\sc i\kern-.025em b}\kern-.08em\TeX}

\begin{document}


\pagestyle{fancy}
\fancyhead{}


\maketitle 
\section{Introduction}

 Artificial autonomous agents suffer from a number of challenges during training, such as reward, goal, or task under-specification(s). Most approaches design a specific reward function for a given environment and task, which is not generalizable to any other setting. The reward specification gets more complicated with complex environments and tasks. Most agents are trained to do specific tasks or achieve specific goals, as there are not many efficient ways for specifying multiple or complex goals for a given agent. In this scenario, being able to utilize language as an interface between a user and an artificial agent simplifies a number of these challenges. A setting where an agent understands and communicates using natural language is more efficient for rewarding the agent online or via a text description, as opposed to using an expert-designed reward function. Moreover, it is easier to specify complicated tasks or instructions for the agent via the natural language.
 
\begin{figure*}
        \centering
        \includegraphics[width=0.85\textwidth]{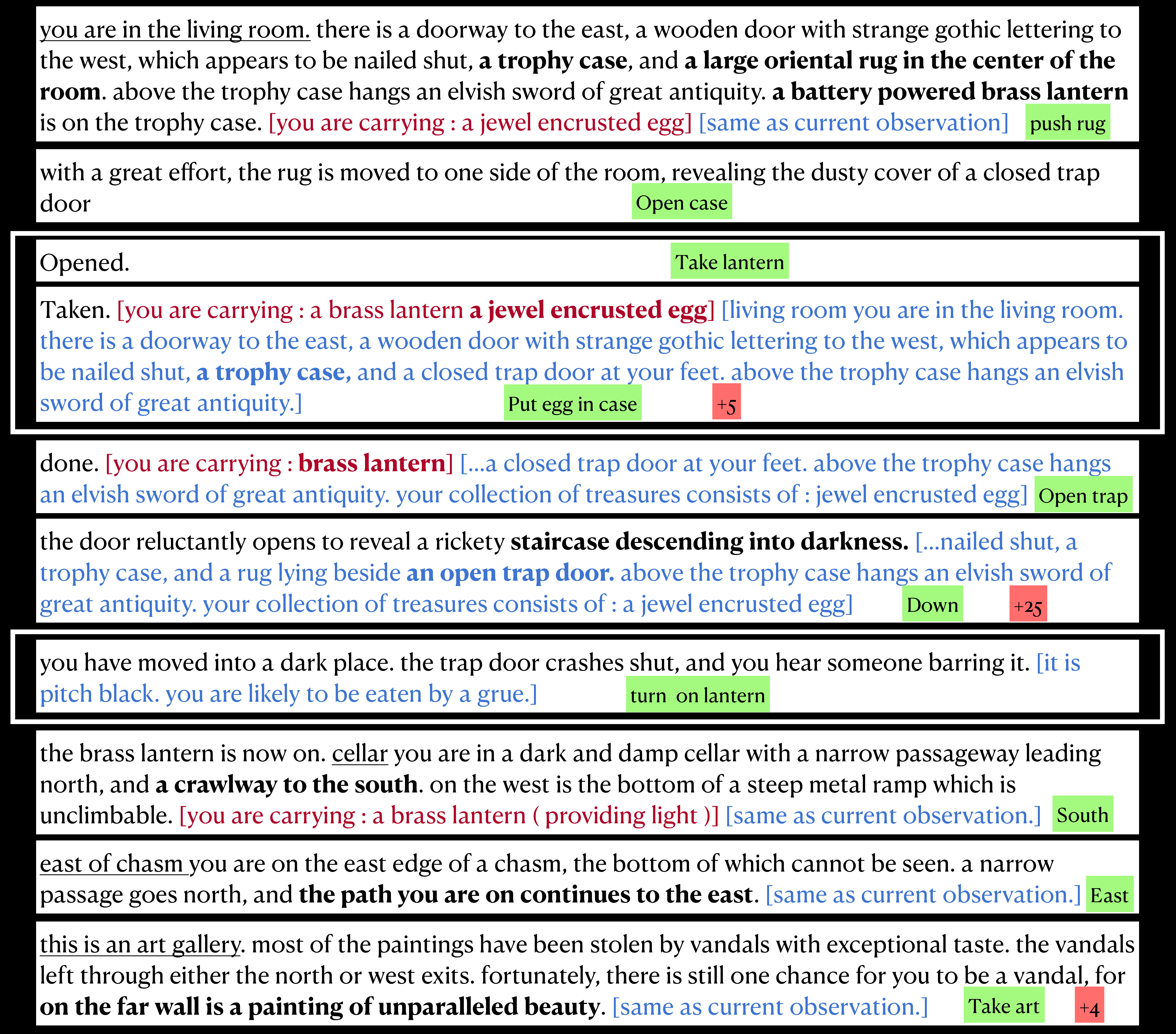}
        \caption{A sample gameplay by our model (DBERT-DRRN) for the classic text game, \texttt{Zork1}. The aim is to solve puzzles and collect 19 treasures in the trophy case, but the agent is not aware of the goal and learns from rewards. Each white box is a state of the game at a step; the green and orange boxes are the action taken, and the reward received correspondingly. The state contains location if visible (underlined), current observation (black), inventory with the player (red), and the current location description (blue). In this example, location, inventory, and description are only mentioned when it changes, while the complete state information is provided to the agent during gameplay. This example shows how our model is able to perform better by learning to use the egg and the lantern in the correct way, as highlighted. 
        }
        \label{fig:example}
    \end{figure*}

Text-based Interactive Fiction (IF) \cite{Hausknecht-2020} games provide such an environment where the agent learns to consume and produce natural language-based inputs and outputs. IF games require an artificial agent to learn policies and operate in a real-world environment created using natural language. These games are designed such that a player receives a textual observation consisting of information about the environment such as a description of the surroundings, objects available for interaction, and states of objects. Based on this information, the player takes an action such as interacting with one of the objects, moving around. etc. A short example of a game-play is shown in Figure  \ref{fig:example}. The textual setting of the games in terms of the observations given to and the actions taken by the agent requires solid language understanding and reasoning. Such a setting has been promoted to develop essential skills required by intelligent agents for various real-life use-cases, including dialogue systems and personal digital assistants. IF games, such as \texttt{Zork1}, have been created to be challenging for human players and consist of texts with dramatic and artful narratives. The game engine, in the case of IF games, generates human-level sophistication and diversity in textual descriptions (as can be seen in the example in Figure \ref{fig:example}) in contrast to games using template-generated synthetic texts. Consequently, such games provide an ideal test-bed for artificial agents that interact with the real-world environment using natural language. Given the setting, agents trained using IF games can be deployed in real-world user-centered robotics applications. 
    
IF games present several challenges for artificial agents as these cover the real world settings.    
It requires an agent to understand the textual state description, handle combinatorial textual action space, and learn a policy in a partially observable environment to maximize the game score. The key challenge is to decipher the long textual observations, extract reward cues from them, and generate a semantically rich representation such that the policy learned on top of it is well informed. Most of the existing works learn textual representations from scratch during the RL training \citep{Hausknecht-2020, Ammanabrolu2020Graph, xu2020deep, ammanabrolu2021how, yao-etal-2020-keep, yao2021blindfolded}, except \citep{guo-etal-2020-interactive} which uses pre-trained GloVe embeddings \citep{pennington-etal-2014-glove}. Online reinforcement learning is known to be sample-inefficient \citep{sutton2018reinforcement}. The reward-based temporal difference objective used for training is proposed for learning the gameplay, and it does not necessarily reinforce the agent to learn the semantics of the game. Therefore, learning textual representations solely from the game-generated text during online learning does not provide rich enough representation to be able to handle such complicated decision-making tasks.  In addition, these games also need some prior knowledge about the functioning of the world to be able to make correct decisions efficiently. For example, as shown in Figure \ref{fig:example}, if it is dark and the agent has a lantern, it should turn it on.
    
Given the requirements for IF games, we propose a pre-trained transformer-based \cite{vaswani2017attention} language model (LM) as a candidate for equipping the RL agent with both language understanding capabilities and real-world knowledge. Transformer-based LMs such as BERT \cite{devlin2019bert}, ALBERT \cite{lan2020albert}, RoBERTa \cite{liu2019roberta} have produced state-of-the-art performance on a plethora of natural language understanding (NLU) tasks. The routine for using the transformer models includes pre-training on large generic English corpora, followed by fine-tuning on a specific downstream task-specific corpus. Moreover, \citet{petroni-etal-2019-language} have shown that pre-trained LMs can act as knowledge bases containing relational and factual knowledge, comparable in utility with other NLP methods having access to oracle knowledge. 
    
In this work, we propose a simple approach that performs better than previously proposed complex approaches. We deploy a pre-trained LM with the existing game agents, namely Deep Reinforcement Relevance Network (DRRN) \citep{he-etal-2016-deep-reinforcement} and Template-Deep Q-Network (TDQN) \citep{Hausknecht-2020}, used in the recently proposed \textit{Jericho} IF game-suite \citep{Hausknecht-2020}. To the best of our knowledge, we are the first to utilize pre-trained LMs for IF games. We use \texttt{DistilBERT}\cite{sanh2020distilbert} (hereafter referred to as \textbf{DBERT}) as the LM owing to its compact size, thereby computing rich text encoding efficiently. We fine-tune DBERT on an independent set of human gameplay transcripts to add a game sense to the pre-trained LM. This set of transcripts contain games other than those in the evaluation game-suite to test the generalizability of the proposed approach. We test both DBERT-DRRN and DBERT-TDQN setup on \texttt{Zork1}, and additionally, we also evaluate DBERT-DRRN on a set of other games from Jericho to test the generalization capability of the model. 
We achieve new state-of-the-art results on 4 out of 14 games while getting comparable performance on others relative to the past approaches (section \ref{sec:related-work}) utilizing sophisticated attention models and knowledge graphs, indicating that pre-trained LMs could be employed as a key component in text-based reinforcement learning models. We release the implementation code via GitHub\footnote{\url{https://github.com/Exploration-Lab/IFG-Pretrained-LM}}. The transcripts (generated when the agent is playing the game) is included in the appendix. 

\section{Related Work} \label{sec:related-work} 
 
Recently, there have been attempts to model real world settings via text based environments. The TextWorld environment \citep{cote18textworld} proposes procedurally generated interactive fiction games, with control over difficulty, and description language. 
Recently, \citet{Hausknecht-2020} proposed a learning environment (Jericho) that supports a set of 32 human-written IF games. These games are written to be challenging for human players, thereby providing a more realistic test-bed for training intelligent agents. \cite{Hausknecht-2020} also present performance results for choice-based (DRRN) and parser-based (TDQN) agents (see section \ref{agents} for details), where the former performed better on average across most of the games. Jericho also provides certain handicaps such as action templates, determining valid actions by detecting world state change, commands to check current items with the agent (\textit{inventory}) or in the current state (\textit{look}). We propose models for solving the games in the Jericho  environment. 

Many of the recent approaches \cite{Ammanabrolu2020Graph, xu2020deep, ammanabrolu2021how} use a dynamically updated Knowledge Graph (KG) to represent the current state of the universe (game). KG-A2C \citep{Ammanabrolu2020Graph} is the first such proposal.  One of the SOTA methods by \citet{xu2020deep}, in addition to KG-A2C architecture, reasons on the KG using low and high-level attention on sub-components of the graph and on the complete graph, respectively, to compute a representation of the game state. Next, they select the actions via recurrent decoding using GRUs, conditioned on the computed game state representation. Similar to KG-A2C approach,  
they train the agent via the Advantage Actor Critic (A2C) method with a supervised auxiliary task of ``valid action prediction” using the action templates handicap.  
Q$^*$BERT \citep{ammanabrolu2021how} presents an open domain QA method to update the KG with more information and an additional intrinsic motivation reward to enable structured exploration. Our agent outperforms all the above agents without an explicit knowledge graph or any additional reward-based supervision or sophisticated reasoning. We show that better language understanding capabilities and appropriate utilization of the world knowledge in game-plays are comparable (and sometimes better) than using explicit KGs.

Multi-Paragraph Reading Comprehension Deep Q-Network \cite{guo-etal-2020-interactive}  (MPRC-DQN) breaks down the problem into two challenges: the partial observability of the environment and large  natural language action space. They solve the partial observability by object-centric retrieval of relevant past observations, and an attention mechanism is deployed to focus on the most relevant context. Template actions from Jericho 
are filled up in question answering (QA) format to generate candidate actions. In contrast, CALM \cite{yao-etal-2020-keep} generates the next set of possible action by fine-tuning a GPT-2 model \cite{radford2019language}. These actions, which are essentially a subset of valid actions, are then fed to a DRRN \cite{he-etal-2016-deep-reinforcement} agent to compute the Q-values. CALM asserts that the action generation model provides linguistic priors to the RL agent. However, since they do not send any feedback to the action generation model and the model only replaces the valid action set by its subset, the RL agent may not necessarily benefit from the linguistic priors. On the other hand, we extract representations from the pre-trained LM and feed them to the DRRN agent for incorporating linguistic priors. With a better observation understanding augmented with the world knowledge via the pre-trained LM, our model also solves the challenge of partial observability to some extent. 
    
\citet{yao2021blindfolded} investigate to what extent semantic information is utilized by the DRRN agent. They show that even in the complete absence of text, the ``DRRN + valid action handicap" setup is able to achieve significant scores, indicating the underlying overfitting to the reward system and memorization tendency of DRRN. They use an inverse-dynamics loss function to regularize the DRRN representation space for improving the semantics.  

Transformers \cite{vaswani2017attention} are a new class of feed-forward neural networks architectures that use a self-attention mechanism to effectively learn short/long-term relationships between tokens in the text. Transformers have been shown to have SOTA performance on almost all the NLP tasks \cite{{wang2018glue, singh2021nlp}}. Transformers are trained using language modeling objective and its variants (e.g., masked language modeling); based on the training objective a number of variants of transformer architectures have been proposed e.g., BERT \cite{devlin2019bert}, DistilBERT \cite{sanh2020distilbert},  ALBERT \cite{lan2020albert}, RoBERTa \cite{liu2019roberta}.   
There have been several transformer-based approaches for both text-based games and RL in general \citep{9231622, pmlr-v119-parisotto20a, chen2021decisiontransformer, janner2021reinforcement}, but these approaches train a transformer from scratch, which is inefficient given that transformers are data-intensive and online RL is sample-inefficient. \citet{9231622} and \citet{pmlr-v119-parisotto20a} modify the transformer structure to add gates and make it light-weight for online training, while \citet{chen2021decisiontransformer} and \citet{janner2021reinforcement} use a smaller transformer but only for offline learning. Using a pre-trained LM brings an additional set of linguistic priors and world knowledge to the RL agent, and in this paper, we propose a way to do so. 
We chose DistilBERT \cite{sanh2020distilbert} for our framework. DistilBERT is a lighter and faster version of BERT trained by knowledge-distilling BERT base with masked language modeling on Toronto Book Corpus and Wikipedia, thus providing us a computationally efficient solution to integrate in the RL setup. 


\begin{figure*}[ht]
\begin{subfigure}{.45\textwidth}
  \centering
  \includegraphics[width=0.9\linewidth]{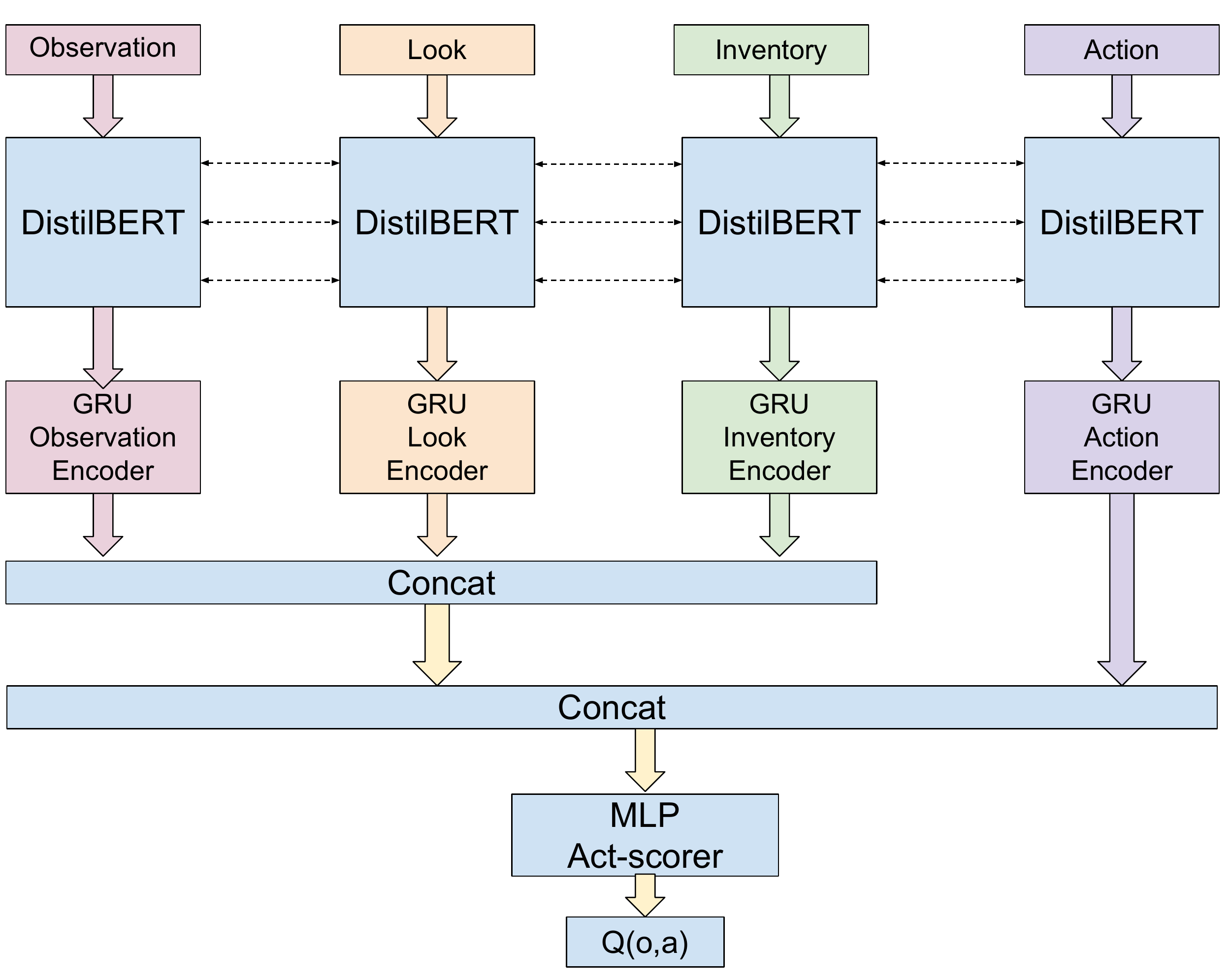}  
  \caption{}
  \label{fig-drrn}
\end{subfigure}
\begin{subfigure}{.45\textwidth}
  \centering
  \includegraphics[width=0.9\linewidth]{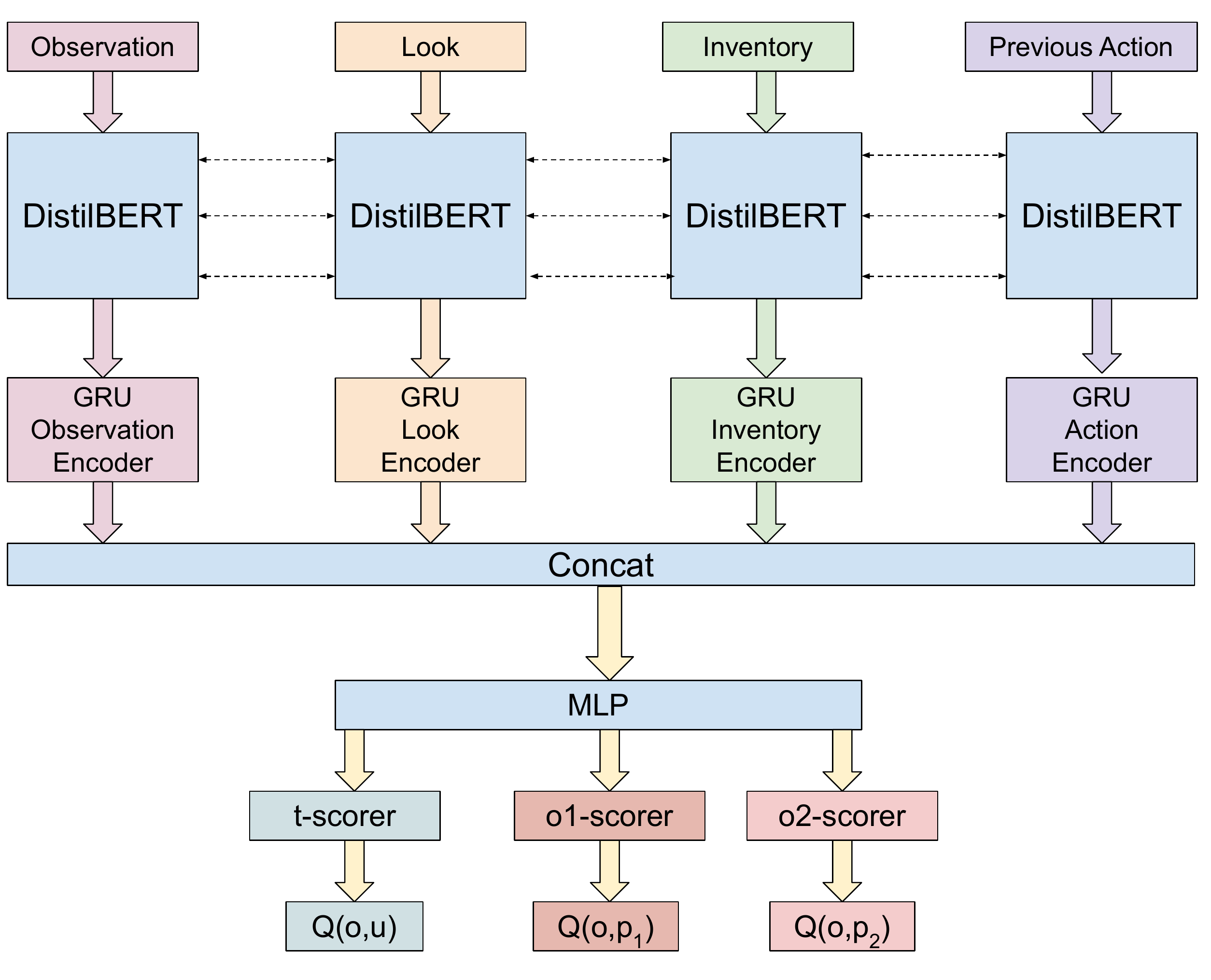}  
  \caption{}
  \label{fig-tdqn}
\end{subfigure}
\caption{RL agents - (a) DRRN: Q-value Q(o,a) is computed for observation o and action a, (b) TDQN: Q-values Q(o,u), Q(o,$p_1$), Q(o,$p_2$) for all
templates $u \in T$ and all vocabulary $p \in V$. 
}
\label{model}
\end{figure*} 
        
\section{Method} 
\subsection{Problem Statement} 
A text-based game can be formulated as a partially observable Markov Decision Process (POMDP) defined by $(S, T, A, O, R)$. An agent interacts with the game environment to receive textual observations $o_t \in O$ at the interaction step $t$. The latent states $s_t \in S$ correspond to $o_t$ along with player, item locations, and inventory contents.  The agent interacts with the environment by executing an action $a_t \in A$ that changes the game state according to a mostly-deterministic but latent transition function $T (s' |s, a)$, and the agent receives rewards $r_t$ from an unknown reward function $R(s, a)$ defined by the game designers. The objective of an RL agent is to take a series of actions that maximize expected cumulative discounted rewards $\mathbf{E}\left[\sum_{i=t}^{\infty} \gamma r_{i}\right]$ at any time $t$. Value-based RL models estimate this reward by learning a Q-function, $Q(o_t, a_t)$ i.e., the expected return when taking an action $a_{t}$, given the observation $o_{t}$. 
The problem statement presents several challenges for the agent: understanding long sophisticated textual state descriptions, extracting reward cues, handling combinatorial textual action space, reasoning-based decision-making, and learning a value function in a partially observable environment to maximize the game score. 

\subsection{RL Agents}\label{agents} 
We build our model on top of existing RL agents. We select the following two agents as these are base architectures \citep{Hausknecht-2020}  on which all other subsequent works build on.

\noindent\textbf{DRRN:} Deep Reinforcement Relevance Network (DRRN) \citep{Hausknecht-2020} uses four GRU encoders to encode the state and each of the actions from the valid action handicap, concatenate the representations, and pass it through an MLP layer to estimate the Q-values ($Q( o_t, a_{ti} )|_{i=1\dots N}$), where $a_{ti}\in \mathcal{A}$ (valid action set). The next action is chosen by softmax sampling the predicted Q-values. The DRRN is trained using tuples $(o_t, a_t, r_t, o_{t+1})$ sampled
from a prioritized experience replay buffer with the traditional temporal difference (TD) loss:
\begin{align*}
    \mathcal{L}_{\mathrm{TD}}(\phi)=\left(r_t+\gamma \max _{a \in \mathcal{A}} Q_{\phi}\left(o_{t+1}, a\right)-Q_{\phi}(o_t, a_t)\right)^{2}
\end{align*}
where $\phi$ represents the overall DRRN parameters, $r_t$ is the reward at $t$-th time step, and $\gamma$ is the discount factor.

\noindent\textbf{TDQN:} Template Deep Q-Network (TDQN) \cite{Hausknecht-2020} like DRRN uses GRU encoders to encode information that consists of the state and previous action in this case. It performs template-based action generation using three Q-values based on the observation $o$: $Q(o,u)$ for all action templates $u\in T$ 
, $Q(o,p_1)$ and $Q(o,p_2)$ for all vocabulary $p \in V$. $p_1, p_2$ fill-up the blanks in the template to generate  candidate actions. The authors used a supervised binary-cross entropy loss to train the model for valid actions in addition to the TD loss.

\subsection{Language Modeling (LM)}
The textual modality of the IF games comes with additional challenges for an RL agent. 
Since the modality of the games is natural language, the modeling of the text-based observations and states becomes important, and this problem has not been explored in-depth in the RL research community. We employ DBERT \cite{sanh2020distilbert} to address this in our model.  We fine-tune DBERT to model the \textit{language} of games. We take inspiration from one of the language modeling techniques, masked language modeling (MLM), used to pre-train DBERT. Given a pair of observation $o_t$ and action $a_t$ composed of tokens ${\textit{[CLS]},o_{t_1}, o_{t_2},...,o_{t_n},\textit{[SEP]},a_{t_1}, a_{t_2},...,a_{t_m}}$, random tokens are masked by a special token, \textit{[MASK]}. Here $n$ is the length of the observation, and $m$ is the length of the action. The masked tokens are predicted from the vocabulary of the model. For example, given sequence \textit{``[CLS] Yes, there is a diorama at one end of the [MASK], but you find yourself unable to [MASK] on it with that fascinating statue sitting in the middle of the room. Once you figure out how it moves, you'll be able to concentrate on the [MASK] parts of the room. [SEP] remove boots"}, the correct predictions would be ``room", ``concentrate" and ``other" respectively. A softmax activation function is used for masked token prediction. Cross entropy loss function is used to learn the parameters, i.e., 
\begin{equation*}
    L = \text{CrossEntropy} \left( x^*, P(x_{\textit{[MASK]}} | S) \right)
\end{equation*}
where $x^*$ is the ground truth, $x_{\textit{[MASK]}}$ is the predicted token, $S$ is the masked sentence,  and $P(x_{\textit{[MASK]}} | S)$ is the predicted distribution over the vocabulary. This training enables DBERT to tune its prior knowledge in the language of games. 

\subsection{RL with Pre-trained LM} 
 We use the fine-tuned DBERT to model the input for the requisite RL agent. We encode the state and the actions separately using DBERT and feed the representations to the respective GRUs in DRRN and TDQN agent. The agents are then trained as described in \ref{agents} (Figure \ref{fig-drrn} and \ref{fig-tdqn}). In this way, the trained DBERT is able to transfer its prior knowledge about the world to different agents across different games. The agents, in turn, learn more informed policies on the top of DBERT optimized for specific games. We use a single instance of the trained DBERT and do not fine-tune DBERT further for specific games to keep the language model general and adaptable to any RL agent or game.


\begin{table*}[h!]
 \caption{Final (raw) and max scores seen by agents (available via prior work) comparing trained DBERT-DRRN with all the current SOTA baselines across a set of games from Jericho game-suite. For the baselines, we use the results reported in the original papers. The missing scores not in the previous work are  denoted as '-'. \textbf{Max} denotes the maximum possible scores based on human-written optimal walkthroughs for winning the game without the step limit of 100. Note that our model (DBERT-DRRN) is trained for less than half of the training steps used in other baselines. \textbf{Bold} scores denote the best score, \textcolor{blue}{blue}* scores denote the second best. DBERT-DRRN gets an avg. norm of 15.7\% on all games, and 21.1\% on 6 games reported for INV-DY. *the games with rising learning curve for DBERT-DRRN till the last training step. $^p$(possible/easy), $^d$(difficult), $^e$(extreme) refers to game difficulty level as given in \citep{Hausknecht-2020}.
    }
    \label{tab:scores}
    \centering
    \begin{adjustbox}{width=1\textwidth}
    {\renewcommand{\arraystretch}{1.3}%
    \begin{tabular}{c|ccccccc|c|c}
    \hline
       \textbf{Game} & \textbf{TDQN} & \textbf{DRRN} & \textbf{MPRC-DQN} & \textbf{SHA-KG} & \textbf{Q*BERT} & \textbf{CALM} & \textbf{INV-DY} & \textbf{DBERT-DRRN} & \textbf{Max}  \\
        & \citep{Hausknecht-2020} & \citep{Hausknecht-2020} & \citep{guo-etal-2020-interactive} & \citep{xu2020deep} & \citep{ammanabrolu2021how} & \citep{yao-etal-2020-keep} & \citep{yao2021blindfolded} & (ours) &  \\
         & raw  & raw / max & raw & raw  & max  & raw & raw / max  & raw / max  &   \\
       \hline
       \hline
         \texttt{Inhumane}$^{*p}$ &0.7 &0& 0 & 5.4 &-  & \textcolor{blue}{25.7}*  & 19.6 / 45  & \textbf{32.8} / \textbf{50} & 90  \\
        
         \texttt{Jewel}$^{d}$ & 0 & 1.6  & \textcolor{blue}{4.46}* & 1.8 & -  & 0.3  & - & \textbf{6.5} / 13  & 90  \\
        
        \texttt{Library}$^{*p}$ &6.3 &\textcolor{blue}{17.0}* & \textbf{17.7} &15.8  & 19 &9.0   & 16.2 / \textbf{21} & \textcolor{blue}{17.0}* / \textbf{21}  &30  \\
        
        \texttt{Ludicorp}$^d$ &6 &13.8& \textbf{19.7} & \textcolor{blue}{17.8}* &22.8  &10.1   & 13.5 / \textbf{23} & 12.5 / 18  & 150  \\
        
        \texttt{Omniquest}$^p$ &\textbf{16.8} &5& 10.0 & - & - & \textcolor{blue}{6.9}*  &5.3 / \textbf{10}  & 4.9 / 5  &50  \\
        
        \texttt{Reverb}$^{p}$ & 0.3 & \textcolor{blue}{8.2}* & 2.0  & \textbf{10.6} & - & - & - & 6.1 / 12  & 50  \\
        
        \texttt{Snacktime}$^{p}$ & 9.7 & 0 / 0.25 & 0  & - & - & \textcolor{blue}{19.4}* & - & \textbf{20.0} / \textbf{20}  & 50  \\
        
        
        \texttt{Spellbrkr}$^{*e}$ & 18.7 & 37.8  & 25  & \textbf{40} & - & \textbf{40} & - &  \textcolor{blue}{38.2}* / 40  &  600 \\

        \texttt{Spirit}$^e$ & 0.6 & 0.8 & \textbf{3.8}  & \textbf{3.8}  & - & 1.4 & - & \textcolor{blue}{2.1}* / 8  & 250  \\
        
        \texttt{Temple}$^p$ & \textcolor{blue}{7.9}* & 7.4  & \textbf{8.0}  & \textcolor{blue}{7.9}*  & \textbf{8.0}  & 0 & - &  \textbf{8.0} / 8.0  & 35  \\
        
        \texttt{Tryst205}$^{*e}$ & 0 & \textcolor{blue}{9.6}*  & \textbf{10.0} & 6.9  & - & - &  & 9.3 / 17  & 350 \\
        
        \texttt{Zork1}$^d$ &9.9 &32.6 / 53 & 38.3 &34.5  & 41.6 & 30.4  & \textcolor{blue}{43.1}* / \textbf{87} & \textbf{44.7} / 55   & 350  \\
        
        \texttt{Zork3}$^d$ &0 &0.5& \textbf{3.63} &\textcolor{blue}{0.7}*  &-  &0.5   & 0.4 / \textbf{4}  & 0.2 / \textbf{4} & 7 \\
        
        \texttt{Yomomma}$^{*d}$ & 0 & 0.4  & \textbf{1.0} & -  & - & - & - &  \textcolor{blue}{0.5}* / 1.0  &  35 \\
        \hline \hline
        Avg. Norm (\%) & 7.7 & 10.2 & 14.2 & 13.3 &  - & 12.8  & 18.9 &\textbf{15.7} (\textbf{21.1}) & 100 \\
        \hline
    \end{tabular}}
    \end{adjustbox}
   
\end{table*}

\section{Experiments} 
We evaluate how the addition of a pre-trained LM facilitates different agents and games. We also present ablation studies to demonstrate the need for training the pre-trained LM. Lastly, we present a study on how our model is able to achieve new state-of-the-art results on \texttt{Zork1}. 
\subsection{LM training} 
 We use the ClubFloyd dataset \cite{yao-etal-2020-keep} for fine-tuning DBERT. It is a collection of human game-play trajectories on 590 games. These are not the optimal trajectories, but it does impart a general game-play sense. We pre-process this data to obtain around 217K pairs of observation and action, ($o_t,a_t$). The transcripts are from games not included in our evaluation (section \ref{sec-rl-expt}) to keep our approach general. We use a 768-dimensional vanilla DBERT (base cased) model pre-trained on English Wikipedia and Toronto Book Corpus. We train this model for 2 epochs on pre-processed ClubFloyd dataset using the MLM technique. 
         
\subsection{RL Training} \label{sec-rl-expt} 
\textbf{Jericho Environment: } We evaluate our agent on 14 games with different difficulty levels\footnote{\noindent We choose games across varying difficulty level (possible/easy, difficult, extreme) as indicated in Table \ref{tab:scores}. The difficulty level assigned to each game \cite{Hausknecht-2020} is based on optimal solution length, reward density, puzzle complexity, etc. Please refer to \citep{Hausknecht-2020} for more details on difficulty level.} 
using the Jericho framework, and use the same Jericho handicaps as in the respective DRRN and TDQN baselines \citep{Hausknecht-2020}. The states are observations ($o_t$) concatenated with items in possession of the player and its current location description provided by the game engine using commands \textit{inventory} and \textit{look}. We also use valid actions and action templates for DBERT-DRRN and DBERT-TDQN respectively, following the baselines. A single game episode runs for 100 environment steps at max, or gets terminated before the game is over or won. 

\noindent\textbf{Training Setup:}
We test our framework on two RL models: DRRN (Figure \ref{fig-drrn}) and TDQN (Figure \ref{fig-tdqn}). The GRUs used in the agents have an embedding dimension of 768 features and a hidden dimension of 128 features. Similar to \citep{yao-etal-2020-keep, Hausknecht-2020} we collect interaction data for DBERT-DRRN and DBERT-TDQN on 8 parallel game environment instances and use a prioritized experience replay buffer to store trajectories with best scores. We sample transitions from this priority buffer with a priority fraction of 0.5, while taking the remaining training data from a general replay buffer. Most importantly, we train our agents for a maximum of 5$\times 10^4$ steps, which is less than half of the training steps used in most other baselines, and achieve SOTA and comparable scores on different games. Each training step makes an interaction with the environment in all 8 instances, therefore we only use half the amount of interaction data for training in comparison to the baselines. We use a softmax exploration policy and a learning rate of $10^{-4}$, along with other presets the same as the baseline. We report the average of scores on the last 100 finished episodes as the score on a game run. We train two independent runs for each model and game, and report the average of their scores as the final (raw) score, along with the average maximum score seen by the runs. The maximum seen scores are a measure of the exploration ability, while the raw scores are that of the learning ability of an agent. We report the learning curves for \texttt{Zork1} in Figure \ref{fig-ablation}.

\section{Results and Analysis} 
We performed a detailed analysis of our method. We present overall game-scores as compared to baseline models across a set of games. We also present an ablation study of our model done on \texttt{Zork1}, along with a semantic understanding analysis. Finally, to actually interpret the agent from inside beyond numbers, we also present a qualitative analysis of our model on \texttt{Zork1} via a case study discussing the factors leading to higher score.

\subsection{Overall Scores}
\begin{figure}[th!]
\begin{subfigure}{.45\textwidth}
  \centering
  \includegraphics[width=\linewidth]{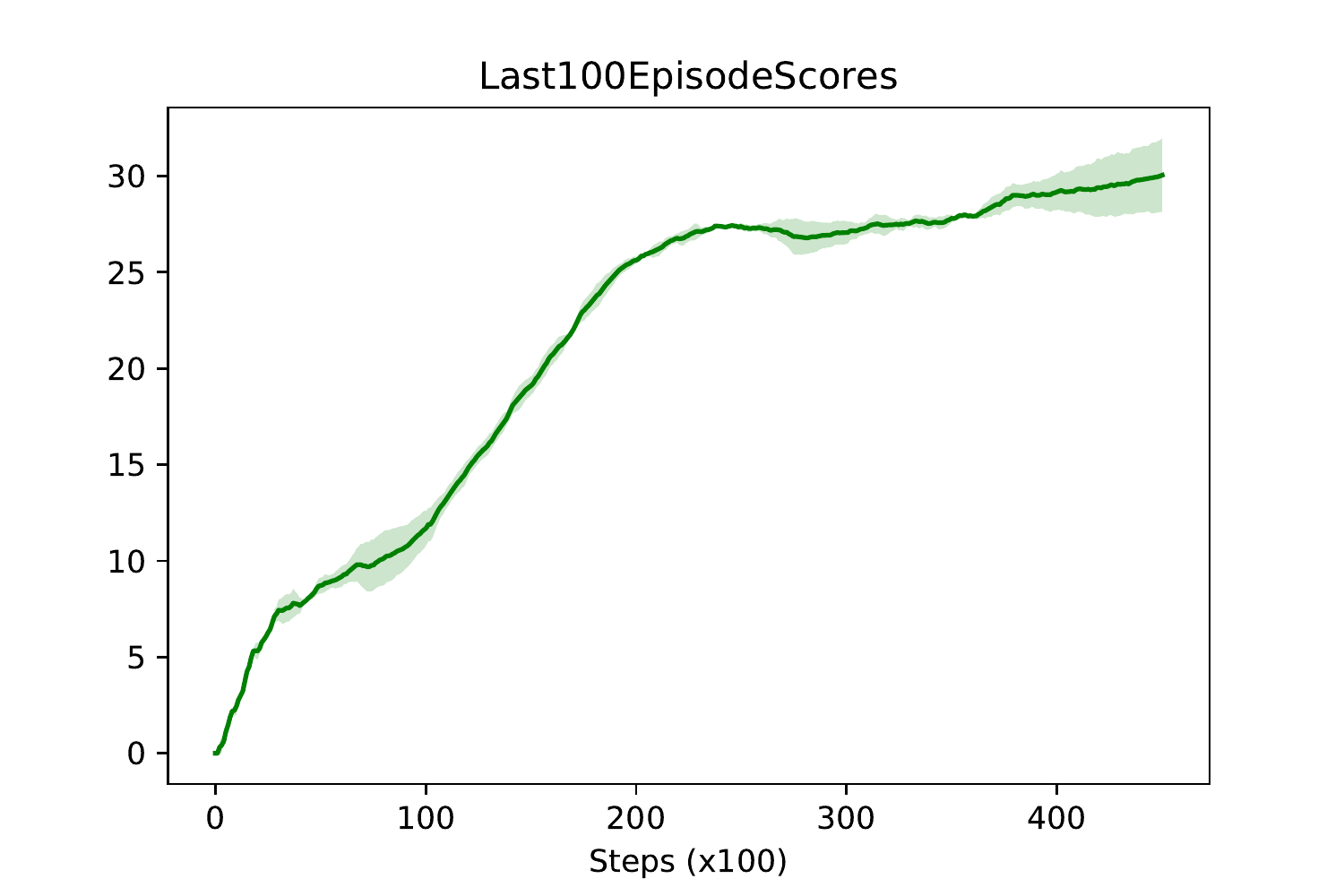}  
  \caption{}
  \label{last100-1}
\end{subfigure}
\begin{subfigure}{.45\textwidth}
  \centering
  \includegraphics[width=\linewidth]{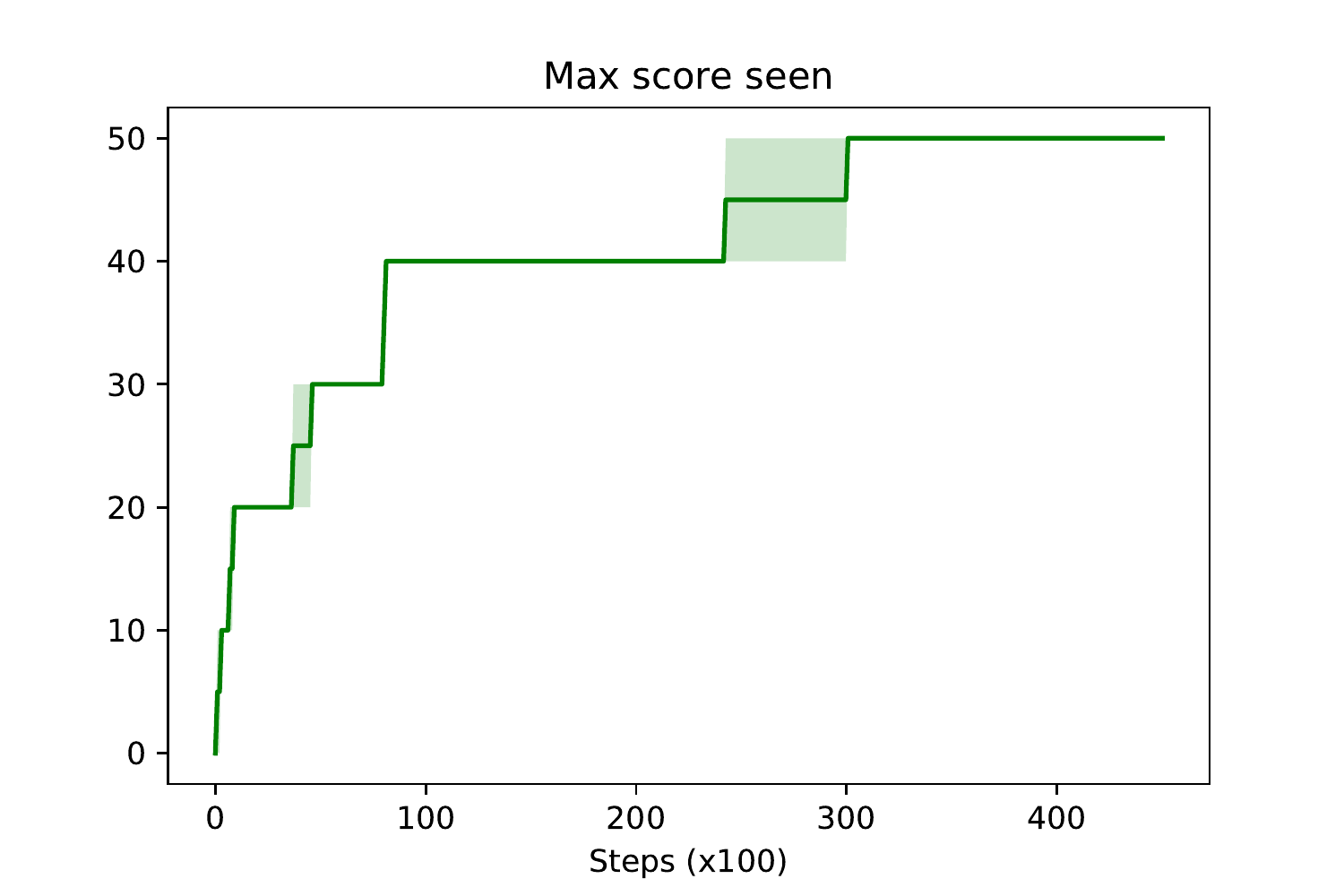}  
  \caption{}
  \label{max-1}
\end{subfigure}
\caption{DBERT-DRRN performance results on \texttt{Inhumane}
}
\label{fig-ablation-1}
\end{figure}

\begin{figure}[th!]
\begin{subfigure}{.45\textwidth}
  \centering
  \includegraphics[width=\linewidth]{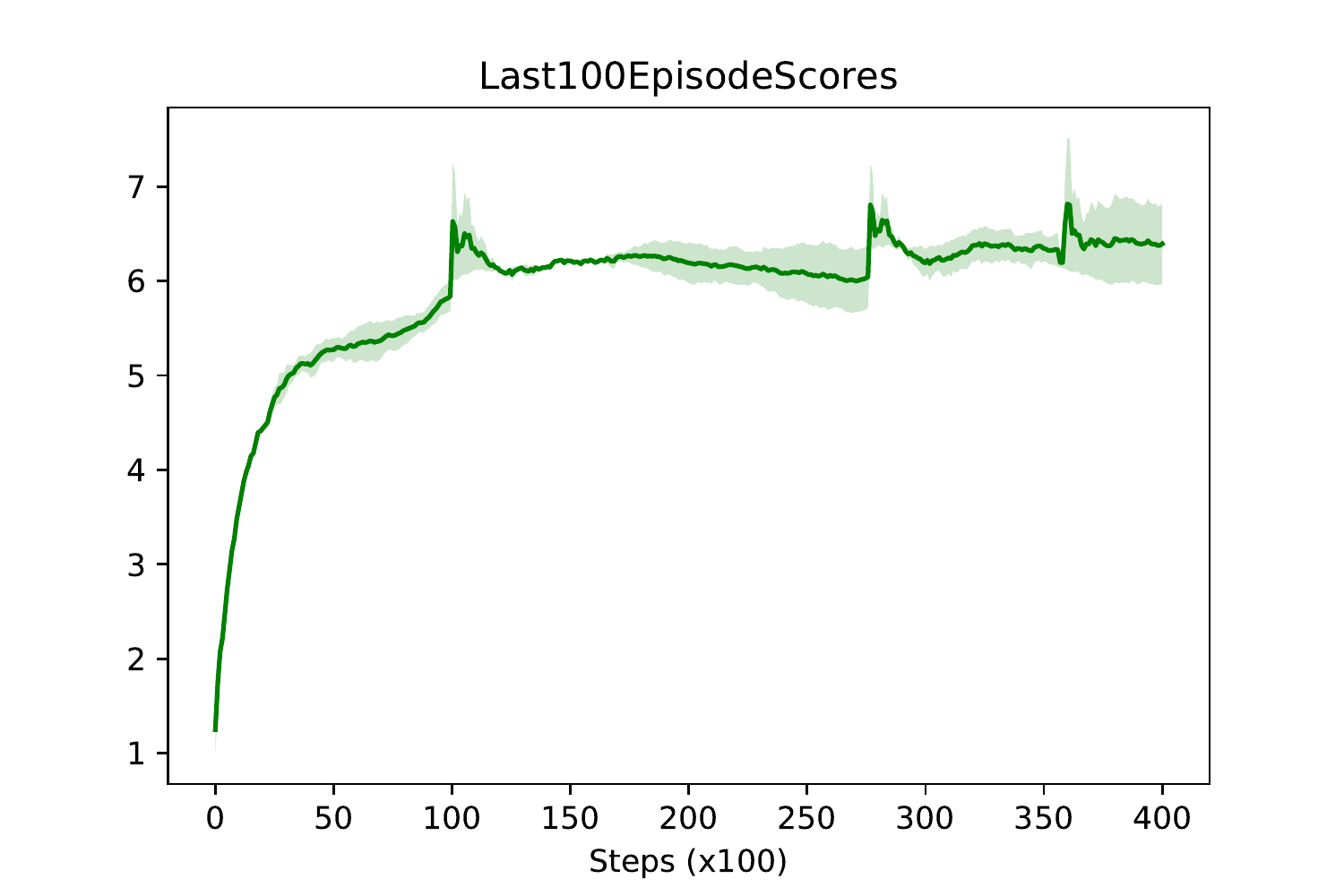}  
  \caption{}
  \label{last100-2}
\end{subfigure}
\begin{subfigure}{.45\textwidth}
  \centering
  \includegraphics[width=\linewidth]{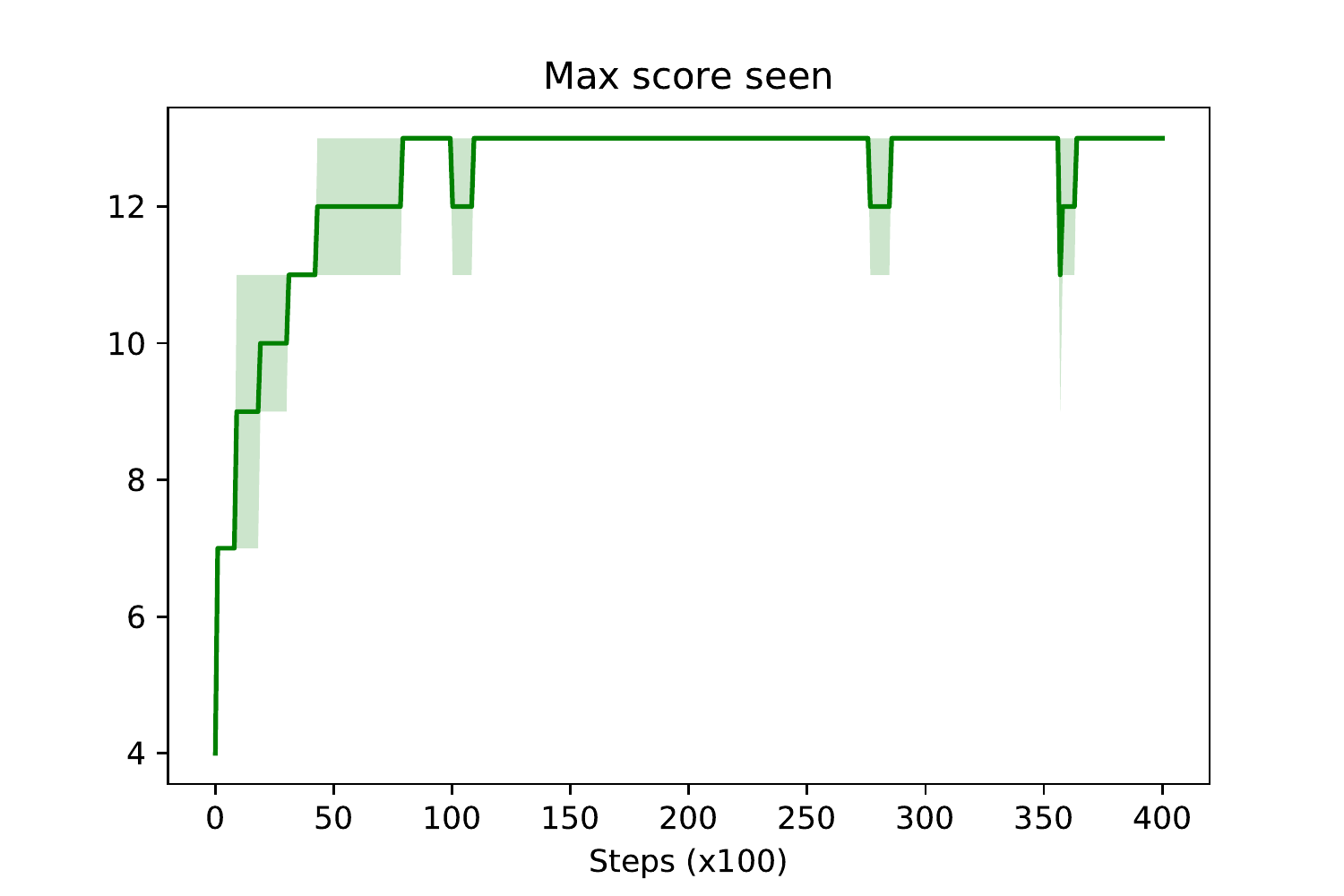}  
  \caption{}
  \label{max-2}
\end{subfigure}
\caption{DBERT-DRRN performance results on \texttt{Jewel}
}
\label{fig-ablation-2}
\end{figure}

Table \ref{tab:scores} reports the final scores of our best performing model (trained DBERT with DRRN) in comparison with 7 existing baselines on 14 games. We report raw or maximum or both the scores as given in original papers. 
Different baselines achieve SOTA scores on different games. 
Our model achieves SOTA results on 4 games: \texttt{Zork1}, \texttt{Inhumane}, \texttt{Snacktime}, and \texttt{Jewel}, while being second best or comparable on most of the other games. We present the learning curves for \texttt{Zork1}, \texttt{Inhumane}, and \texttt{Jewel} in Figures \ref{fig-ablation}, \ref{fig-ablation-1}, and \ref{fig-ablation-2} respectively. INV-DY \cite{yao2021blindfolded} uses additional loss objectives inspired from curiosity-driven exploration \cite{pathak2017curiositydriven}. While it helps them achieve higher maximum scores on \texttt{Zork1}, but are not able to learn the high score trajectories. On the other hand, our agent efficiently learns the max score trajectories explored by it, thereby indicating that with a better exploration strategy our model has the potential to achieve better scores. None of the other agents, with a max score of 55, are able to stably reach a score as high as our model, that maintains a margin of 6.4 from the best model \cite{guo-etal-2020-interactive}. Our agent explores higher max score on  \texttt{Inhumane}, but more importantly, it is able to learn the best-explored trajectories, thereby plateauing closer to the max scores for many games (\texttt{Inhumane, Jewel, Omniquest, Zork1}), indicating that the trained DBERT is an important learning component, and it also facilitates the exploration to some extent. We also report the normalized score (raw score as a factor of max possible score collected from human-written optimal walk-through) averaged across all games. We get an overall norm of 15.7\%, followed by 14.2\% achieved by MPRC-DQN \cite{guo-etal-2020-interactive}. Our model does not suffer a lot on any game, while the second best (MPRC-DQN) gets 0 scores on \texttt{Inhumane} and \texttt{Snacktime}, and notably higher scores on others (\texttt{Ludicorp, Spirit, Zork3}). It indicates both the generalization tendency and the necessity of the pre-trained LM deployed in our model. When compared to the average norm of 18.9\% for INV-DY evaluated on 6 of the 14 games they reported, we get 21.1\%. The average norm is also a measure of human-machine gap for text-based games, indicating that IF games are at best only 15.7\% solved. Hence, it is a good benchmark for developing language understanding agents.
We have included the learning curves for independent runs on \texttt{Zork1} and its game transcripts in the Appendix.

\subsection{DBERT Training Ablation}
\begin{figure*}[th!]
\begin{subfigure}{.45\textwidth}
  \centering
  \includegraphics[width=\linewidth]{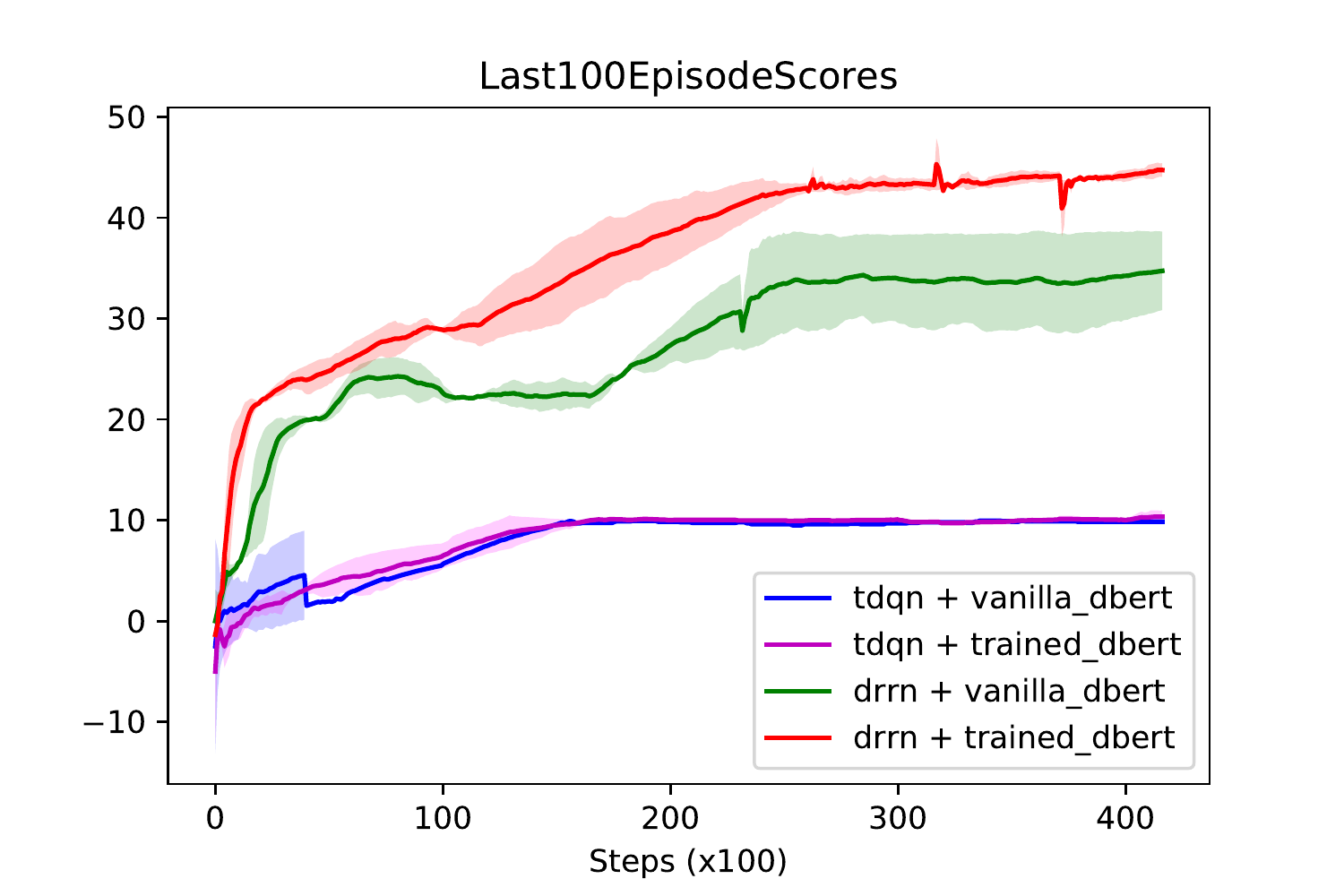}  
  \caption{}
  \label{last100}
\end{subfigure}
\begin{subfigure}{.45\textwidth}
  \centering
  \includegraphics[width=\linewidth]{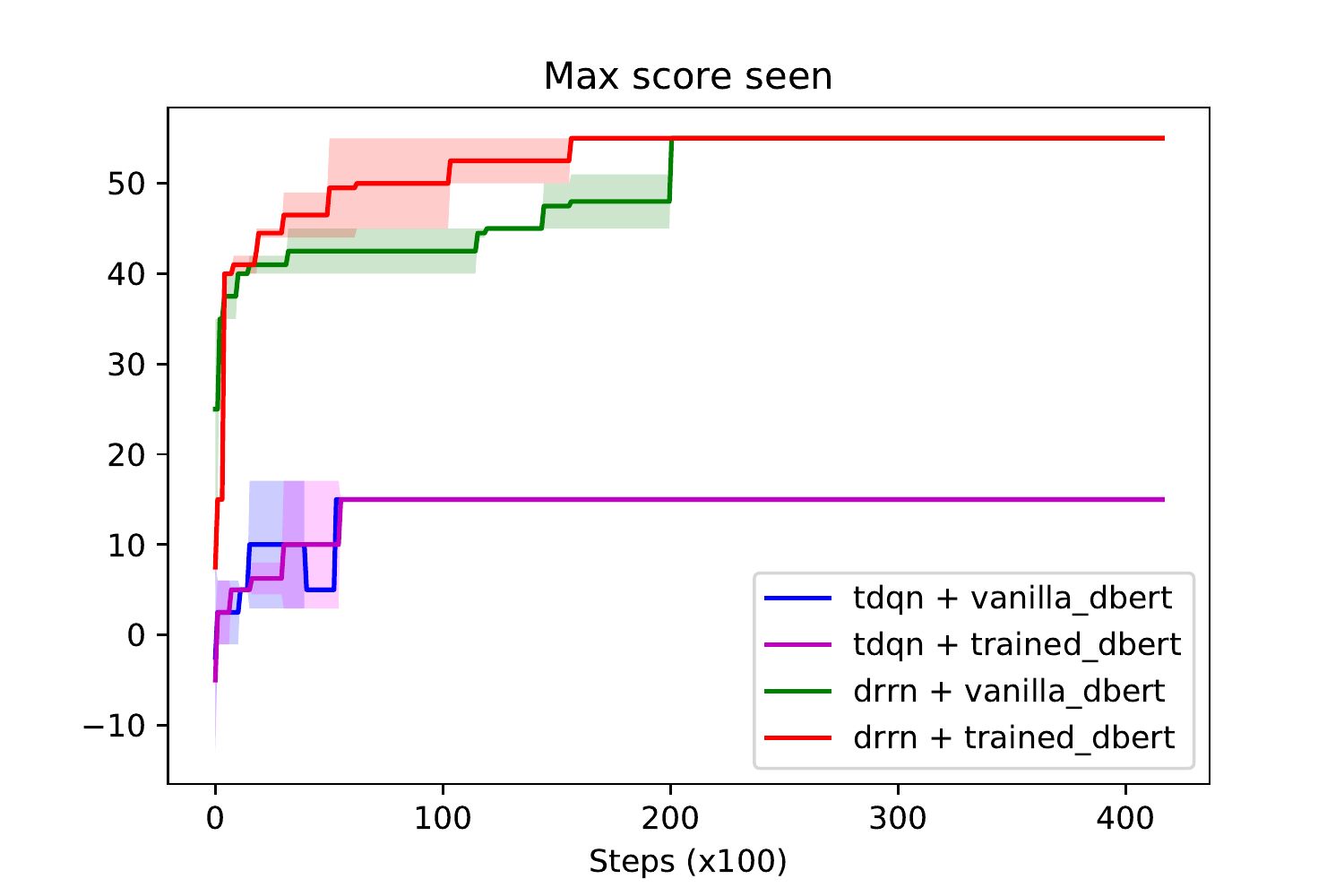}  
  \caption{}
  \label{max}
\end{subfigure}
\caption{DBERT performance ablation results on \texttt{Zork1} 
}
\label{fig-ablation}
\end{figure*}

Our framework includes a DBERT trained on human trajectories from different games. This training induces a language acquaintance with games. We tested empirically how this affected the model. We obtain runs on both DRRN and TDQN based models for the vanilla DBERT and DBERT pretrained on games from ClubFloyd dataset. We present the results in Figure \ref{fig-ablation} where we see DRRN with pretrained DBERT outperforms DRRN with vanilla DBERT with a final score of 44.72 compared to 34.73. The TDQN based models do not exhibit much difference, but pretrained version still has a higher score of 10.35, whereas vanilla DBERT version of the framework obtains 9.85. It indicates the importance of acquainting the pre-trained LM with the game language, so that it can transfer its prior knowledge appropriately to the agents.

\subsection{Semantic Understanding: Transfer}
\citet{yao2021blindfolded} show that ``TD loss + valid action handicap" setup with mostly deterministic rewards can lead to overfitting and memorization, thus, hindering the actual goal of understanding and learning to operate in natural language. Therefore, we also test whether an agent trained for one game can transfer its learning to other similar games to see if the agent is learning the semantics or memorizing the trajectories. We train an agent on \texttt{Zork1} and test it without any training on \texttt{Zork3}, as these are the two most similar games in the Jericho game-suite. The average episode score over 300 episodes was 0.06 for DBERT-DRRN and 0.007 for DRRN. These scores are not as high as achieved by training fresh agents on \texttt{Zork3}; however, DBERT-DRRN is able
to transfer its knowledge better than DRRN.
        

\subsection{Avoiding being eaten by a grue and thereafter}
\begin{figure}[t!]
\centering
\includegraphics[scale=.50]{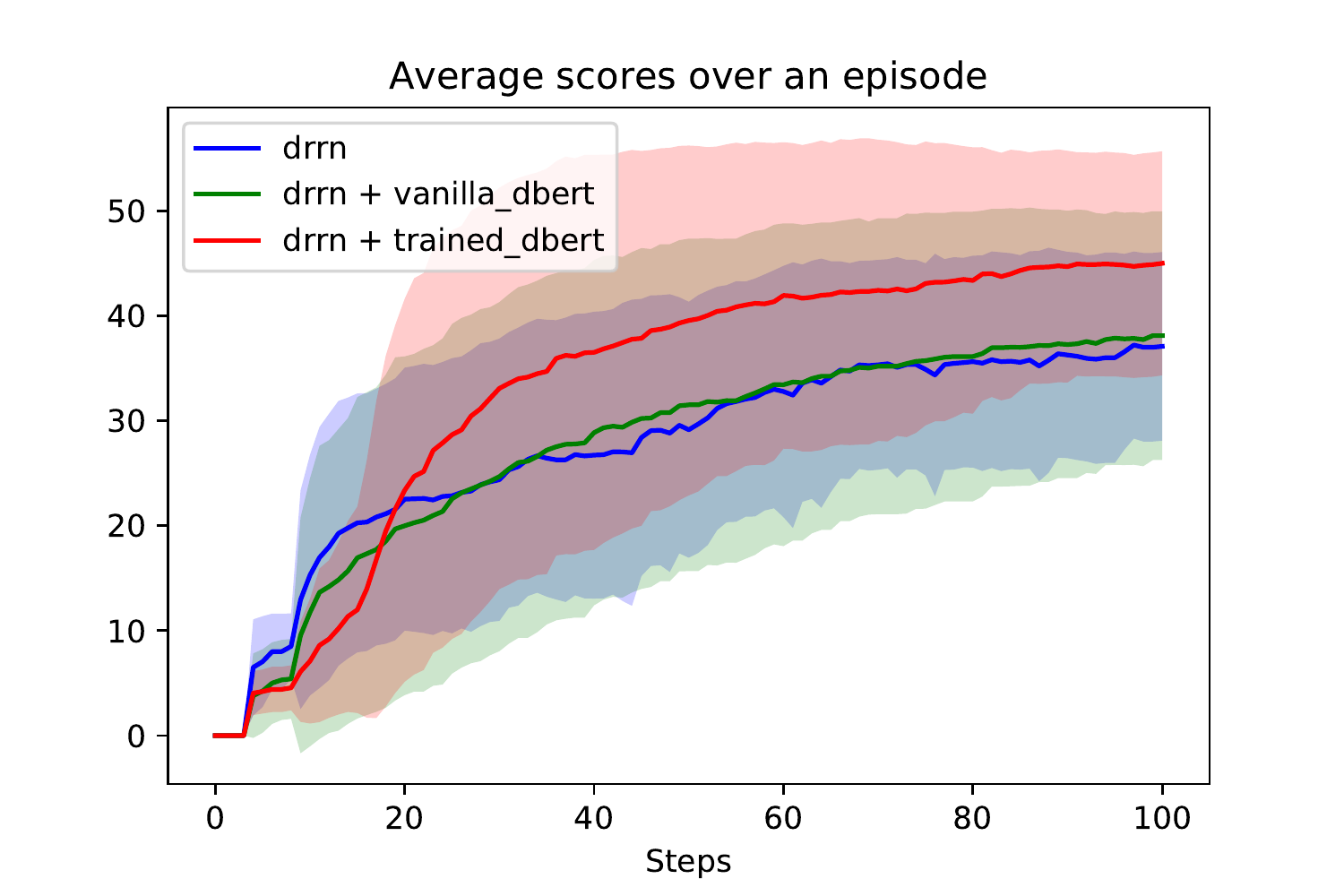}
\caption{Avoiding being eaten by a grue: Average scores over an episode for \texttt{Zork1} 
}
\label{grue}
\end{figure}

We compare scores over an episode for (vanilla and trained) DBERT-DRRN and DRRN. We sample 100 random episodes from the last few hundred episodes during the training, and plot the average scores over an episode in Figure \ref{grue}. An episode is set to terminate after 100 steps or before (if the game is over or won). The goal in \texttt{Zork1} is to collect 19 treasures in the \textit{trophy case}, which agents are not aware of, and they learn via in-game rewards. Existing agents traverse the \textit{Forest} for the \textit{Egg} (1st treasure), followed by \textit{Kitchen}, \textit{Cellar}, and \textit{Gallery} for the \textit{painting} (next treasure), while solving various puzzle and escaping the \textit{grue}.  In DRRN, the agent gets to the \textit{Kitchen} with a reward of +10 at around step-10 and then to the \textit{Cellar}  with a reward of +25 around step-15 as we can see in the plot. It doesn't learn to take and carry the \textit{Egg} for a reward of +5 (and another +5 for putting it in the case) before moving to the \textit{Kitchen} even though the observations present the \textit{Egg} as something precious \textit{``..in the bird's nest is a large egg encrusted with precious jewels, apparently scavenged by a childless songbird. the egg is covered with fine gold inlay,.."}. \citet{ammanabrolu2021how} use an intrinsic motivation reward in addition to the rewards provided by the game engine, in order to get past when stuck in the game, i.e., when the observations do not change. This motivation also helps them provide rewards to add dependencies to their KG for the \textit{Egg}. However, the LM in trained DBERT-DRRN guesses that the \textit{Egg} is important without any extra supervision, and takes it around step-10, hence, achieving higher scores later on. Moreover, trained DBERT-DRRN avoids being eaten by a grue in the \textit{Cellar} which says \textit{``the trap door crashes shut, and you hear someone barring it. it is pitch black. you are likely to be eaten by a grue."} by instantly turning on the \textit{lantern} without any reward, while DRRN takes random actions at this point and dies. Then our agent finds the way to the \textit{gallery}, and gets the \textit{painting} for +4 points. Thus, merely improving the understanding of the agent enables it to take better actions without any additional supervision. A part of a sample episode transcript is shown in Figure \ref{fig:example}, and the complete episode is provided in the Appendix.  
    
In summary, the pre-trained LM enables the DBERT-DRRN agent to extract reward cues from the observations as well as use its priors on world sense and game sense to get past the grue. We observed that the Vanilla-DBERT is only as good as the base DRRN agent. Training it on a set of gameplays is improving the model considerably, indicating the importance of this training, which is essentially channeling the world sense of Vanilla-DBERT into a gameplay mode. The DBERT-DRRN agent understands that a \textit{``jewel encrusted egg appearing extremely fragile"} should be taken, and should not be thrown, or if it's dark the lantern should be turned on. Trained DBERT-DRRN assigns reasonably high Q-values to such actions in the valid action set, while DRRN assigns similar values when rewards are absent for the same (as can be seen in the trajectories provided in the Appendix). Both DBERT-DRRN and all other baselines are able to reach the gallery. Our agent does so more often as it escapes the grue in the Cellar using the lantern. Moreover, it extracts intermediate rewards cues and ends up with a higher score at the end of an episode. The baselines with a score around 39 or less are the ones which could not learn these aspects of the game. After getting the painting, the agents explore several options, but none of them, including ours, are able to find and learn to find the third treasure. Moving past trained DBERT-DRRN score will likely require a more intelligent agent with better exploration and learning strategies.

\section{Conclusion and Discussion} 
This paper proposes using a pre-trained LM fine-tuned on game dynamics for RL agents trained for text-based games. It provides three-fold benefits to the RL agent: linguistic priors, world sense priors, and game sense priors. It facilitates the agent to achieve SOTA results on several text-based games, even though being a simpler approach than all other baselines. The proposed approach indicates the importance of using pre-trained LMs for RL agents in text-based games. Through this work, we want to draw the research community's attention and motivate research in this direction to create even better priors using pre-trained LM for language understanding agents, such as distilling its knowledge to agents with better architectures. Such unsupervised components for exploration and learning will also be more useful in real-life scenarios where there are no explicit determinate rewards than reward-based representation learning. We have shown that training RL agents for text-based games in the absence of a prior is inefficient. The two key reasons are: learning language from scratch using the limited text data with a TD loss or related objective does not produce rich enough representations, and the agent has no source of world knowledge to be able to take appropriate actions given the state. The agents thus trained are not really understanding their state in the game, leading to the suspicion that the agents are simply memorizing the best-explored trajectories \cite{yao2021blindfolded}. 

IF games are still far from solved. We need agents with better exploration strategies (for instance curiosity-driven learning \citep{pathak2017curiositydriven}) to find high score trajectories, as well as learning architectures and objectives that can facilitate learning these trajectories while efficiently leveraging priors from pre-trained language models. Current results show the performance of the agents trained on the same game environments. Another line of future work is to test trained agents on unseen games. This will require a strong understanding of the world functioning and generalized training strategies to be able to acquire skills from multiple games and transfer them to the unseen test environments. This will also be a robust test-bed for semantic understanding of the agent, as memorization will no longer make any points in unseen games. Leveraging commonsense knowledge for text-based game agents can be another interesting direction for improving the knowledge priors. Many actions taken during these games are associated with object affordances. With a commonsense knowledge source, the agent will be able to extract rational interaction possibilities with the objects. 


\bibliographystyle{ACM-Reference-Format} 
\bibliography{references}


\newpage
\appendix
\input{appendix}

\end{document}

%% file: appendix.tex
\section*{Appendix}

%


\section{Learning Curves}
\begin{figure}[H]
\centering
        \begin{subfigure}{0.50\textwidth}
             \includegraphics[width=0.8\linewidth]{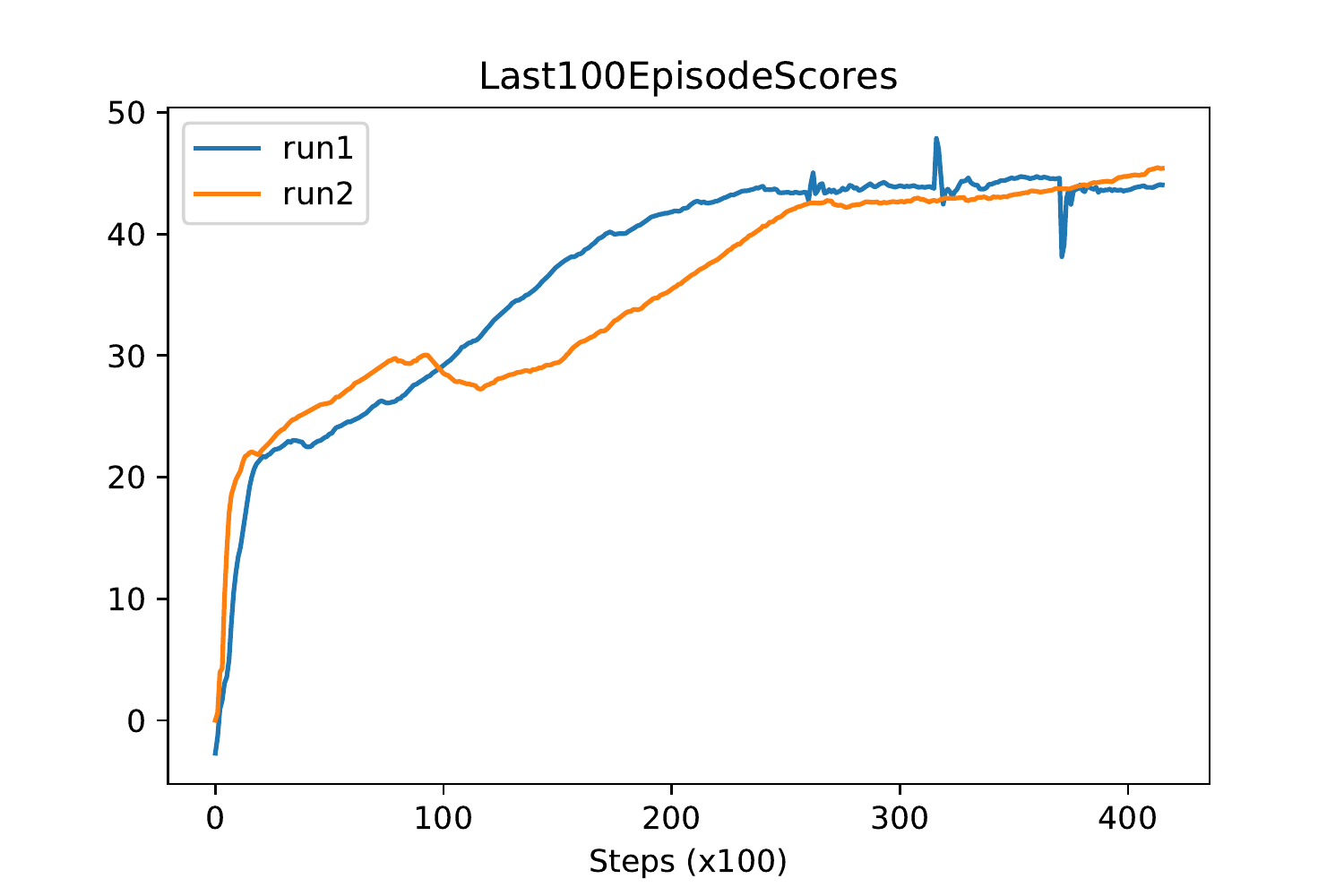}
                \caption{}
                \label{fig:gull}
        \end{subfigure}%
        \begin{subfigure}{0.50\textwidth}
                \includegraphics[width=0.8\linewidth]{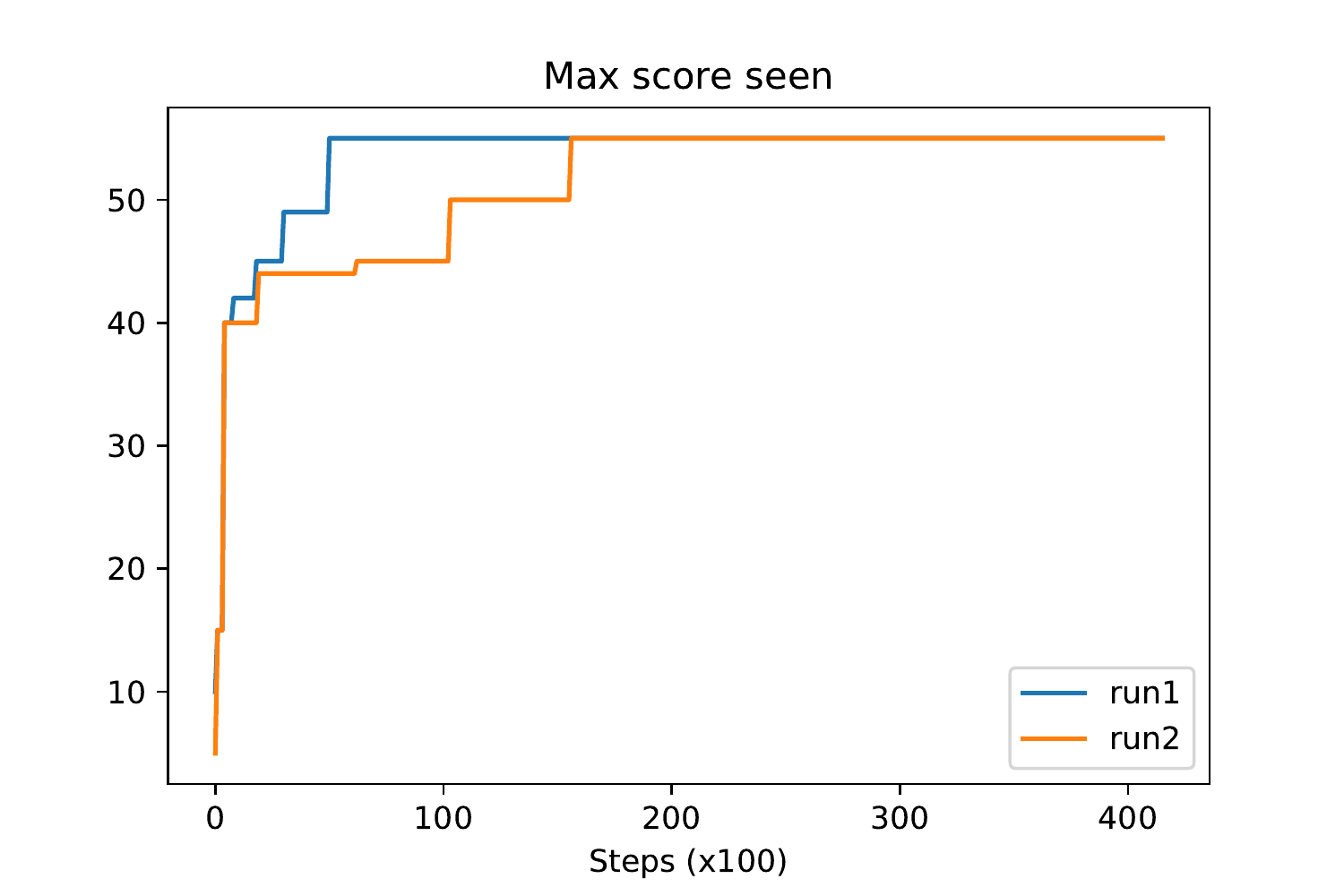}
                \caption{}
                \label{fig:gull2}
        \end{subfigure}%
        
        \begin{subfigure}{0.50\textwidth}
         \includegraphics[width=0.8\linewidth]{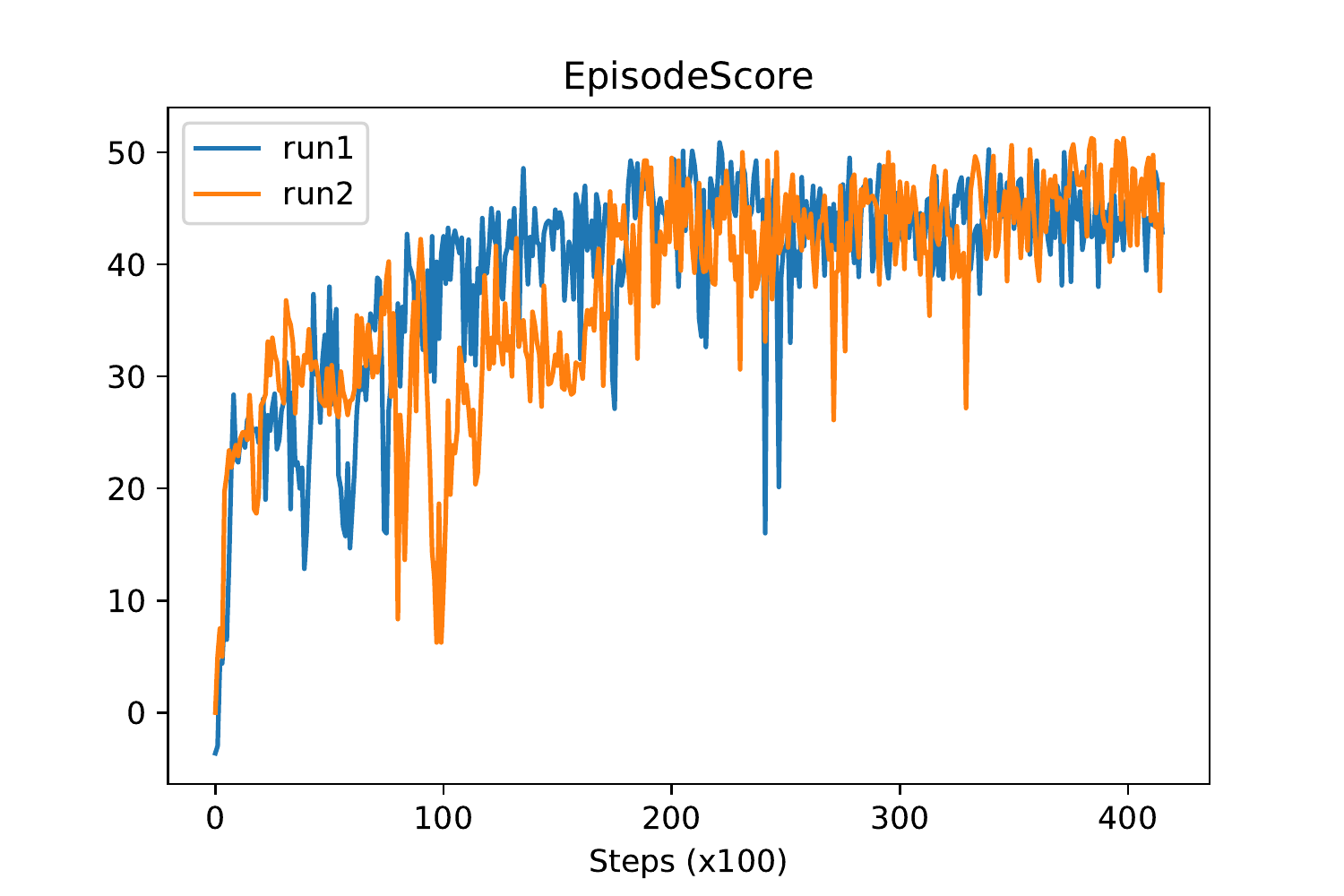}
                \caption{}
                \label{fig:tiger}
        \end{subfigure}%
        \begin{subfigure}{0.50\textwidth}
                \includegraphics[width=0.8\linewidth]{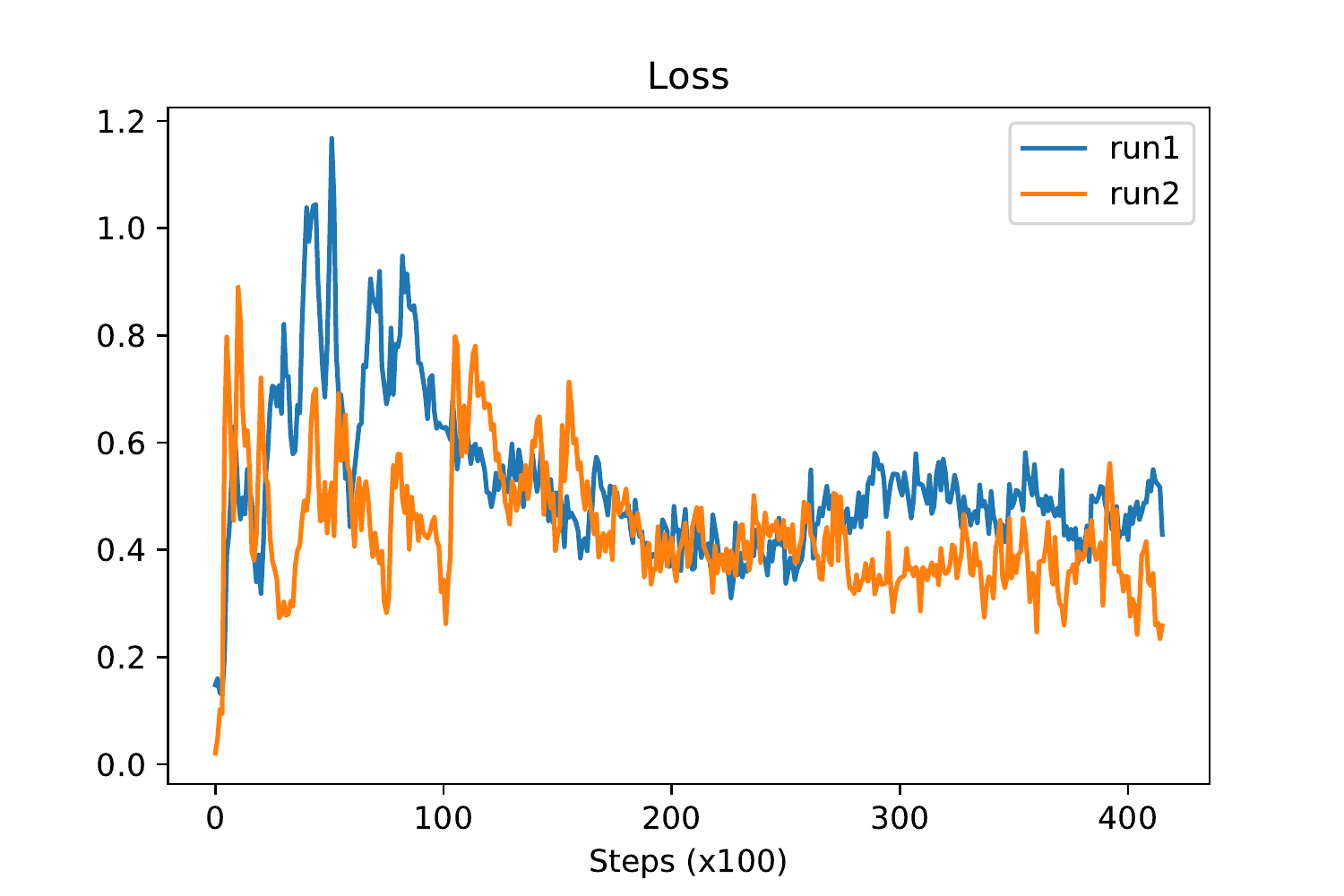}
                \caption{}
                \label{fig:mouse}
        \end{subfigure}
        \caption{Results for \texttt{Zork1}: Two independent runs of (DBERT-DRRN)}
        \label{fig:animals}
\end{figure}

The learning curves for \texttt{Zork1} are shown in Figure \ref{fig:animals}. 

\newpage 
\clearpage

\section{Gameplay Transcripts for Trained DBERT-DRRN}

The transcripts presents the game-play over an episode of 100 steps. For each step, the game state (observation, inventory, description), list of valid actions and corresponding Q-Values estimated by the agent, the action taken, reward received, cumulative score, and whether the episode is over (Done) is given.

List of transcripts:
\begin{itemize}
\item \hyperref[appendix:zork1]{\textbf{D.1} \texttt{Zork1}}
\begin{itemize}
    \item \hyperref[appendix:drrn]{\textbf{D.1.1} DRRN (for comparison)}
\end{itemize}
\item \hyperref[appendix:Inhuman]{\textbf{D.2} \texttt{Inhumane}}
\end{itemize}

\subsection{\texttt{Zork1}} \label{appendix:zork1}

STATE 0\\ Copyright ( c ) 1981, 1982, 1983 Infocom, Inc. All rights reserved. ZORK is a registered trademark of Infocom, Inc. Revision 88 / Serial number 840726 West of House You are standing in an open field west of a white house, with a boarded front door. There is a small mailbox here. You are empty handed. West of House You are standing in an open field west of a white house, with a boarded front door. There is a small mailbox here. 
Actions:  ['west', 'north', 'south', 'open mailbox']
Qvalues:  [9.07, 7.74, 7.42, 5.98]
  Action:  west
Reward:  0, Score 0, Done False
\\ \\ STATE 1 \\
forest this is a forest, with trees in all directions. to the east, there appears to be sunlight. you are empty handed. forest this is a forest, with trees in all directions. to the east, there appears to be sunlight. 
Actions:  ['east', 'go around trees', 'north', 'south']
Qvalues:  [10.05, 7.14, 6.8, 6.64]
  Action:  south
Reward:  0, Score 0, Done False
\\ \\ STATE 2 \\
forest this is a dimly lit forest, with large trees all around. you are empty handed. forest this is a dimly lit forest, with large trees all around. 
Actions:  ['northwest', 'north', 'go around trees', 'west']
Qvalues:  [6.59, 6.52, 6.1, 5.37]
  Action:  northwest
Reward:  0, Score 0, Done False
\\ \\ STATE 3 \\ south of house you are facing the south side of a white house. there is no door here, and all the windows are boarded. you are empty handed. south of house you are facing the south side of a white house. there is no door here, and all the windows are boarded. 
Actions:  ['east', 'south', 'west']
Qvalues:  [7.71, 6.15, 5.75]
  Action:  east
Reward:  0, Score 0, Done False
\\ \\ STATE 4 \\ behind house you are behind the white house. a path leads into the forest to the east. in one corner of the house there is a small window which is slightly ajar. you are empty handed. behind house you are behind the white house. a path leads into the forest to the east. in one corner of the house there is a small window which is slightly ajar. 
Actions:  ['open window', 'east', 'north', 'south']
Qvalues:  [9.39, 7.99, 7.91, 7.4]
  Action:  east
Reward:  0, Score 0, Done False
\\ \\ STATE 5 \\ clearing you are in a small clearing in a well marked forest path that extends to the east and west. you are empty handed. clearing you are in a small clearing in a well marked forest path that extends to the east and west. 
Actions:  ['west', 'north', 'east', 'south', 'go around forest']
Qvalues:  [7.97, 6.53, 6.35, 6.21, 5.76]
  Action:  north
Reward:  0, Score 0, Done False
\\ \\ STATE 6 \\ forest this is a dimly lit forest, with large trees all around. you are empty handed. you hear in the distance the chirping of a song bird. forest this is a dimly lit forest, with large trees all around. you hear in the distance the chirping of a song bird. 
Actions:  ['south', 'go around trees', 'west', 'east']
Qvalues:  [6.83, 6.45, 6.22, 5.98]
  Action:  west
Reward:  0, Score 0, Done False
\\ \\ STATE 7 \\ forest path this is a path winding through a dimly lit forest. the path heads north south here. one particularly large tree with some low branches stands at the edge of the path. you are empty handed. forest path this is a path winding through a dimly lit forest. the path heads north south here. one particularly large tree with some low branches stands at the edge of the path. you hear in the distance the chirping of a song bird. 
Actions:  ['up', 'go around forest', 'south', 'east', 'north', 'west']
Qvalues:  [10.91, 8.65, 8.4, 7.61, 7.09, 5.01]
  Action:  up
Reward:  0, Score 0, Done False
\\ \\ STATE 8 \\ up a tree you are about 10 feet above the ground nestled among some large branches. the nearest branch above you is above your reach. beside you on the branch is a small bird's nest. in the bird's nest is a large egg encrusted with precious jewels, apparently scavenged by a childless songbird. the egg is covered with fine gold inlay, and ornamented in lapis lazuli and mother of pearl. unlike most eggs, this one is hinged and closed with a delicate looking clasp. the egg appears extremely fragile. you are empty handed. up a tree you are about 10 feet above the ground nestled among some large branches. the nearest branch above you is above your reach. beside you on the branch is a small bird's nest. in the bird's nest is a large egg encrusted with precious jewels, apparently scavenged by a childless songbird. the egg is covered with fine gold inlay, and ornamented in lapis lazuli and mother of pearl. unlike most eggs, this one is hinged and closed with a delicate looking clasp. the egg appears extremely fragile. 
Actions:  ['take egg', 'close nest', 'take nest', 'take on egg', 'down']
Qvalues:  [11.59, 4.09, 2.25, 0.82, -4.72]
  Action:  take egg
Reward:  5, Score 5, Done False
\\ \\ STATE 9 \\ taken. you are carrying : a jewel encrusted egg up a tree you are about 10 feet above the ground nestled among some large branches. the nearest branch above you is above your reach. beside you on the branch is a small bird's nest. you hear in the distance the chirping of a song bird. 
Actions:  ['down', 'close nest', 'put egg in nest', 'take nest', 'take on egg', 'throw egg at nest', 'put down egg']
Qvalues:  [8.07, 6.47, 5.62, 5.51, 5.29, 5.04, 4.88]
  Action:  down
Reward:  0, Score 5, Done False
\\ \\ STATE 10 \\ forest path you are carrying : a jewel encrusted egg forest path this is a path winding through a dimly lit forest. the path heads north south here. one particularly large tree with some low branches stands at the edge of the path. you hear in the distance the chirping of a song bird. 
Actions:  ['south', 'go around forest', 'north', 'east', 'west', 'open egg with all', 'up', 'take on egg', 'throw egg at tree', 'put down egg']
Qvalues:  [8.94, 8.82, 8.04, 7.9, 7.49, 7.45, 7.02, 6.23, 6.19, 5.91]
  Action:  east
Reward:  0, Score 5, Done False
\\ \\ STATE 11 \\ forest you are carrying : a jewel encrusted egg you hear in the distance the chirping of a song bird. forest this is a dimly lit forest, with large trees all around. 
Actions:  ['south', 'go around trees', 'east', 'west', 'open egg with all', 'throw egg at large', 'take on egg', 'put down egg']
Qvalues:  [8.72, 7.81, 7.81, 7.49, 7.4, 6.49, 6.14, 6.09]
  Action:  go around trees
Reward:  0, Score 5, Done False
\\ \\ STATE 12 \\ forest you are carrying : a jewel encrusted egg forest this is a dimly lit forest, with large trees all around. 
Actions:  ['northwest', 'north', 'go around trees', 'west', 'open egg with all', 'take on egg', 'throw egg at large', 'put down egg']
Qvalues:  [8.56, 8.06, 7.97, 7.58, 7.44, 6.38, 6.35, 6.14]
  Action:  northwest
Reward:  0, Score 5, Done False
\\ \\ STATE 13 \\ south of house you are carrying : a jewel encrusted egg south of house you are facing the south side of a white house. there is no door here, and all the windows are boarded. 
Actions:  ['east', 'south', 'open egg with all', 'west', 'take on egg', 'put down egg', 'throw egg at house']
Qvalues:  [9.39, 8.03, 7.8, 7.47, 5.81, 5.77, 5.59]
  Action:  east
Reward:  0, Score 5, Done False
\\ \\ STATE 14 \\ behind house you are carrying : a jewel encrusted egg behind house you are behind the white house. a path leads into the forest to the east. in one corner of the house there is a small window which is slightly ajar. 
Actions:  ['open window', 'open egg with all', 'south', 'east', 'north', 'put down egg', 'take on egg', 'throw egg at path']
Qvalues:  [10.97, 9.79, 9.03, 8.74, 8.71, 7.78, 7.72, 6.85]
  Action:  open window
Reward:  0, Score 5, Done False
\\ \\ STATE 15 \\ with great effort, you open the window far enough to allow entry. you are carrying : a jewel encrusted egg behind house you are behind the white house. a path leads into the forest to the east. in one corner of the house there is a small window which is open. 
Actions:  ['west', 'close window', 'open egg with all', 'east', 'south', 'north', 'put down egg', 'take on egg', 'throw egg at path']
Qvalues:  [11.56, 9.85, 9.56, 9.54, 9.12, 8.97, 7.34, 6.58, 6.2]
  Action:  east
Reward:  0, Score 5, Done False
\\ \\ STATE 16 \\ clearing you are carrying : a jewel encrusted egg clearing you are in a small clearing in a well marked forest path that extends to the east and west. 
Actions:  ['west', 'open egg with all', 'east', 'south', 'north', 'go around forest', 'put down egg', 'throw egg at forest', 'take on egg']
Qvalues:  [9.27, 8.16, 8.13, 7.81, 7.76, 7.49, 6.35, 6.08, 5.98]
  Action:  west
Reward:  0, Score 5, Done False
\\ \\ STATE 17 \\ behind house you are carrying : a jewel encrusted egg behind house you are behind the white house. a path leads into the forest to the east. in one corner of the house there is a small window which is open. 
Actions:  ['open egg with all', 'close window', 'west', 'east', 'south', 'north', 'put down egg', 'take on egg', 'throw egg at path']
Qvalues:  [9.71, 9.64, 9.54, 8.65, 8.58, 8.54, 7.81, 7.27, 6.8]
  Action:  west
Reward:  10, Score 15, Done False
\\ \\ STATE 18 \\ kitchen you are in the kitchen of the white house. a table seems to have been used recently for the preparation of food. a passage leads to the west and a dark staircase can be seen leading upward. a dark chimney leads down and to the east is a small window which is open. on the table is an elongated brown sack, smelling of hot peppers. a bottle is sitting on the table. the glass bottle contains : a quantity of water you are carrying : a jewel encrusted egg kitchen you are in the kitchen of the white house. a table seems to have been used recently for the preparation of food. a passage leads to the west and a dark staircase can be seen leading upward. a dark chimney leads down and to the east is a small window which is open. on the table is an elongated brown sack, smelling of hot peppers. a bottle is sitting on the table. the glass bottle contains : a quantity of water 
Actions:  ['west', 'close window', 'open sack', 'east', 'open bottle', 'take sack', 'put down egg', 'put egg on table', 'throw egg at bottle', 'take on egg', 'take bottle', 'take all', 'up', 'jump']
Qvalues:  [21.67, 9.94, 9.52, 9.26, 8.74, 7.63, 7.6, 7.41, 6.13, 5.58, 4.72, 3.89, -2.46, -11.57]
  Action:  west
Reward:  0, Score 15, Done False
\\ \\ STATE 19 \\ living room you are in the living room. there is a doorway to the east, a wooden door with strange gothic lettering to the west, which appears to be nailed shut, a trophy case, and a large oriental rug in the center of the room. above the trophy case hangs an elvish sword of great antiquity. a battery powered brass lantern is on the trophy case. you are carrying : a jewel encrusted egg living room you are in the living room. there is a doorway to the east, a wooden door with strange gothic lettering to the west, which appears to be nailed shut, a trophy case, and a large oriental rug in the center of the room. above the trophy case hangs an elvish sword of great antiquity. a battery powered brass lantern is on the trophy case. 
Actions:  ['take lantern', 'push rug', 'open case', 'turn on lantern', 'throw egg at case', 'throw egg at lantern', 'take sword', 'put down egg', 'take on egg', 'take all', 'east']
Qvalues:  [25.17, 25.08, 24.82, 20.51, 20.26, 20.21, 17.84, 17.42, 17.37, 13.44, 13.38]
  Action:  push rug
Reward:  0, Score 15, Done False
\\ \\ STATE 20 \\ with a great effort, the rug is moved to one side of the room, revealing the dusty cover of a closed trap door. you are carrying : a jewel encrusted egg living room you are in the living room. there is a doorway to the east, a wooden door with strange gothic lettering to the west, which appears to be nailed shut, a trophy case, and a closed trap door at your feet. above the trophy case hangs an elvish sword of great antiquity. a battery powered brass lantern is on the trophy case. 
Actions:  ['take lantern', 'open case', 'open trap', 'turn on lantern', 'throw egg at case', 'throw egg at lantern', 'take sword', 'take on egg', 'put down egg', 'take all', 'east']
Qvalues:  [27.57, 26.49, 26.36, 22.97, 22.6, 21.73, 19.57, 19.16, 19.04, 15.18, 13.41]
  Action:  open case
Reward:  0, Score 15, Done False
\\ \\ STATE 21 \\ opened. you are carrying : a jewel encrusted egg living room you are in the living room. there is a doorway to the east, a wooden door with strange gothic lettering to the west, which appears to be nailed shut, a trophy case, and a closed trap door at your feet. above the trophy case hangs an elvish sword of great antiquity. a battery powered brass lantern is on the trophy case. 
Actions:  ['put egg in case', 'take lantern', 'open trap', 'close case', 'turn on lantern', 'throw egg at case', 'throw egg at lantern', 'fill case with egg', 'take sword', 'take on egg', 'put down egg', 'take all', 'east']
Qvalues:  [27.91, 27.9, 25.05, 24.38, 23.28, 22.29, 21.63, 20.13, 19.53, 19.27, 18.89, 15.31, 12.56]
  Action:  take lantern
Reward:  0, Score 15, Done False
\\ \\ STATE 22 \\ taken. you are carrying : a brass lantern a jewel encrusted egg living room you are in the living room. there is a doorway to the east, a wooden door with strange gothic lettering to the west, which appears to be nailed shut, a trophy case, and a closed trap door at your feet. above the trophy case hangs an elvish sword of great antiquity. 
Actions:  ['put egg in case', 'open trap', 'close case', 'open egg with lantern', 'put all in case', 'turn on lantern', 'throw egg at case', 'put lantern in case', 'put down lantern', 'fill case with egg', 'take sword', 'throw lantern at case', 'take on egg', 'throw lantern at egg', 'put down all', 'put down egg', 'east']
Qvalues:  [30.7, 28.07, 27.18, 26.67, 24.57, 23.87, 22.85, 22.05, 21.95, 21.19, 20.45, 20.32, 19.95, 19.49, 19.48, 19.32, 14.5]
  Action:  put egg in case
Reward:  5, Score 20, Done False
\\ \\ STATE 23 \\ done. you are carrying : a brass lantern living room you are in the living room. there is a doorway to the east, a wooden door with strange gothic lettering to the west, which appears to be nailed shut, a trophy case, and a closed trap door at your feet. above the trophy case hangs an elvish sword of great antiquity. your collection of treasures consists of : a jewel encrusted egg 
Actions:  ['open trap', 'close case', 'open egg with lantern', 'take egg from sword', 'turn on lantern', 'take egg', 'put down lantern', 'put lantern in case', 'throw lantern at case', 'take sword', 'throw lantern at egg', 'take on egg', 'east']
Qvalues:  [30.74, 27.85, 27.84, 27.56, 25.61, 24.67, 24.36, 23.05, 22.62, 22.33, 21.85, 21.0, 15.86]
  Action:  open trap
Reward:  0, Score 20, Done False
\\ \\ STATE 24 \\ the door reluctantly opens to reveal a rickety staircase descending into darkness. you are carrying : a brass lantern living room you are in the living room. there is a doorway to the east, a wooden door with strange gothic lettering to the west, which appears to be nailed shut, a trophy case, and a rug lying beside an open trap door. above the trophy case hangs an elvish sword of great antiquity. your collection of treasures consists of : a jewel encrusted egg 
Actions:  ['down', 'open egg with lantern', 'close case', 'take egg from sword', 'close trap', 'turn on lantern', 'take egg', 'put down lantern', 'put lantern in case', 'throw lantern at case', 'take sword', 'throw lantern at egg', 'take on egg', 'east']
Qvalues:  [32.52, 27.82, 27.35, 27.33, 26.53, 26.21, 24.31, 24.03, 23.0, 22.09, 21.81, 21.24, 20.86, 15.54]
  Action:  down
Reward:  25, Score 45, Done False
\\ \\ STATE 25 \\ you have moved into a dark place. the trap door crashes shut, and you hear someone barring it. it is pitch black. you are likely to be eaten by a grue. you are carrying : a brass lantern it is pitch black. you are likely to be eaten by a grue. 
Actions:  ['turn on lantern', 'put down lantern', 'throw lantern at grue', 'south', 'east', 'north']
Qvalues:  [7.76, -1.33, -3.82, -4.74, -7.48, -10.65]
  Action:  turn on lantern
Reward:  0, Score 45, Done False
\\ \\ STATE 26 \\ the brass lantern is now on. cellar you are in a dark and damp cellar with a narrow passageway leading north, and a crawlway to the south. on the west is the bottom of a steep metal ramp which is unclimbable. you are carrying : a brass lantern ( providing light ) cellar you are in a dark and damp cellar with a narrow passageway leading north, and a crawlway to the south. on the west is the bottom of a steep metal ramp which is unclimbable. 
Actions:  ['south', 'put out lantern', 'put down lantern', 'north', 'throw lantern at crawlway']
Qvalues:  [8.37, 4.86, 4.5, 4.2, 0.44]
  Action:  south
Reward:  0, Score 45, Done False
\\ \\ STATE 27 \\ east of chasm you are on the east edge of a chasm, the bottom of which cannot be seen. a narrow passage goes north, and the path you are on continues to the east. you are carrying : a brass lantern ( providing light ) east of chasm you are on the east edge of a chasm, the bottom of which cannot be seen. a narrow passage goes north, and the path you are on continues to the east. 
Actions:  ['east', 'north', 'put out lantern', 'put down lantern', 'throw lantern at path', 'jump']
Qvalues:  [9.28, 4.51, 4.16, 3.46, 0.51, -7.56]
  Action:  east
Reward:  0, Score 45, Done False
\\ \\ STATE 28 \\ gallery this is an art gallery. most of the paintings have been stolen by vandals with exceptional taste. the vandals left through either the north or west exits. fortunately, there is still one chance for you to be a vandal, for on the far wall is a painting of unparalleled beauty. you are carrying : a brass lantern ( providing light ) gallery this is an art gallery. most of the paintings have been stolen by vandals with exceptional taste. the vandals left through either the north or west exits. fortunately, there is still one chance for you to be a vandal, for on the far wall is a painting of unparalleled beauty. 
Actions:  ['take art', 'put out lantern', 'put down lantern', 'throw lantern at art', 'north', 'west']
Qvalues:  [10.35, 10.21, 8.89, 6.58, 6.06, 3.92]
  Action:  take art
Reward:  4, Score 49, Done False
\\ \\ STATE 29 \\ taken. you are carrying : a painting a brass lantern ( providing light ) gallery this is an art gallery. most of the paintings have been stolen by vandals with exceptional taste. the vandals left through either the north or west exits. 
Actions:  ['north', 'put out lantern', 'put down lantern', 'put down art', 'west', 'put down all', 'throw art at lantern']
Qvalues:  [7.74, 7.08, 6.64, 6.36, 6.36, 6.35, 4.02]
  Action:  north
Reward:  0, Score 49, Done False
\\ \\ STATE 30 \\ studio this appears to have been an artist's studio. the walls and floors are splattered with paints of 69 different colors. strangely enough, nothing of value is hanging here. at the south end of the room is an open door ( also covered with paint ). a dark and narrow chimney leads up from a fireplace ; although you might be able to get up it, it seems unlikely you could get back down. loosely attached to a wall is a small piece of paper. you are carrying : a painting a brass lantern ( providing light ) studio this appears to have been an artist's studio. the walls and floors are splattered with paints of 69 different colors. strangely enough, nothing of value is hanging here. at the south end of the room is an open door ( also covered with paint ). a dark and narrow chimney leads up from a fireplace ; although you might be able to get up it, it seems unlikely you could get back down. loosely attached to a wall is a small piece of paper. 
Actions:  ['up', 'put out lantern', 'south', 'put down lantern', 'take manual', 'put down painting', 'put down all', 'throw lantern at manual']
Qvalues:  [8.6, 7.77, 6.81, 6.69, 6.59, 6.38, 6.37, 4.9]
  Action:  put out lantern
Reward:  0, Score 49, Done False
\\ \\ STATE 31 \\ the brass lantern is now off. it is now pitch black. you are carrying : a painting a brass lantern it is pitch black. you are likely to be eaten by a grue. 
Actions:  ['up', 'turn on lantern', 'south', 'put down lantern', 'put down all', 'put down painting', 'throw lantern at grue', 'north']
Qvalues:  [8.42, 5.97, 5.91, 4.04, 3.45, 3.42, 2.89, 1.35]
  Action:  up
Reward:  0, Score 49, Done False
\\ \\ STATE 32 \\ kitchen on the table is an elongated brown sack, smelling of hot peppers. a bottle is sitting on the table. the glass bottle contains : a quantity of water you are carrying : a painting a brass lantern kitchen you are in the kitchen of the white house. a table seems to have been used recently for the preparation of food. a passage leads to the west and a dark staircase can be seen leading upward. a dark chimney leads down and to the east is a small window which is open. on the table is an elongated brown sack, smelling of hot peppers. a bottle is sitting on the table. the glass bottle contains : a quantity of water 
Actions:  ['west', 'open bottle', 'close window', 'open sack', 'east', 'put lantern on table', 'take sack', 'put down lantern', 'throw lantern at bottle', 'turn on lantern', 'put down all', 'put down painting', 'take bottle', 'take all', 'put painting on table', 'up', 'jump']
Qvalues:  [10.1, 9.01, 8.93, 8.91, 8.41, 8.23, 8.14, 8.08, 7.97, 7.87, 7.63, 7.56, 7.29, 6.6, 6.56, 5.92, -2.09]
  Action:  open bottle
Reward:  0, Score 49, Done False
\\ \\ STATE 33 \\ opened. you are carrying : a painting a brass lantern kitchen you are in the kitchen of the white house. a table seems to have been used recently for the preparation of food. a passage leads to the west and a dark staircase can be seen leading upward. a dark chimney leads down and to the east is a small window which is open. on the table is an elongated brown sack, smelling of hot peppers. sitting on the kitchen table is : a glass bottle the glass bottle contains : a quantity of water 
Actions:  ['west', 'open sack', 'close window', 'close bottle', 'east', 'put lantern on table', 'take sack', 'throw lantern at bottle', 'put down lantern', 'turn on lantern', 'put down all', 'put down painting', 'take bottle', 'put painting on table', 'take all', 'up', 'jump']
Qvalues:  [10.07, 9.15, 9.04, 8.95, 8.73, 8.54, 8.45, 8.35, 8.34, 8.17, 7.95, 7.73, 7.63, 6.95, 6.95, 5.69, -2.51]
  Action:  put lantern on table
Reward:  0, Score 49, Done False
\\ \\ STATE 34 \\ done. you are carrying : a painting kitchen you are in the kitchen of the white house. a table seems to have been used recently for the preparation of food. a passage leads to the west and a dark staircase can be seen leading upward. a dark chimney leads down and to the east is a small window which is open. on the table is an elongated brown sack, smelling of hot peppers. sitting on the kitchen table is : a brass lantern a glass bottle the glass bottle contains : a quantity of water 
Actions:  ['west', 'take lantern', 'open sack', 'close bottle', 'east', 'take sack', 'close window', 'throw painting at lantern', 'turn on lantern', 'take bottle', 'put down painting', 'take all', 'up', 'jump']
Qvalues:  [9.0, 8.75, 8.31, 8.18, 8.11, 8.08, 8.02, 7.95, 7.84, 7.16, 7.12, 6.95, 4.29, -4.68]
  Action:  close bottle
Reward:  0, Score 49, Done False
\\ \\ STATE 35 \\ closed. you are carrying : a painting kitchen you are in the kitchen of the white house. a table seems to have been used recently for the preparation of food. a passage leads to the west and a dark staircase can be seen leading upward. a dark chimney leads down and to the east is a small window which is open. on the table is an elongated brown sack, smelling of hot peppers. sitting on the kitchen table is : a brass lantern a glass bottle the glass bottle contains : a quantity of water 
Actions:  ['west', 'take lantern', 'open bottle', 'open sack', 'close window', 'take sack', 'east', 'throw painting at lantern', 'turn on lantern', 'take bottle', 'put down painting', 'take all', 'up', 'jump']
Qvalues:  [9.01, 8.76, 8.58, 8.53, 8.18, 8.18, 8.17, 7.97, 7.94, 7.29, 7.17, 7.02, 4.29, -4.77]
  Action:  take lantern
Reward:  0, Score 49, Done False
\\ \\ STATE 36 \\ taken. you are carrying : a brass lantern a painting kitchen you are in the kitchen of the white house. a table seems to have been used recently for the preparation of food. a passage leads to the west and a dark staircase can be seen leading upward. a dark chimney leads down and to the east is a small window which is open. on the table is an elongated brown sack, smelling of hot peppers. sitting on the kitchen table is : a glass bottle the glass bottle contains : a quantity of water 
Actions:  ['west', 'open bottle', 'open sack', 'close window', 'east', 'put lantern on table', 'take sack', 'throw lantern at bottle', 'put down lantern', 'turn on lantern', 'put down all', 'put down painting', 'take bottle', 'take all', 'put painting on table', 'up', 'jump']
Qvalues:  [9.82, 9.15, 9.08, 9.02, 8.61, 8.37, 8.31, 8.15, 8.15, 8.07, 7.82, 7.61, 7.53, 6.82, 6.8, 6.16, -1.96]
  Action:  open bottle
Reward:  0, Score 49, Done False
\\ \\ STATE 37 \\ opened. you are carrying : a brass lantern a painting kitchen you are in the kitchen of the white house. a table seems to have been used recently for the preparation of food. a passage leads to the west and a dark staircase can be seen leading upward. a dark chimney leads down and to the east is a small window which is open. on the table is an elongated brown sack, smelling of hot peppers. sitting on the kitchen table is : a glass bottle the glass bottle contains : a quantity of water 
Actions:  ['west', 'open sack', 'close window', 'close bottle', 'east', 'put lantern on table', 'take sack', 'put down lantern', 'throw lantern at bottle', 'turn on lantern', 'put down all', 'take bottle', 'put down painting', 'take all', 'put painting on table', 'up', 'jump']
Qvalues:  [9.93, 9.02, 8.99, 8.76, 8.6, 8.48, 8.27, 8.14, 8.12, 7.97, 7.84, 7.38, 7.38, 6.84, 6.52, 5.84, -2.44]
  Action:  close window
Reward:  0, Score 49, Done False
\\ \\ STATE 38 \\ the window closes ( more easily than it opened ). you are carrying : a brass lantern a painting kitchen you are in the kitchen of the white house. a table seems to have been used recently for the preparation of food. a passage leads to the west and a dark staircase can be seen leading upward. a dark chimney leads down and to the east is a small window which is slightly ajar. on the table is an elongated brown sack, smelling of hot peppers. sitting on the kitchen table is : a glass bottle the glass bottle contains : a quantity of water 
Actions:  ['open window', 'west', 'open sack', 'close bottle', 'put lantern on table', 'take sack', 'put down lantern', 'throw lantern at bottle', 'turn on lantern', 'put down all', 'put down painting', 'take bottle', 'take all', 'put painting on table', 'up', 'jump']
Qvalues:  [9.55, 9.15, 8.8, 8.59, 8.13, 8.0, 7.89, 7.86, 7.71, 7.48, 7.33, 7.22, 6.5, 6.43, 6.4, -1.57]
  Action:  put down all
Reward:  0, Score 49, Done False
\\ \\ STATE 39 \\ brass lantern : dropped. painting : dropped. you are empty handed. kitchen you are in the kitchen of the white house. a table seems to have been used recently for the preparation of food. a passage leads to the west and a dark staircase can be seen leading upward. a dark chimney leads down and to the east is a small window which is slightly ajar. a painting by a neglected genius is here. there is a brass lantern ( battery powered ) here. on the table is an elongated brown sack, smelling of hot peppers. sitting on the kitchen table is : a glass bottle the glass bottle contains : a quantity of water 
Actions:  ['open window', 'open sack', 'close bottle', 'turn on lantern', 'west', 'take painting', 'take all', 'take lantern', 'take sack', 'take bottle', 'take all from table', 'up', 'jump']
Qvalues:  [6.81, 6.37, 6.11, 6.08, 5.92, 5.9, 5.89, 5.83, 5.7, 5.25, 5.0, 2.44, -6.37]
  Action:  open window
Reward:  0, Score 49, Done False
\\ \\ STATE 40 \\ with great effort, you open the window far enough to allow entry. you are empty handed. kitchen you are in the kitchen of the white house. a table seems to have been used recently for the preparation of food. a passage leads to the west and a dark staircase can be seen leading upward. a dark chimney leads down and to the east is a small window which is open. a painting by a neglected genius is here. there is a brass lantern ( battery powered ) here. on the table is an elongated brown sack, smelling of hot peppers. sitting on the kitchen table is : a glass bottle the glass bottle contains : a quantity of water 
Actions:  ['east', 'open sack', 'west', 'close bottle', 'close window', 'turn on lantern', 'take sack', 'take lantern', 'take all', 'take painting', 'take bottle', 'take all from table', 'up', 'jump']
Qvalues:  [6.76, 6.36, 6.29, 6.08, 6.06, 5.9, 5.59, 5.56, 5.48, 5.17, 5.14, 4.9, 0.7, -8.47]
  Action:  east
Reward:  0, Score 49, Done False
\\ \\ STATE 41 \\ behind house you are empty handed. behind house you are behind the white house. a path leads into the forest to the east. in one corner of the house there is a small window which is open. 
Actions:  ['close window', 'north', 'west', 'east', 'south']
Qvalues:  [7.41, 6.37, 6.31, 6.13, 5.86]
  Action:  west
Reward:  0, Score 49, Done False
\\ \\ STATE 42 \\ kitchen a painting by a neglected genius is here. there is a brass lantern ( battery powered ) here. on the table is an elongated brown sack, smelling of hot peppers. sitting on the kitchen table is : a glass bottle the glass bottle contains : a quantity of water you are empty handed. kitchen you are in the kitchen of the white house. a table seems to have been used recently for the preparation of food. a passage leads to the west and a dark staircase can be seen leading upward. a dark chimney leads down and to the east is a small window which is open. a painting by a neglected genius is here. there is a brass lantern ( battery powered ) here. on the table is an elongated brown sack, smelling of hot peppers. sitting on the kitchen table is : a glass bottle the glass bottle contains : a quantity of water 
Actions:  ['west', 'east', 'open sack', 'close window', 'close bottle', 'take lantern', 'turn on lantern', 'take sack', 'take painting', 'take bottle', 'take all', 'take all from table', 'up', 'jump']
Qvalues:  [9.43, 8.18, 7.32, 7.28, 6.96, 6.84, 6.75, 6.41, 5.8, 5.8, 5.79, 5.22, 1.47, -7.43]
  Action:  take painting
Reward:  0, Score 49, Done False
\\ \\ STATE 43 \\ taken. you are carrying : a painting kitchen you are in the kitchen of the white house. a table seems to have been used recently for the preparation of food. a passage leads to the west and a dark staircase can be seen leading upward. a dark chimney leads down and to the east is a small window which is open. there is a brass lantern ( battery powered ) here. on the table is an elongated brown sack, smelling of hot peppers. sitting on the kitchen table is : a glass bottle the glass bottle contains : a quantity of water 
Actions:  ['take lantern', 'west', 'east', 'open sack', 'close bottle', 'take sack', 'close window', 'turn on lantern', 'throw painting at lantern', 'take bottle', 'take all', 'put down painting', 'take all from table', 'put painting on table', 'up', 'jump']
Qvalues:  [8.45, 8.2, 8.0, 7.88, 7.75, 7.68, 7.63, 7.61, 7.42, 6.97, 6.81, 6.75, 6.24, 6.19, 5.3, -3.67]
  Action:  take lantern
Reward:  0, Score 49, Done False
\\ \\ STATE 44 \\ taken. you are carrying : a brass lantern a painting kitchen you are in the kitchen of the white house. a table seems to have been used recently for the preparation of food. a passage leads to the west and a dark staircase can be seen leading upward. a dark chimney leads down and to the east is a small window which is open. on the table is an elongated brown sack, smelling of hot peppers. sitting on the kitchen table is : a glass bottle the glass bottle contains : a quantity of water 
Actions:  ['west', 'open sack', 'close window', 'close bottle', 'east', 'put lantern on table', 'take sack', 'turn on lantern', 'put down lantern', 'throw lantern at bottle', 'put down all', 'take bottle', 'put down painting', 'take all', 'put painting on table', 'up', 'jump']
Qvalues:  [9.89, 8.94, 8.93, 8.69, 8.64, 8.29, 8.14, 8.0, 8.0, 7.92, 7.64, 7.39, 7.27, 6.79, 6.4, 6.21, -2.21]
  Action:  west
Reward:  0, Score 49, Done False
\\ \\ STATE 45 \\ living room above the trophy case hangs an elvish sword of great antiquity. your collection of treasures consists of : a jewel encrusted egg you are carrying : a brass lantern a painting living room you are in the living room. there is a doorway to the east, a wooden door with strange gothic lettering to the west, which appears to be nailed shut, a trophy case, and a closed trap door at your feet. above the trophy case hangs an elvish sword of great antiquity. your collection of treasures consists of : a jewel encrusted egg 
Actions:  ['put painting in case', 'put all in case', 'put down painting', 'take treasures from sword', 'put down all', 'open egg with lantern', 'put down lantern', 'put lantern in case', 'take sword', 'close case', 'turn on lantern', 'throw lantern at case', 'throw painting at treasures', 'open trap', 'throw lantern at treasures', 'take treasures', 'take on treasures', 'east']
Qvalues:  [10.96, 10.79, 10.25, 10.18, 9.93, 9.89, 9.77, 9.74, 9.61, 9.59, 9.55, 9.5, 9.43, 9.4, 8.77, 8.09, 7.49, 6.26]
  Action:  put all in case
Reward:  6, Score 55, Done False
\\ \\ STATE 46 \\ brass lantern : done. painting : done. you are empty handed. living room you are in the living room. there is a doorway to the east, a wooden door with strange gothic lettering to the west, which appears to be nailed shut, a trophy case, and a closed trap door at your feet. above the trophy case hangs an elvish sword of great antiquity. your collection of treasures consists of : a painting a brass lantern a jewel encrusted egg 
Actions:  ['east', 'open trap', 'turn on lantern', 'take lantern', 'close case', 'take on egg', 'take egg from sword', 'take sword', 'take painting', 'take egg', 'take all from case']
Qvalues:  [4.89, 4.27, 4.15, 3.98, 3.98, 3.72, 3.7, 3.69, 3.19, 3.05, 1.15]
  Action:  east
Reward:  0, Score 55, Done False
\\ \\ STATE 47 \\ kitchen on the table is an elongated brown sack, smelling of hot peppers. sitting on the kitchen table is : a glass bottle the glass bottle contains : a quantity of water you are empty handed. kitchen you are in the kitchen of the white house. a table seems to have been used recently for the preparation of food. a passage leads to the west and a dark staircase can be seen leading upward. a dark chimney leads down and to the east is a small window which is open. on the table is an elongated brown sack, smelling of hot peppers. sitting on the kitchen table is : a glass bottle the glass bottle contains : a quantity of water 
Actions:  ['west', 'east', 'open sack', 'close window', 'close bottle', 'take sack', 'take bottle', 'take all', 'up', 'jump']
Qvalues:  [7.6, 7.38, 6.33, 6.29, 6.05, 5.44, 5.09, 4.99, 1.82, -7.47]
  Action:  east
Reward:  0, Score 55, Done False
\\ \\ STATE 48 \\ behind house you are empty handed. behind house you are behind the white house. a path leads into the forest to the east. in one corner of the house there is a small window which is open. 
Actions:  ['close window', 'north', 'east', 'west', 'south']
Qvalues:  [7.72, 6.65, 6.54, 6.41, 6.24]
  Action:  close window
Reward:  0, Score 55, Done False
\\ \\ STATE 49 \\ the window closes ( more easily than it opened ). you are empty handed. behind house you are behind the white house. a path leads into the forest to the east. in one corner of the house there is a small window which is slightly ajar. 
Actions:  ['open window', 'north', 'south', 'east']
Qvalues:  [8.61, 6.8, 6.53, 6.42]
  Action:  open window
Reward:  0, Score 55, Done False
\\ \\ STATE 50 \\ with great effort, you open the window far enough to allow entry. you are empty handed. behind house you are behind the white house. a path leads into the forest to the east. in one corner of the house there is a small window which is open. 
Actions:  ['west', 'close window', 'north', 'east', 'south']
Qvalues:  [9.23, 7.55, 6.84, 6.76, 6.48]
  Action:  west
Reward:  0, Score 55, Done False
\\ \\ STATE 51 \\ kitchen on the table is an elongated brown sack, smelling of hot peppers. sitting on the kitchen table is : a glass bottle the glass bottle contains : a quantity of water you are empty handed. kitchen you are in the kitchen of the white house. a table seems to have been used recently for the preparation of food. a passage leads to the west and a dark staircase can be seen leading upward. a dark chimney leads down and to the east is a small window which is open. on the table is an elongated brown sack, smelling of hot peppers. sitting on the kitchen table is : a glass bottle the glass bottle contains : a quantity of water 
Actions:  ['east', 'open sack', 'close bottle', 'close window', 'take sack', 'take bottle', 'take all', 'west', 'up', 'jump']
Qvalues:  [7.0, 6.25, 5.99, 5.77, 5.28, 5.25, 5.2, 5.18, 3.37, -6.41]
  Action:  take sack
Reward:  0, Score 55, Done False
\\ \\ STATE 52 \\ taken. you are carrying : a brown sack kitchen you are in the kitchen of the white house. a table seems to have been used recently for the preparation of food. a passage leads to the west and a dark staircase can be seen leading upward. a dark chimney leads down and to the east is a small window which is open. sitting on the kitchen table is : a glass bottle the glass bottle contains : a quantity of water 
Actions:  ['put sack on table', 'east', 'put down sack', 'west', 'take bottle', 'close bottle', 'open sack', 'close window', 'up', 'jump']
Qvalues:  [5.79, 5.65, 5.52, 5.38, 5.1, 5.03, 5.0, 4.56, 3.88, -4.0]
  Action:  put sack on table
Reward:  0, Score 55, Done False
\\ \\ STATE 53 \\ done. you are empty handed. kitchen you are in the kitchen of the white house. a table seems to have been used recently for the preparation of food. a passage leads to the west and a dark staircase can be seen leading upward. a dark chimney leads down and to the east is a small window which is open. sitting on the kitchen table is : a brown sack a glass bottle the glass bottle contains : a quantity of water 
Actions:  ['east', 'open sack', 'west', 'close bottle', 'close window', 'take sack', 'take bottle', 'take all', 'up', 'jump']
Qvalues:  [6.97, 5.93, 5.76, 5.66, 5.45, 5.06, 5.0, 4.82, 3.54, -5.91]
  Action:  east
Reward:  0, Score 55, Done False
\\ \\ STATE 54 \\ behind house you are empty handed. behind house you are behind the white house. a path leads into the forest to the east. in one corner of the house there is a small window which is open. 
Actions:  ['close window', 'north', 'east', 'west', 'south']
Qvalues:  [7.59, 6.53, 6.4, 6.29, 6.18]
  Action:  south
Reward:  0, Score 55, Done False
\\ \\ STATE 55 \\ south of house you are empty handed. south of house you are facing the south side of a white house. there is no door here, and all the windows are boarded. 
Actions:  ['east', 'south', 'west']
Qvalues:  [6.57, 5.66, 4.74]
  Action:  east
Reward:  0, Score 55, Done False
\\ \\ STATE 56 \\ behind house you are empty handed. behind house you are behind the white house. a path leads into the forest to the east. in one corner of the house there is a small window which is open. 
Actions:  ['close window', 'north', 'east', 'south', 'west']
Qvalues:  [7.37, 6.67, 6.36, 6.29, 5.95]
  Action:  close window
Reward:  0, Score 55, Done False
\\ \\ STATE 57 \\ the window closes ( more easily than it opened ). you are empty handed. behind house you are behind the white house. a path leads into the forest to the east. in one corner of the house there is a small window which is slightly ajar. 
Actions:  ['open window', 'north', 'south', 'east']
Qvalues:  [8.24, 6.57, 6.44, 6.12]
  Action:  open window
Reward:  0, Score 55, Done False
\\ \\ STATE 58 \\ with great effort, you open the window far enough to allow entry. you are empty handed. behind house you are behind the white house. a path leads into the forest to the east. in one corner of the house there is a small window which is open. 
Actions:  ['west', 'close window', 'north', 'east', 'south']
Qvalues:  [8.37, 7.33, 6.58, 6.38, 6.19]
  Action:  close window
Reward:  0, Score 55, Done False
\\ \\ STATE 59 \\ the window closes ( more easily than it opened ). you are empty handed. behind house you are behind the white house. a path leads into the forest to the east. in one corner of the house there is a small window which is slightly ajar. 
Actions:  ['open window', 'north', 'south', 'east']
Qvalues:  [8.14, 6.58, 6.43, 6.16]
  Action:  east
Reward:  0, Score 55, Done False
\\ \\ STATE 60 \\ clearing you are empty handed. clearing you are in a small clearing in a well marked forest path that extends to the east and west. 
Actions:  ['west', 'south', 'north', 'east', 'go around forest']
Qvalues:  [6.88, 5.64, 5.62, 5.59, 5.35]
  Action:  south
Reward:  0, Score 55, Done False
\\ \\ STATE 61 \\ forest you are empty handed. forest this is a dimly lit forest, with large trees all around. 
Actions:  ['northwest', 'north', 'go around trees', 'west']
Qvalues:  [6.36, 6.01, 5.76, 5.31]
  Action:  northwest
Reward:  0, Score 55, Done False
\\ \\ STATE 62 \\ south of house you are empty handed. south of house you are facing the south side of a white house. there is no door here, and all the windows are boarded. 
Actions:  ['east', 'south', 'west']
Qvalues:  [6.54, 5.42, 5.03]
  Action:  east
Reward:  0, Score 55, Done False
\\ \\ STATE 63 \\ behind house you are empty handed. behind house you are behind the white house. a path leads into the forest to the east. in one corner of the house there is a small window which is slightly ajar. 
Actions:  ['open window', 'north', 'south', 'east']
Qvalues:  [8.04, 6.16, 6.05, 5.93]
  Action:  open window
Reward:  0, Score 55, Done False
\\ \\ STATE 64 \\ with great effort, you open the window far enough to allow entry. you are empty handed. behind house you are behind the white house. a path leads into the forest to the east. in one corner of the house there is a small window which is open. 
Actions:  ['west', 'close window', 'north', 'east', 'south']
Qvalues:  [8.2, 7.53, 7.02, 6.89, 6.46]
  Action:  west
Reward:  0, Score 55, Done False
\\ \\ STATE 65 \\ kitchen sitting on the kitchen table is : a brown sack a glass bottle the glass bottle contains : a quantity of water you are empty handed. kitchen you are in the kitchen of the white house. a table seems to have been used recently for the preparation of food. a passage leads to the west and a dark staircase can be seen leading upward. a dark chimney leads down and to the east is a small window which is open. sitting on the kitchen table is : a brown sack a glass bottle the glass bottle contains : a quantity of water 
Actions:  ['east', 'west', 'open sack', 'close bottle', 'close window', 'take sack', 'take bottle', 'take all', 'up', 'jump']
Qvalues:  [7.18, 6.16, 5.93, 5.53, 5.5, 4.9, 4.69, 4.64, 2.8, -6.25]
  Action:  west
Reward:  0, Score 55, Done False
\\ \\ STATE 66 \\ living room above the trophy case hangs an elvish sword of great antiquity. your collection of treasures consists of : a painting a brass lantern a jewel encrusted egg you are empty handed. living room you are in the living room. there is a doorway to the east, a wooden door with strange gothic lettering to the west, which appears to be nailed shut, a trophy case, and a closed trap door at your feet. above the trophy case hangs an elvish sword of great antiquity. your collection of treasures consists of : a painting a brass lantern a jewel encrusted egg 
Actions:  ['east', 'open trap', 'close case', 'turn on lantern', 'take egg from sword', 'take sword', 'take on egg', 'take lantern', 'take egg', 'take painting', 'take all from case']
Qvalues:  [5.2, 4.77, 4.54, 4.41, 4.18, 4.15, 4.09, 4.06, 3.41, 3.14, 1.54]
  Action:  east
Reward:  0, Score 55, Done False
\\ \\ STATE 67 \\ kitchen sitting on the kitchen table is : a brown sack a glass bottle the glass bottle contains : a quantity of water you are empty handed. kitchen you are in the kitchen of the white house. a table seems to have been used recently for the preparation of food. a passage leads to the west and a dark staircase can be seen leading upward. a dark chimney leads down and to the east is a small window which is open. sitting on the kitchen table is : a brown sack a glass bottle the glass bottle contains : a quantity of water 
Actions:  ['east', 'open sack', 'close bottle', 'west', 'close window', 'take all', 'take sack', 'take bottle', 'up', 'jump']
Qvalues:  [6.58, 5.57, 5.2, 5.13, 5.03, 4.57, 4.56, 4.5, 3.0, -6.25]
  Action:  open sack
Reward:  0, Score 55, Done False
\\ \\ STATE 68 \\ opening the brown sack reveals a lunch, and a clove of garlic. you are empty handed. kitchen you are in the kitchen of the white house. a table seems to have been used recently for the preparation of food. a passage leads to the west and a dark staircase can be seen leading upward. a dark chimney leads down and to the east is a small window which is open. sitting on the kitchen table is : a brown sack the brown sack contains : a lunch a clove of garlic a glass bottle the glass bottle contains : a quantity of water 
Actions:  ['east', 'west', 'close sack', 'close bottle', 'close window', 'take all from sack', 'take sack', 'take garlic', 'take food', 'take all', 'take bottle', 'up', 'jump']
Qvalues:  [6.66, 5.33, 5.29, 5.0, 4.84, 4.49, 4.48, 4.46, 4.45, 4.26, 4.17, 2.18, -6.95]
  Action:  close bottle
Reward:  0, Score 55, Done False
\\ \\ STATE 69 \\ closed. you are empty handed. kitchen you are in the kitchen of the white house. a table seems to have been used recently for the preparation of food. a passage leads to the west and a dark staircase can be seen leading upward. a dark chimney leads down and to the east is a small window which is open. sitting on the kitchen table is : a brown sack the brown sack contains : a lunch a clove of garlic a glass bottle the glass bottle contains : a quantity of water 
Actions:  ['east', 'open bottle', 'close sack', 'west', 'take all from sack', 'take sack', 'close window', 'take food', 'take garlic', 'take bottle', 'take all', 'up', 'jump']
Qvalues:  [6.26, 5.2, 5.13, 4.97, 4.6, 4.55, 4.52, 4.43, 4.39, 4.33, 4.22, 2.04, -7.58]
  Action:  take all from sack
Reward:  0, Score 55, Done False
\\ \\ STATE 70 \\ lunch : taken. clove of garlic : taken. you are carrying : a clove of garlic a lunch kitchen you are in the kitchen of the white house. a table seems to have been used recently for the preparation of food. a passage leads to the west and a dark staircase can be seen leading upward. a dark chimney leads down and to the east is a small window which is open. sitting on the kitchen table is : a brown sack a glass bottle the glass bottle contains : a quantity of water 
Actions:  ['put all in sack', 'put all on table', 'put down all', 'put food on table', 'put garlic on table', 'put down food', 'west', 'east', 'put food in sack', 'take sack', 'open bottle', 'put garlic in sack', 'put down garlic', 'close sack', 'take bottle', 'take all', 'close window', 'up', 'jump']
Qvalues:  [5.74, 5.73, 5.59, 5.43, 5.38, 5.34, 5.34, 5.3, 5.27, 5.26, 5.25, 5.18, 5.16, 4.96, 4.91, 4.61, 4.48, 4.17, -2.97]
  Action:  put down garlic
Reward:  0, Score 55, Done False
\\ \\ STATE 71 \\ dropped. you are carrying : a lunch kitchen you are in the kitchen of the white house. a table seems to have been used recently for the preparation of food. a passage leads to the west and a dark staircase can be seen leading upward. a dark chimney leads down and to the east is a small window which is open. there is a clove of garlic here. sitting on the kitchen table is : a brown sack a glass bottle the glass bottle contains : a quantity of water 
Actions:  ['put food on table', 'put down food', 'east', 'west', 'take garlic', 'put food in sack', 'take sack', 'open bottle', 'close sack', 'take bottle', 'take all', 'close window', 'take all from table', 'up', 'jump']
Qvalues:  [5.24, 5.24, 5.21, 5.16, 5.11, 4.98, 4.98, 4.86, 4.77, 4.57, 4.45, 4.25, 4.16, 2.27, -6.13]
  Action:  put food in sack
Reward:  0, Score 55, Done False
\\ \\ STATE 72 \\ done. you are empty handed. kitchen you are in the kitchen of the white house. a table seems to have been used recently for the preparation of food. a passage leads to the west and a dark staircase can be seen leading upward. a dark chimney leads down and to the east is a small window which is open. there is a clove of garlic here. sitting on the kitchen table is : a brown sack the brown sack contains : a lunch a glass bottle the glass bottle contains : a quantity of water 
Actions:  ['east', 'close sack', 'open bottle', 'west', 'take food', 'take garlic', 'take sack', 'take all', 'close window', 'take bottle', 'take all from table', 'up', 'jump']
Qvalues:  [6.19, 5.16, 5.15, 4.86, 4.61, 4.54, 4.51, 4.48, 4.48, 4.37, 3.97, 2.19, -7.34]
  Action:  take sack
Reward:  0, Score 55, Done False
\\ \\ STATE 73 \\ taken. you are carrying : a brown sack the brown sack contains : a lunch kitchen you are in the kitchen of the white house. a table seems to have been used recently for the preparation of food. a passage leads to the west and a dark staircase can be seen leading upward. a dark chimney leads down and to the east is a small window which is open. there is a clove of garlic here. sitting on the kitchen table is : a glass bottle the glass bottle contains : a quantity of water 
Actions:  ['put sack on table', 'put food on table', 'put down sack', 'put down food', 'west', 'take food', 'open bottle', 'shake sack', 'take garlic', 'east', 'close sack', 'take bottle', 'take all', 'close window', 'up', 'jump']
Qvalues:  [5.56, 5.37, 5.37, 5.31, 5.31, 5.22, 5.13, 5.13, 5.1, 5.01, 4.91, 4.82, 4.57, 4.44, 3.26, -4.31]
  Action:  shake sack
Reward:  0, Score 55, Done False
\\ \\ STATE 74 \\ the contents of the brown sack spills to the ground. you are carrying : a brown sack kitchen you are in the kitchen of the white house. a table seems to have been used recently for the preparation of food. a passage leads to the west and a dark staircase can be seen leading upward. a dark chimney leads down and to the east is a small window which is open. a hot pepper sandwich is here. there is a clove of garlic here. sitting on the kitchen table is : a glass bottle the glass bottle contains : a quantity of water 
Actions:  ['put sack on table', 'west', 'put down sack', 'take food', 'east', 'take garlic', 'open bottle', 'take bottle', 'close sack', 'take all', 'close window', 'up', 'jump']
Qvalues:  [5.61, 5.48, 5.42, 5.32, 5.28, 5.14, 5.1, 5.01, 4.85, 4.49, 4.37, 3.37, -4.37]
  Action:  close sack
Reward:  0, Score 55, Done False
\\ \\ STATE 75 \\ closed. you are carrying : a brown sack kitchen you are in the kitchen of the white house. a table seems to have been used recently for the preparation of food. a passage leads to the west and a dark staircase can be seen leading upward. a dark chimney leads down and to the east is a small window which is open. a hot pepper sandwich is here. there is a clove of garlic here. sitting on the kitchen table is : a glass bottle the glass bottle contains : a quantity of water 
Actions:  ['put sack on table', 'west', 'put down sack', 'take food', 'east', 'take garlic', 'open bottle', 'open sack', 'take bottle', 'take all', 'close window', 'up', 'jump']
Qvalues:  [5.67, 5.52, 5.47, 5.4, 5.32, 5.26, 5.2, 5.01, 4.97, 4.7, 4.48, 2.85, -4.86]
  Action:  put down sack
Reward:  0, Score 55, Done False
\\ \\ STATE 76 \\ dropped. you are empty handed. kitchen you are in the kitchen of the white house. a table seems to have been used recently for the preparation of food. a passage leads to the west and a dark staircase can be seen leading upward. a dark chimney leads down and to the east is a small window which is open. there is a brown sack here. a hot pepper sandwich is here. there is a clove of garlic here. sitting on the kitchen table is : a glass bottle the glass bottle contains : a quantity of water 
Actions:  ['east', 'west', 'open sack', 'open bottle', 'take sack', 'take food', 'take garlic', 'take all', 'close window', 'take bottle', 'up', 'jump']
Qvalues:  [6.33, 5.31, 5.27, 5.17, 4.77, 4.69, 4.63, 4.58, 4.57, 4.47, 1.86, -7.42]
  Action:  east
Reward:  0, Score 55, Done False
\\ \\ STATE 77 \\ behind house you are empty handed. behind house you are behind the white house. a path leads into the forest to the east. in one corner of the house there is a small window which is open. 
Actions:  ['close window', 'north', 'east', 'south', 'west']
Qvalues:  [6.43, 5.68, 5.43, 5.25, 5.15]
  Action:  close window
Reward:  0, Score 55, Done False
\\ \\ STATE 78 \\ the window closes ( more easily than it opened ). you are empty handed. behind house you are behind the white house. a path leads into the forest to the east. in one corner of the house there is a small window which is slightly ajar. 
Actions:  ['open window', 'north', 'south', 'east']
Qvalues:  [7.54, 6.26, 5.95, 5.91]
  Action:  north
Reward:  0, Score 55, Done False
\\ \\ STATE 79 \\ north of house you are facing the north side of a white house. there is no door here, and all the windows are boarded up. to the north a narrow path winds through the trees. you are empty handed. north of house you are facing the north side of a white house. there is no door here, and all the windows are boarded up. to the north a narrow path winds through the trees. 
Actions:  ['east', 'north', 'west']
Qvalues:  [7.13, 6.52, 5.61]
  Action:  east
Reward:  0, Score 55, Done False
\\ \\ STATE 80 \\ behind house you are empty handed. behind house you are behind the white house. a path leads into the forest to the east. in one corner of the house there is a small window which is slightly ajar. 
Actions:  ['open window', 'north', 'south', 'east']
Qvalues:  [6.88, 5.47, 5.28, 5.27]
  Action:  open window
Reward:  0, Score 55, Done False
\\ \\ STATE 81 \\ with great effort, you open the window far enough to allow entry. you are empty handed. behind house you are behind the white house. a path leads into the forest to the east. in one corner of the house there is a small window which is open. 
Actions:  ['west', 'close window', 'north', 'east', 'south']
Qvalues:  [7.23, 6.18, 5.72, 5.56, 5.36]
  Action:  west
Reward:  0, Score 55, Done False
\\ \\ STATE 82 \\ kitchen there is a brown sack here. a hot pepper sandwich is here. there is a clove of garlic here. sitting on the kitchen table is : a glass bottle the glass bottle contains : a quantity of water you are empty handed. kitchen you are in the kitchen of the white house. a table seems to have been used recently for the preparation of food. a passage leads to the west and a dark staircase can be seen leading upward. a dark chimney leads down and to the east is a small window which is open. there is a brown sack here. a hot pepper sandwich is here. there is a clove of garlic here. sitting on the kitchen table is : a glass bottle the glass bottle contains : a quantity of water 
Actions:  ['west', 'east', 'open sack', 'open bottle', 'close window', 'take garlic', 'take food', 'take sack', 'take bottle', 'take all', 'up', 'jump']
Qvalues:  [8.05, 7.66, 6.21, 5.98, 5.91, 5.52, 5.44, 5.4, 4.96, 4.93, 2.11, -6.64]
  Action:  east
Reward:  0, Score 55, Done False
\\ \\ STATE 83 \\ behind house you are empty handed. behind house you are behind the white house. a path leads into the forest to the east. in one corner of the house there is a small window which is open. 
Actions:  ['close window', 'north', 'east', 'south', 'west']
Qvalues:  [6.52, 6.0, 5.7, 5.69, 5.22]
  Action:  west
Reward:  0, Score 55, Done False
\\ \\ STATE 84 \\ kitchen there is a brown sack here. a hot pepper sandwich is here. there is a clove of garlic here. sitting on the kitchen table is : a glass bottle the glass bottle contains : a quantity of water you are empty handed. kitchen you are in the kitchen of the white house. a table seems to have been used recently for the preparation of food. a passage leads to the west and a dark staircase can be seen leading upward. a dark chimney leads down and to the east is a small window which is open. there is a brown sack here. a hot pepper sandwich is here. there is a clove of garlic here. sitting on the kitchen table is : a glass bottle the glass bottle contains : a quantity of water 
Actions:  ['east', 'west', 'open sack', 'open bottle', 'close window', 'take garlic', 'take food', 'take sack', 'take bottle', 'take all', 'up', 'jump']
Qvalues:  [7.35, 7.25, 6.16, 6.01, 5.81, 5.52, 5.44, 5.41, 5.16, 5.06, 2.48, -6.1]
  Action:  take food
Reward:  0, Score 55, Done False
\\ \\ STATE 85 \\ taken. you are carrying : a lunch kitchen you are in the kitchen of the white house. a table seems to have been used recently for the preparation of food. a passage leads to the west and a dark staircase can be seen leading upward. a dark chimney leads down and to the east is a small window which is open. there is a brown sack here. there is a clove of garlic here. sitting on the kitchen table is : a glass bottle the glass bottle contains : a quantity of water 
Actions:  ['put down food', 'put food on table', 'east', 'take garlic', 'take sack', 'west', 'open bottle', 'open sack', 'take bottle', 'take all', 'close window', 'up', 'jump']
Qvalues:  [5.65, 5.57, 5.53, 5.48, 5.47, 5.43, 5.25, 5.19, 5.13, 4.96, 4.56, 3.21, -4.72]
  Action:  take garlic
Reward:  0, Score 55, Done False
\\ \\ STATE 86 \\ taken. you are carrying : a clove of garlic a lunch kitchen you are in the kitchen of the white house. a table seems to have been used recently for the preparation of food. a passage leads to the west and a dark staircase can be seen leading upward. a dark chimney leads down and to the east is a small window which is open. there is a brown sack here. sitting on the kitchen table is : a glass bottle the glass bottle contains : a quantity of water 
Actions:  ['put all on table', 'put down all', 'west', 'put food on table', 'take sack', 'put down food', 'put garlic on table', 'east', 'open bottle', 'put down garlic', 'take bottle', 'open sack', 'take all', 'close window', 'up', 'jump']
Qvalues:  [5.93, 5.82, 5.62, 5.59, 5.57, 5.56, 5.51, 5.47, 5.45, 5.42, 5.32, 5.28, 4.87, 4.68, 3.83, -3.17]
  Action:  put all on table
Reward:  0, Score 55, Done False
\\ \\ STATE 87 \\ clove of garlic : done. lunch : done. you are empty handed. kitchen you are in the kitchen of the white house. a table seems to have been used recently for the preparation of food. a passage leads to the west and a dark staircase can be seen leading upward. a dark chimney leads down and to the east is a small window which is open. there is a brown sack here. sitting on the kitchen table is : a lunch a clove of garlic a glass bottle the glass bottle contains : a quantity of water 
Actions:  ['east', 'west', 'open sack', 'open bottle', 'close window', 'take garlic', 'take food', 'take sack', 'take all', 'take bottle', 'take all from table', 'up', 'jump']
Qvalues:  [6.17, 5.34, 5.27, 5.11, 4.66, 4.62, 4.55, 4.55, 4.39, 4.28, 3.72, 2.16, -6.62]
  Action:  take sack
Reward:  0, Score 55, Done False
\\ \\ STATE 88 \\ taken. you are carrying : a brown sack kitchen you are in the kitchen of the white house. a table seems to have been used recently for the preparation of food. a passage leads to the west and a dark staircase can be seen leading upward. a dark chimney leads down and to the east is a small window which is open. sitting on the kitchen table is : a lunch a clove of garlic a glass bottle the glass bottle contains : a quantity of water 
Actions:  ['put sack on table', 'put down sack', 'west', 'take food', 'east', 'take garlic', 'open bottle', 'take bottle', 'open sack', 'take all', 'close window', 'up', 'jump']
Qvalues:  [5.72, 5.61, 5.46, 5.46, 5.37, 5.31, 5.26, 5.25, 5.13, 4.8, 4.53, 3.63, -3.52]
  Action:  take food
Reward:  0, Score 55, Done False
\\of the white house. a table seems to have been used recently for the preparation of food. a passage leads to the west and a dark staircase can be seen leading upward. a dark chimney leads down and to the east is a small window which is open. sitting on the kitchen table is : a clove of garlic a glass bottle the glass bottle contains : a quantity of water 
Actions:  ['put all on table', 'put down all', 'put sack on table', 'put down sack', 'west', 'put down food', 'put food on table', 'take garlic', 'east', 'open bottle', 'open sack', 'take bottle', 'take all', 'close window', 'up', 'jump']
Qvalues:  [5.32, 5.29, 5.25, 5.12, 5.09, 5.06, 5.05, 4.91, 4.9, 4.79, 4.66, 4.56, 4.33, 4.07, 3.08, -4.21]
  Action:  take bottle
Reward:  0, Score 55, Done False
\\ \\ STATE 90 \\ taken. you are carrying : a glass bottle the glass bottle contains : a quantity of water a lunch a brown sack kitchen you are in the kitchen of the white house. a table seems to have been used recently for the preparation of food. a passage leads to the west and a dark staircase can be seen leading upward. a dark chimney leads down and to the east is a small window which is open. sitting on the kitchen table is : a clove of garlic 
Actions:  ['put all on table', 'put down bottle', 'put down all', 'put bottle on table', 'east', 'put sack on table', 'west', 'put down sack', 'put down food', 'put food on table', 'open bottle', 'put water on table', 'open sack', 'take garlic', 'up', 'close window', 'jump']
Qvalues:  [5.07, 5.04, 5.02, 5.0, 4.88, 4.87, 4.82, 4.74, 4.7, 4.62, 4.59, 4.55, 4.35, 4.34, 3.94, 3.75, -2.22]
  Action:  put down bottle
Reward:  0, Score 55, Done False
\\ \\ STATE 91 \\ dropped. you are carrying : a lunch a brown sack kitchen you are in the kitchen of the white house. a table seems to have been used recently for the preparation of food. a passage leads to the west and a dark staircase can be seen leading upward. a dark chimney leads down and to the east is a small window which is open. there is a glass bottle here. the glass bottle contains : a quantity of water sitting on the kitchen table is : a clove of garlic 
Actions:  ['put all on table', 'put sack on table', 'put down sack', 'put down all', 'west', 'put down food', 'put food on table', 'east', 'take garlic', 'open bottle', 'open sack', 'take bottle', 'take all', 'close window', 'up', 'jump']
Qvalues:  [5.43, 5.42, 5.32, 5.3, 5.3, 5.25, 5.17, 5.02, 4.94, 4.83, 4.66, 4.62, 4.28, 4.06, 3.09, -4.1]
  Action:  put down sack
Reward:  0, Score 55, Done False
\\ \\ STATE 92 \\ dropped. you are carrying : a lunch kitchen you are in the kitchen of the white house. a table seems to have been used recently for the preparation of food. a passage leads to the west and a dark staircase can be seen leading upward. a dark chimney leads down and to the east is a small window which is open. there is a brown sack here. there is a glass bottle here. the glass bottle contains : a quantity of water sitting on the kitchen table is : a clove of garlic 
Actions:  ['east', 'put down food', 'west', 'put food on table', 'take sack', 'take garlic', 'open bottle', 'open sack', 'take bottle', 'take all', 'close window', 'up', 'jump']
Qvalues:  [5.35, 5.25, 5.23, 5.12, 4.97, 4.97, 4.73, 4.73, 4.72, 4.69, 4.11, 3.21, -4.36]
  Action:  east
Reward:  0, Score 55, Done False
\\ \\ STATE 93 \\ behind house you are carrying : a lunch behind house you are behind the white house. a path leads into the forest to the east. in one corner of the house there is a small window which is open. 
Actions:  ['put down lunch', 'north', 'close window', 'south', 'east', 'west']
Qvalues:  [5.7, 4.52, 4.49, 4.3, 4.28, 4.03]
  Action:  close window
Reward:  0, Score 55, Done False
\\ \\ STATE 94 \\ the window closes ( more easily than it opened ). you are carrying : a lunch behind house you are behind the white house. a path leads into the forest to the east. in one corner of the house there is a small window which is slightly ajar. 
Actions:  ['put down lunch', 'open window', 'south', 'east', 'north']
Qvalues:  [5.45, 4.72, 4.17, 4.1, 3.96]
  Action:  south
Reward:  0, Score 55, Done False
\\ \\ STATE 95 \\ south of house you are carrying : a lunch south of house you are facing the south side of a white house. there is no door here, and all the windows are boarded. 
Actions:  ['put down lunch', 'east', 'south', 'west']
Qvalues:  [4.75, 4.42, 4.13, 3.16]
  Action:  east
Reward:  0, Score 55, Done False
\\ \\ STATE 96 \\ behind house you are carrying : a lunch behind house you are behind the white house. a path leads into the forest to the east. in one corner of the house there is a small window which is slightly ajar. 
Actions:  ['put down lunch', 'open window', 'south', 'north', 'east']
Qvalues:  [5.48, 4.63, 4.06, 4.03, 3.82]
  Action:  open window
Reward:  0, Score 55, Done False
\\ \\ STATE 97 \\ with great effort, you open the window far enough to allow entry. you are carrying : a lunch behind house you are behind the white house. a path leads into the forest to the east. in one corner of the house there is a small window which is open. 
Actions:  ['put down lunch', 'west', 'south', 'north', 'east', 'close window']
Qvalues:  [5.21, 4.88, 4.3, 4.26, 4.1, 3.92]
  Action:  west
Reward:  0, Score 55, Done False
\\ \\ STATE 98 \\ kitchen there is a brown sack here. there is a glass bottle here. the glass bottle contains : a quantity of water sitting on the kitchen table is : a clove of garlic you are carrying : a lunch kitchen you are in the kitchen of the white house. a table seems to have been used recently for the preparation of food. a passage leads to the west and a dark staircase can be seen leading upward. a dark chimney leads down and to the east is a small window which is open. there is a brown sack here. there is a glass bottle here. the glass bottle contains : a quantity of water sitting on the kitchen table is : a clove of garlic 
Actions:  ['put down food', 'put food on table', 'west', 'take sack', 'take garlic', 'east', 'open bottle', 'open sack', 'take bottle', 'take all', 'close window', 'up', 'jump']
Qvalues:  [4.8, 4.68, 4.6, 4.58, 4.46, 4.44, 4.44, 4.38, 4.36, 4.0, 3.7, 3.02, -4.39]
  Action:  take garlic
Reward:  0, Score 55, Done False
\\ \\ STATE 99 \\ taken. you are carrying : a clove of garlic a lunch kitchen you are in the kitchen of the white house. a table seems to have been used recently for the preparation of food. a passage leads to the west and a dark staircase can be seen leading upward. a dark chimney leads down and to the east is a small window which is open. there is a brown sack here. there is a glass bottle here. the glass bottle contains : a quantity of water 
Actions:  ['put down all', 'put all on table', 'put down food', 'west', 'put food on table', 'put garlic on table', 'take sack', 'put down garlic', 'open bottle', 'east', 'open sack', 'take bottle', 'take all', 'close window', 'up', 'jump']
Qvalues:  [5.33, 5.29, 5.05, 5.04, 4.98, 4.92, 4.91, 4.81, 4.79, 4.65, 4.61, 4.54, 4.26, 3.94, 3.29, -3.82]
  Action:  put down food
Reward:  0, Score 55, Done True
\\
\\ STATE 100 \\ dropped. you are carrying : a clove of garlic kitchen you are in the kitchen of the white house. a table seems to have been used recently for the preparation of food. a passage leads to the west and a dark staircase can be seen leading upward. a dark chimney leads down and to the east is a small window which is open. a hot pepper sandwich is here. there is a brown sack here. there is a glass bottle here. the glass bottle contains : a quantity of water 
Actions:  ['west', 'put garlic on table', 'take sack', 'open bottle', 'take food', 'put down garlic', 'open sack', 'take bottle', 'east', 'take all', 'close window', 'up', 'jump']
Qvalues:  [5.1, 4.79, 4.73, 4.68, 4.67, 4.65, 4.57, 4.32, 4.3, 3.95, 3.91, 2.32, -5.11]
  Action:  reset
Reward:  0, Score 0, Done False

\subsubsection{DRRN (for comparison)} \label{appendix:drrn}

STATE 0 \\  Copyright (c) 1981, 1982, 1983 Infocom, Inc. All rights reserved. ZORK is a registered trademark of Infocom, Inc. Revision 88 / Serial number 840726  West of House You are standing in an open field west of a white house, with a boarded front door. There is a small mailbox here.  You are empty handed.  West of House You are standing in an open field west of a white house, with a boarded front door. There is a small mailbox here. Actions:  ['north', 'south', 'west', 'open mailbox']
Qvalues:  [9.25, 8.75, 8.4, 8.11]
   Action:  north
Reward:  0, Score 0, Done False
\\ \\ STATE 1 \\  north of house you are facing the north side of a white house. there is no door here, and all the windows are boarded up. to the north a narrow path winds through the trees.  you are empty handed.  north of house you are facing the north side of a white house. there is no door here, and all the windows are boarded up. to the north a narrow path winds through the trees. Actions:  ['north', 'east', 'west']
Qvalues:  [10.23, 10.22, 8.8]
   Action:  north
Reward:  0, Score 0, Done False
\\ \\ STATE 2 \\  forest path this is a path winding through a dimly lit forest. the path heads north south here. one particularly large tree with some low branches stands at the edge of the path.  you are empty handed.  forest path this is a path winding through a dimly lit forest. the path heads north south here. one particularly large tree with some low branches stands at the edge of the path. Actions:  ['up', 'go around forest', 'south', 'east', 'north', 'west']
Qvalues:  [11.3, 9.84, 9.53, 8.64, 7.8, 5.7]
   Action:  up
Reward:  0, Score 0, Done False
\\ \\ STATE 3 \\  up a tree you are about 10 feet above the ground nestled among some large branches. the nearest branch above you is above your reach. beside you on the branch is a small bird's nest. in the bird's nest is a large egg encrusted with precious jewels, apparently scavenged by a childless songbird. the egg is covered with fine gold inlay, and ornamented in lapis lazuli and mother of pearl. unlike most eggs, this one is hinged and closed with a delicate looking clasp. the egg appears extremely fragile.  you are empty handed.  up a tree you are about 10 feet above the ground nestled among some large branches. the nearest branch above you is above your reach. beside you on the branch is a small bird's nest. in the bird's nest is a large egg encrusted with precious jewels, apparently scavenged by a childless songbird. the egg is covered with fine gold inlay, and ornamented in lapis lazuli and mother of pearl. unlike most eggs, this one is hinged and closed with a delicate looking clasp. the egg appears extremely fragile. Actions:  ['take egg', 'take on egg', 'take nest', 'close nest', 'down']
Qvalues:  [12.46, 10.87, 10.22, 9.65, 5.91]
   Action:  take egg
Reward:  5, Score 5, Done False
\\ \\ STATE 4 \\  taken.  you are carrying:   a jewel encrusted egg you hear in the distance the chirping of a song bird.  up a tree you are about 10 feet above the ground nestled among some large branches. the nearest branch above you is above your reach. beside you on the branch is a small bird's nest. you hear in the distance the chirping of a song bird. Actions:  ['take on egg', 'put egg in nest', 'down', 'take nest', 'put down egg', 'close nest', 'throw egg at nest']
Qvalues:  [8.24, 8.11, 7.77, 7.37, 7.34, 7.27, 6.01]
   Action:  down
Reward:  0, Score 5, Done False
\\ \\ STATE 5 \\  forest path  you are carrying:   a jewel encrusted egg  forest path this is a path winding through a dimly lit forest. the path heads north south here. one particularly large tree with some low branches stands at the edge of the path. you hear in the distance the chirping of a song bird. Actions:  ['south', 'take on egg', 'up', 'open egg with all', 'go around forest', 'put down egg', 'east', 'north', 'throw egg at tree', 'west']
Qvalues:  [9.33, 9.13, 9.02, 8.27, 8.21, 7.74, 7.41, 6.32, 4.05, 3.81]
   Action:  south
Reward:  0, Score 5, Done False
\\ \\ STATE 6 \\  north of house  you are carrying:   a jewel encrusted egg  north of house you are facing the north side of a white house. there is no door here, and all the windows are boarded up. to the north a narrow path winds through the trees. Actions:  ['east', 'take on egg', 'put down egg', 'north', 'open egg with all', 'west', 'throw egg at path']
Qvalues:  [10.13, 8.82, 7.17, 6.75, 6.43, 4.81, 3.01]
   Action:  east
Reward:  0, Score 5, Done False
\\ \\ STATE 7 \\  behind house you are behind the white house. a path leads into the forest to the east. in one corner of the house there is a small window which is slightly ajar.  you are carrying:   a jewel encrusted egg  behind house you are behind the white house. a path leads into the forest to the east. in one corner of the house there is a small window which is slightly ajar. Actions:  ['take on egg', 'open window', 'put down egg', 'open egg with all', 'north', 'east', 'south', 'throw egg at path']
Qvalues:  [11.34, 10.97, 10.23, 10.2, 8.65, 8.62, 8.47, 5.23]
   Action:  put down egg
Reward:  0, Score 5, Done False
\\ \\ STATE 8 \\  dropped.  you are empty handed.  behind house you are behind the white house. a path leads into the forest to the east. in one corner of the house there is a small window which is slightly ajar. there is a jewel encrusted egg here. Actions:  ['take egg', 'take on egg', 'open window', 'north', 'east', 'south']
Qvalues:  [11.04, 10.84, 9.91, 9.62, 9.43, 9.09]
   Action:  north
Reward:  0, Score 5, Done False
\\ \\ STATE 9 \\  north of house  you are empty handed.  north of house you are facing the north side of a white house. there is no door here, and all the windows are boarded up. to the north a narrow path winds through the trees. Actions:  ['east', 'north', 'west']
Qvalues:  [10.09, 8.65, 6.82]
   Action:  east
Reward:  0, Score 5, Done False
\\ \\ STATE 10 \\  behind house there is a jewel encrusted egg here.  you are empty handed.  behind house you are behind the white house. a path leads into the forest to the east. in one corner of the house there is a small window which is slightly ajar. there is a jewel encrusted egg here. Actions:  ['take egg', 'take on egg', 'north', 'open window', 'south', 'east']
Qvalues:  [10.77, 10.26, 10.26, 10.17, 10.01, 9.07]
   Action:  north
Reward:  0, Score 5, Done False
\\ \\ STATE 11 \\  north of house  you are empty handed.  north of house you are facing the north side of a white house. there is no door here, and all the windows are boarded up. to the north a narrow path winds through the trees. Actions:  ['east', 'north', 'west']
Qvalues:  [10.04, 8.6, 6.73]
   Action:  north
Reward:  0, Score 5, Done False
\\ \\ STATE 12 \\  forest path  you are empty handed.  forest path this is a path winding through a dimly lit forest. the path heads north south here. one particularly large tree with some low branches stands at the edge of the path. Actions:  ['up', 'south', 'go around forest', 'east', 'north', 'west']
Qvalues:  [9.02, 8.89, 8.28, 7.72, 7.19, 4.43]
   Action:  south
Reward:  0, Score 5, Done False
\\ \\ STATE 13 \\  north of house  you are empty handed.  north of house you are facing the north side of a white house. there is no door here, and all the windows are boarded up. to the north a narrow path winds through the trees. Actions:  ['east', 'north', 'west']
Qvalues:  [10.03, 8.58, 6.68]
   Action:  east
Reward:  0, Score 5, Done False
\\ \\ STATE 14 \\  behind house there is a jewel encrusted egg here.  you are empty handed.  behind house you are behind the white house. a path leads into the forest to the east. in one corner of the house there is a small window which is slightly ajar. there is a jewel encrusted egg here. Actions:  ['take egg', 'north', 'take on egg', 'open window', 'south', 'east']
Qvalues:  [10.64, 10.2, 10.18, 10.07, 9.98, 9.04]
   Action:  north
Reward:  0, Score 5, Done False
\\ \\ STATE 15 \\  north of house  you are empty handed.  north of house you are facing the north side of a white house. there is no door here, and all the windows are boarded up. to the north a narrow path winds through the trees. Actions:  ['east', 'north', 'west']
Qvalues:  [9.99, 8.53, 6.58]
   Action:  east
Reward:  0, Score 5, Done False
\\ \\ STATE 16 \\  behind house there is a jewel encrusted egg here.  you are empty handed.  behind house you are behind the white house. a path leads into the forest to the east. in one corner of the house there is a small window which is slightly ajar. there is a jewel encrusted egg here. Actions:  ['take egg', 'north', 'take on egg', 'open window', 'south', 'east']
Qvalues:  [10.54, 10.14, 10.09, 9.99, 9.93, 8.99]
   Action:  open window
Reward:  0, Score 5, Done False
\\ \\ STATE 17 \\  with great effort, you open the window far enough to allow entry.  you are empty handed.  behind house you are behind the white house. a path leads into the forest to the east. in one corner of the house there is a small window which is open. there is a jewel encrusted egg here. Actions:  ['take egg', 'take on egg', 'west', 'close window', 'east', 'south', 'north']
Qvalues:  [11.46, 9.71, 9.65, 8.11, 6.49, 6.19, 5.6]
   Action:  take egg
Reward:  0, Score 5, Done False
\\ \\ STATE 18 \\  taken.  you are carrying:   a jewel encrusted egg  behind house you are behind the white house. a path leads into the forest to the east. in one corner of the house there is a small window which is open. Actions:  ['take on egg', 'west', 'north', 'open egg with all', 'put down egg', 'close window', 'south', 'east', 'throw egg at path']
Qvalues:  [10.51, 10.42, 10.08, 9.93, 9.81, 9.6, 9.4, 9.12, 6.64]
   Action:  north
Reward:  0, Score 5, Done False
\\ \\ STATE 19 \\  north of house  you are carrying:   a jewel encrusted egg  north of house you are facing the north side of a white house. there is no door here, and all the windows are boarded up. to the north a narrow path winds through the trees. Actions:  ['east', 'take on egg', 'put down egg', 'north', 'open egg with all', 'west', 'throw egg at path']
Qvalues:  [9.59, 8.34, 6.73, 6.21, 5.94, 4.12, 2.61]
   Action:  take on egg
Reward:  0, Score 5, Done False
\\ \\ STATE 20 \\  there is a noticeable crunch from beneath you, and inspection reveals that the egg is lying open, badly damaged. there is a golden clockwork canary nestled in the egg. it seems to have recently had a bad experience. the mountings for its jewel like eyes are empty, and its silver beak is crumpled. through a cracked crystal window below its left wing you can see the remains of intricate machinery. it is not clear what result winding it would have, as the mainspring seems sprung.  you are carrying:   a broken jewel encrusted egg   the broken jewel encrusted egg contains:     a broken clockwork canary  north of house you are facing the north side of a white house. there is no door here, and all the windows are boarded up. to the north a narrow path winds through the trees. Actions:  ['take canary', 'shake egg', 'close egg', 'north', 'east', 'put down canary', 'put down egg', 'west']
Qvalues:  [10.64, 10.34, 10.01, 9.8, 9.8, 9.68, 9.67, 8.01]
   Action:  north
Reward:  0, Score 5, Done False
\\ \\ STATE 21 \\  forest path  you are carrying:   a broken jewel encrusted egg   the broken jewel encrusted egg contains:     a broken clockwork canary  forest path this is a path winding through a dimly lit forest. the path heads north south here. one particularly large tree with some low branches stands at the edge of the path. you hear in the distance the chirping of a song bird. Actions:  ['south', 'take canary', 'up', 'shake egg', 'put down canary', 'east', 'close egg', 'go around forest', 'north', 'put down egg', 'west']
Qvalues:  [11.15, 9.93, 9.35, 9.24, 8.89, 8.88, 8.79, 8.74, 8.22, 8.19, 5.5]
   Action:  south
Reward:  0, Score 5, Done False
\\ \\ STATE 22 \\  north of house  you are carrying:   a broken jewel encrusted egg   the broken jewel encrusted egg contains:     a broken clockwork canary  north of house you are facing the north side of a white house. there is no door here, and all the windows are boarded up. to the north a narrow path winds through the trees. Actions:  ['east', 'shake egg', 'take canary', 'close egg', 'north', 'put down canary', 'put down egg', 'west']
Qvalues:  [11.93, 10.06, 9.78, 9.49, 8.68, 8.53, 8.35, 6.3]
   Action:  east
Reward:  0, Score 5, Done False
\\ \\ STATE 23 \\  behind house  you are carrying:   a broken jewel encrusted egg   the broken jewel encrusted egg contains:     a broken clockwork canary  behind house you are behind the white house. a path leads into the forest to the east. in one corner of the house there is a small window which is open. Actions:  ['close window', 'shake egg', 'close egg', 'take canary', 'east', 'south', 'north', 'put down egg', 'put down canary', 'west']
Qvalues:  [11.21, 10.7, 10.61, 10.2, 10.12, 9.76, 9.66, 9.53, 8.86, 8.7]
   Action:  close egg
Reward:  0, Score 5, Done False
\\ \\ STATE 24 \\  closed.  you are carrying:   a broken jewel encrusted egg  behind house you are behind the white house. a path leads into the forest to the east. in one corner of the house there is a small window which is open. Actions:  ['close window', 'south', 'north', 'west', 'east', 'open egg', 'put down egg']
Qvalues:  [12.79, 12.65, 11.54, 11.37, 10.81, 10.31, 10.01]
   Action:  south
Reward:  0, Score 5, Done False
\\ \\ STATE 25 \\  south of house you are facing the south side of a white house. there is no door here, and all the windows are boarded.  you are carrying:   a broken jewel encrusted egg  south of house you are facing the south side of a white house. there is no door here, and all the windows are boarded. Actions:  ['east', 'south', 'open egg', 'put down egg', 'west']
Qvalues:  [14.39, 10.78, 10.45, 9.31, 8.39]
   Action:  east
Reward:  0, Score 5, Done False
\\ \\ STATE 26 \\  behind house  you are carrying:   a broken jewel encrusted egg  behind house you are behind the white house. a path leads into the forest to the east. in one corner of the house there is a small window which is open. Actions:  ['close window', 'south', 'east', 'open egg', 'north', 'west', 'put down egg']
Qvalues:  [13.61, 12.05, 11.88, 11.29, 11.11, 10.28, 10.27]
   Action:  close window
Reward:  0, Score 5, Done False
\\ \\ STATE 27 \\  the window closes (more easily than it opened).  you are carrying:   a broken jewel encrusted egg  behind house you are behind the white house. a path leads into the forest to the east. in one corner of the house there is a small window which is slightly ajar. Actions:  ['open window', 'north', 'east', 'south', 'open egg', 'put down egg']
Qvalues:  [15.18, 12.67, 12.56, 12.56, 12.07, 11.34]
   Action:  open window
Reward:  0, Score 5, Done False
\\ \\ STATE 28 \\  with great effort, you open the window far enough to allow entry.  you are carrying:   a broken jewel encrusted egg  behind house you are behind the white house. a path leads into the forest to the east. in one corner of the house there is a small window which is open. Actions:  ['west', 'close window', 'open egg', 'north', 'put down egg', 'east', 'south']
Qvalues:  [18.49, 15.61, 14.4, 13.72, 13.06, 12.87, 12.11]
   Action:  west
Reward:  10, Score 15, Done False
\\ \\ STATE 29 \\  kitchen you are in the kitchen of the white house. a table seems to have been used recently for the preparation of food. a passage leads to the west and a dark staircase can be seen leading upward. a dark chimney leads down and to the east is a small window which is open. on the table is an elongated brown sack, smelling of hot peppers. a bottle is sitting on the table. the glass bottle contains:   a quantity of water  you are carrying:   a broken jewel encrusted egg  kitchen you are in the kitchen of the white house. a table seems to have been used recently for the preparation of food. a passage leads to the west and a dark staircase can be seen leading upward. a dark chimney leads down and to the east is a small window which is open. on the table is an elongated brown sack, smelling of hot peppers. a bottle is sitting on the table. the glass bottle contains:   a quantity of water Actions:  ['east', 'west', 'close window', 'take bottle', 'take all', 'open bottle', 'open sack', 'take sack', 'open egg', 'up', 'put egg on table', 'put down egg', 'jump']
Qvalues:  [12.6, 12.33, 12.29, 11.66, 11.3, 11.3, 11.0, 10.48, 10.26, 10.12, 9.82, 9.81, 4.29]
   Action:  put down egg
Reward:  0, Score 15, Done False
\\ \\ STATE 30 \\  dropped.  you are empty handed.  kitchen you are in the kitchen of the white house. a table seems to have been used recently for the preparation of food. a passage leads to the west and a dark staircase can be seen leading upward. a dark chimney leads down and to the east is a small window which is open. there is a somewhat ruined egg here. on the table is an elongated brown sack, smelling of hot peppers. a bottle is sitting on the table. the glass bottle contains:   a quantity of water Actions:  ['take egg', 'east', 'open sack', 'west', 'take bottle', 'take sack', 'take all from table', 'close window', 'take all', 'open bottle', 'up', 'open egg', 'jump']
Qvalues:  [9.16, 9.1, 9.02, 8.95, 8.85, 8.65, 8.54, 8.49, 8.48, 8.42, 8.29, 8.02, 0.73]
   Action:  open sack
Reward:  0, Score 15, Done False
\\ \\ STATE 31 \\  opening the brown sack reveals a lunch, and a clove of garlic.  you are empty handed.  kitchen you are in the kitchen of the white house. a table seems to have been used recently for the preparation of food. a passage leads to the west and a dark staircase can be seen leading upward. a dark chimney leads down and to the east is a small window which is open. there is a somewhat ruined egg here. a bottle is sitting on the table. the glass bottle contains:   a quantity of water   a brown sack   the brown sack contains:     a lunch     a clove of garlic Actions:  ['take egg', 'take all from table', 'take all', 'take all from sack', 'west', 'take sack', 'close sack', 'take food', 'take bottle', 'take garlic', 'open egg', 'open bottle', 'close window', 'east', 'up', 'jump']
Qvalues:  [9.83, 9.13, 8.97, 8.85, 8.69, 8.59, 8.55, 8.51, 8.43, 8.31, 8.09, 7.93, 7.75, 7.02, 6.99, -1.14]
   Action:  close sack
Reward:  0, Score 15, Done False
\\ \\ STATE 32 \\  closed.  you are empty handed.  kitchen you are in the kitchen of the white house. a table seems to have been used recently for the preparation of food. a passage leads to the west and a dark staircase can be seen leading upward. a dark chimney leads down and to the east is a small window which is open. there is a somewhat ruined egg here. a bottle is sitting on the table. the glass bottle contains:   a quantity of water   a brown sack Actions:  ['take egg', 'take all', 'take all from table', 'take sack', 'open sack', 'take bottle', 'close window', 'open egg', 'open bottle', 'west', 'east', 'up', 'jump']
Qvalues:  [11.85, 11.45, 11.15, 11.06, 10.74, 10.47, 10.35, 10.14, 10.06, 9.89, 9.32, 9.24, 2.32]
   Action:  take bottle
Reward:  0, Score 15, Done False
\\ \\ STATE 33 \\  taken.  you are carrying:   a glass bottle   the glass bottle contains:     a quantity of water  kitchen you are in the kitchen of the white house. a table seems to have been used recently for the preparation of food. a passage leads to the west and a dark staircase can be seen leading upward. a dark chimney leads down and to the east is a small window which is open. there is a somewhat ruined egg here.   sitting on the kitchen table is:      a brown sack Actions:  ['take egg', 'take all', 'west', 'put water on table', 'open sack', 'take sack', 'put bottle on table', 'open egg', 'open bottle', 'close window', 'up', 'put down bottle', 'east', 'jump']
Qvalues:  [11.16, 10.75, 10.52, 9.92, 9.89, 9.88, 9.78, 9.66, 9.51, 9.4, 9.36, 9.12, 8.99, 1.01]
   Action:  close window
Reward:  0, Score 15, Done False
\\ \\ STATE 34 \\  the window closes (more easily than it opened).  you are carrying:   a glass bottle   the glass bottle contains:     a quantity of water  kitchen you are in the kitchen of the white house. a table seems to have been used recently for the preparation of food. a passage leads to the west and a dark staircase can be seen leading upward. a dark chimney leads down and to the east is a small window which is slightly ajar. there is a somewhat ruined egg here.   sitting on the kitchen table is:      a brown sack Actions:  ['take egg', 'west', 'take all', 'open egg', 'open sack', 'take sack', 'put water on table', 'put bottle on table', 'open bottle', 'open window', 'put down bottle', 'up', 'jump']
Qvalues:  [10.72, 9.98, 9.86, 9.09, 9.02, 8.96, 8.92, 8.83, 8.8, 8.65, 8.29, 7.96, -1.33]
   Action:  put water on table
Reward:  0, Score 15, Done False
\\ \\ STATE 35 \\  done.  you are carrying:   a glass bottle  kitchen you are in the kitchen of the white house. a table seems to have been used recently for the preparation of food. a passage leads to the west and a dark staircase can be seen leading upward. a dark chimney leads down and to the east is a small window which is slightly ajar. there is a somewhat ruined egg here.   sitting on the kitchen table is:      a quantity of water     a brown sack Actions:  ['take egg', 'take all', 'west', 'open window', 'open sack', 'take sack', 'open egg', 'open bottle', 'put bottle on table', 'put down bottle', 'up', 'jump']
Qvalues:  [11.52, 10.95, 10.61, 10.33, 10.11, 10.07, 10.03, 9.92, 9.8, 9.75, 8.45, 1.04]
   Action:  put bottle on table
Reward:  0, Score 15, Done False
\\ \\ STATE 36 \\  done.  you are empty handed.  kitchen you are in the kitchen of the white house. a table seems to have been used recently for the preparation of food. a passage leads to the west and a dark staircase can be seen leading upward. a dark chimney leads down and to the east is a small window which is slightly ajar. there is a somewhat ruined egg here.   sitting on the kitchen table is:      a glass bottle     a quantity of water     a brown sack Actions:  ['take egg', 'take all', 'open window', 'take all from table', 'west', 'take sack', 'take bottle', 'open sack', 'open bottle', 'open egg', 'up', 'jump']
Qvalues:  [11.36, 10.61, 10.58, 10.56, 10.46, 10.29, 10.28, 10.13, 9.7, 9.65, 8.73, 1.69]
   Action:  take sack
Reward:  0, Score 15, Done False
\\ \\ STATE 37 \\  taken.  you are carrying:   a brown sack  kitchen you are in the kitchen of the white house. a table seems to have been used recently for the preparation of food. a passage leads to the west and a dark staircase can be seen leading upward. a dark chimney leads down and to the east is a small window which is slightly ajar. there is a somewhat ruined egg here.   sitting on the kitchen table is:      a glass bottle     a quantity of water Actions:  ['west', 'take egg', 'take all', 'put sack on table', 'open window', 'open sack', 'take bottle', 'put down sack', 'open egg', 'open bottle', 'up', 'jump']
Qvalues:  [9.59, 9.28, 8.93, 8.76, 8.47, 8.47, 8.34, 8.32, 7.95, 7.83, 7.77, 0.83]
   Action:  west
Reward:  0, Score 15, Done False
\\ \\ STATE 38 \\  living room you are in the living room. there is a doorway to the east, a wooden door with strange gothic lettering to the west, which appears to be nailed shut, a trophy case, and a large oriental rug in the center of the room. above the trophy case hangs an elvish sword of great antiquity. a battery powered brass lantern is on the trophy case.  you are carrying:   a brown sack  living room you are in the living room. there is a doorway to the east, a wooden door with strange gothic lettering to the west, which appears to be nailed shut, a trophy case, and a large oriental rug in the center of the room. above the trophy case hangs an elvish sword of great antiquity. a battery powered brass lantern is on the trophy case. Actions:  ['take all', 'push rug', 'open sack', 'put down sack', 'open case', 'take sword', 'take lantern', 'turn on lantern', 'throw sack at lantern', 'east']
Qvalues:  [10.66, 10.2, 9.58, 9.11, 9.0, 8.97, 8.77, 8.68, 8.31, 7.35]
   Action:  push rug
Reward:  0, Score 15, Done False
\\ \\ STATE 39 \\  with a great effort, the rug is moved to one side of the room, revealing the dusty cover of a closed trap door.  you are carrying:   a brown sack  living room you are in the living room. there is a doorway to the east, a wooden door with strange gothic lettering to the west, which appears to be nailed shut, a trophy case, and a closed trap door at your feet. above the trophy case hangs an elvish sword of great antiquity. a battery powered brass lantern is on the trophy case. Actions:  ['take all', 'open trap', 'open sack', 'take sword', 'put down sack', 'take lantern', 'open case', 'turn on lantern', 'throw sack at lantern', 'east']
Qvalues:  [11.53, 11.1, 10.37, 10.0, 9.76, 9.69, 9.66, 9.61, 9.1, 8.05]
   Action:  open trap
Reward:  0, Score 15, Done False
\\ \\ STATE 40 \\  the door reluctantly opens to reveal a rickety staircase descending into darkness.  you are carrying:   a brown sack  living room you are in the living room. there is a doorway to the east, a wooden door with strange gothic lettering to the west, which appears to be nailed shut, a trophy case, and a rug lying beside an open trap door. above the trophy case hangs an elvish sword of great antiquity. a battery powered brass lantern is on the trophy case. Actions:  ['take all', 'take lantern', 'open sack', 'put down sack', 'take sword', 'open case', 'throw sack at lantern', 'turn on lantern', 'close trap', 'east', 'down']
Qvalues:  [12.63, 11.7, 11.21, 10.74, 10.64, 10.54, 10.47, 10.07, 9.7, 9.09, 7.73]
   Action:  take all
Reward:  0, Score 15, Done False
\\ \\ STATE 41 \\  trophy case: the trophy case is securely fastened to the wall. sword: taken. brass lantern: taken. carpet: the rug is extremely heavy and cannot be carried.  you are carrying:   a brass lantern   a sword   a brown sack  living room you are in the living room. there is a doorway to the east, a wooden door with strange gothic lettering to the west, which appears to be nailed shut, a trophy case, and a rug lying beside an open trap door. Actions:  ['open sack', 'put down all', 'down', 'open case', 'put down sack', 'throw sack at lantern', 'close trap', 'put down lantern', 'put down sword', 'turn on lantern', 'east']
Qvalues:  [14.22, 14.12, 13.88, 13.79, 13.66, 13.2, 13.14, 13.11, 13.01, 12.88, 12.33]
   Action:  put down all
Reward:  0, Score 15, Done False
\\ \\ STATE 42 \\  brass lantern: dropped. sword: dropped. brown sack: dropped.  you are empty handed.  living room you are in the living room. there is a doorway to the east, a wooden door with strange gothic lettering to the west, which appears to be nailed shut, a trophy case, and a rug lying beside an open trap door. there is a brown sack here. there is a sword here. there is a brass lantern (battery powered) here. Actions:  ['open sack', 'take lantern', 'take sack', 'take all', 'take sword', 'turn on lantern', 'open case', 'close trap', 'down', 'east']
Qvalues:  [11.7, 11.61, 11.57, 11.39, 11.37, 11.37, 10.79, 10.57, 10.13, 10.1]
   Action:  down
Reward:  25, Score 40, Done False
\\ \\ STATE 43 \\  you have moved into a dark place. the trap door crashes shut, and you hear someone barring it.  it is pitch black. you are likely to be eaten by a grue.  you are empty handed.  it is pitch black. you are likely to be eaten by a grue. Actions:  ['south', 'north']
Qvalues:  [-7.44, -9.98]
   Action:  south
Reward:  0, Score 40, Done False
\\ \\ STATE 44 \\  you have moved into a dark place. it is pitch black. you are likely to be eaten by a grue.  you are empty handed.  it is pitch black. you are likely to be eaten by a grue. Actions:  ['east', 'north', 'west']
Qvalues:  [1.4, -3.4, -8.58]
   Action:  east
Reward:  0, Score 40, Done False
\\ \\ STATE 45 \\  gallery this is an art gallery. most of the paintings have been stolen by vandals with exceptional taste. the vandals left through either the north or west exits. fortunately, there is still one chance for you to be a vandal, for on the far wall is a painting of unparalleled beauty.  you are empty handed.  gallery this is an art gallery. most of the paintings have been stolen by vandals with exceptional taste. the vandals left through either the north or west exits. fortunately, there is still one chance for you to be a vandal, for on the far wall is a painting of unparalleled beauty. Actions:  ['take art', 'north', 'west']
Qvalues:  [6.02, 1.29, 0.19]
   Action:  take art
Reward:  4, Score 44, Done False
\\ \\ STATE 46 \\  taken.  you are carrying:   a painting  gallery this is an art gallery. most of the paintings have been stolen by vandals with exceptional taste. the vandals left through either the north or west exits. Actions:  ['north', 'put down art', 'west']
Qvalues:  [2.41, 2.35, 2.15]
   Action:  put down art
Reward:  0, Score 44, Done False
\\ \\ STATE 47 \\  dropped.  you are empty handed.  gallery this is an art gallery. most of the paintings have been stolen by vandals with exceptional taste. the vandals left through either the north or west exits. a painting by a neglected genius is here. Actions:  ['take art', 'west', 'north']
Qvalues:  [2.91, 1.66, 1.11]
   Action:  take art
Reward:  0, Score 44, Done False
\\ \\ STATE 48 \\  taken.  you are carrying:   a painting  gallery this is an art gallery. most of the paintings have been stolen by vandals with exceptional taste. the vandals left through either the north or west exits. Actions:  ['north', 'put down art', 'west']
Qvalues:  [2.38, 2.36, 2.16]
   Action:  put down art
Reward:  0, Score 44, Done False
\\ \\ STATE 49 \\  dropped.  you are empty handed.  gallery this is an art gallery. most of the paintings have been stolen by vandals with exceptional taste. the vandals left through either the north or west exits. a painting by a neglected genius is here. Actions:  ['take art', 'west', 'north']
Qvalues:  [2.8, 1.52, 0.96]
   Action:  take art
Reward:  0, Score 44, Done False
\\ \\ STATE 50 \\  taken.  you are carrying:   a painting  gallery this is an art gallery. most of the paintings have been stolen by vandals with exceptional taste. the vandals left through either the north or west exits. Actions:  ['put down art', 'north', 'west']
Qvalues:  [2.34, 2.29, 2.1]
   Action:  north
Reward:  0, Score 44, Done False
\\ \\ STATE 51 \\  you have moved into a dark place. it is pitch black. you are likely to be eaten by a grue.  you are carrying:   a painting  it is pitch black. you are likely to be eaten by a grue. Actions:  ['south', 'put down painting', 'north']
Qvalues:  [1.71, -2.26, -3.91]
   Action:  south
Reward:  0, Score 44, Done False
\\ \\ STATE 52 \\  gallery  you are carrying:   a painting  gallery this is an art gallery. most of the paintings have been stolen by vandals with exceptional taste. the vandals left through either the north or west exits. Actions:  ['north', 'put down art', 'west']
Qvalues:  [2.0, 1.94, 1.62]
   Action:  put down art
Reward:  0, Score 44, Done False
\\ \\ STATE 53 \\  dropped.  you are empty handed.  gallery this is an art gallery. most of the paintings have been stolen by vandals with exceptional taste. the vandals left through either the north or west exits. a painting by a neglected genius is here. Actions:  ['take art', 'west', 'north']
Qvalues:  [2.44, 1.08, 0.47]
   Action:  take art
Reward:  0, Score 44, Done False
\\ \\ STATE 54 \\  taken.  you are carrying:   a painting  gallery this is an art gallery. most of the paintings have been stolen by vandals with exceptional taste. the vandals left through either the north or west exits. Actions:  ['put down art', 'north', 'west']
Qvalues:  [2.23, 1.96, 1.84]
   Action:  put down art
Reward:  0, Score 44, Done False
\\ \\ STATE 55 \\  dropped.  you are empty handed.  gallery this is an art gallery. most of the paintings have been stolen by vandals with exceptional taste. the vandals left through either the north or west exits. a painting by a neglected genius is here. Actions:  ['take art', 'west', 'north']
Qvalues:  [2.1, 0.73, 0.08]
   Action:  west
Reward:  0, Score 44, Done False
\\ \\ STATE 56 \\  you have moved into a dark place. it is pitch black. you are likely to be eaten by a grue.  you are empty handed.  it is pitch black. you are likely to be eaten by a grue. Actions:  ['east', 'north', 'west']
Qvalues:  [0.75, -4.31, -9.21]
   Action:  east
Reward:  0, Score 44, Done False
\\ \\ STATE 57 \\  gallery a painting by a neglected genius is here.  you are empty handed.  gallery this is an art gallery. most of the paintings have been stolen by vandals with exceptional taste. the vandals left through either the north or west exits. a painting by a neglected genius is here. Actions:  ['take art', 'west', 'north']
Qvalues:  [1.99, 0.53, -0.2]
   Action:  take art
Reward:  0, Score 44, Done False
\\ \\ STATE 58 \\  taken.  you are carrying:   a painting  gallery this is an art gallery. most of the paintings have been stolen by vandals with exceptional taste. the vandals left through either the north or west exits. Actions:  ['put down art', 'north', 'west']
Qvalues:  [1.93, 1.43, 1.35]
   Action:  put down art
Reward:  0, Score 44, Done False
\\ \\ STATE 59 \\  dropped.  you are empty handed.  gallery this is an art gallery. most of the paintings have been stolen by vandals with exceptional taste. the vandals left through either the north or west exits. a painting by a neglected genius is here. Actions:  ['take art', 'wait', 'west', 'north']
Qvalues:  [1.48, 0.25, 0.06, -0.58]
   Action:  wait
Reward:  0, Score 44, Done False
\\ \\ STATE 60 \\  time passes...  you are empty handed.  gallery this is an art gallery. most of the paintings have been stolen by vandals with exceptional taste. the vandals left through either the north or west exits. Actions:  ['north', 'west']
Qvalues:  [1.24, 0.15]
   Action:  west
Reward:  0, Score 44, Done False
\\ \\ STATE 61 \\  you have moved into a dark place. it is pitch black. you are likely to be eaten by a grue.  you are empty handed.  it is pitch black. you are likely to be eaten by a grue. Actions:  ['east', 'north', 'west']
Qvalues:  [0.87, -4.34, -9.29]
   Action:  east
Reward:  0, Score 44, Done False
\\ \\ STATE 62 \\  gallery  you are empty handed.  gallery this is an art gallery. most of the paintings have been stolen by vandals with exceptional taste. the vandals left through either the north or west exits. Actions:  ['west', 'north']
Qvalues:  [0.38, 0.23]
   Action:  west
Reward:  0, Score 44, Done False
\\ \\ STATE 63 \\  you have moved into a dark place. it is pitch black. you are likely to be eaten by a grue.  you are empty handed.  it is pitch black. you are likely to be eaten by a grue. Actions:  ['east', 'north', 'west']
Qvalues:  [0.92, -4.3, -9.3]
   Action:  east
Reward:  0, Score 44, Done False
\\ \\ STATE 64 \\  gallery  you are empty handed.  gallery this is an art gallery. most of the paintings have been stolen by vandals with exceptional taste. the vandals left through either the north or west exits. Actions:  ['west', 'north']
Qvalues:  [0.29, 0.16]
   Action:  north
Reward:  0, Score 44, Done False
\\ \\ STATE 65 \\  you have moved into a dark place. it is pitch black. you are likely to be eaten by a grue.  you are empty handed.  it is pitch black. you are likely to be eaten by a grue. Actions:  ['south', 'north']
Qvalues:  [0.59, -4.18]
   Action:  south
Reward:  0, Score 44, Done False
\\ \\ STATE 66 \\  gallery  you are empty handed.  gallery this is an art gallery. most of the paintings have been stolen by vandals with exceptional taste. the vandals left through either the north or west exits. Actions:  ['west', 'north']
Qvalues:  [0.29, 0.17]
   Action:  west
Reward:  0, Score 44, Done False
\\ \\ STATE 67 \\  you have moved into a dark place. it is pitch black. you are likely to be eaten by a grue.  you are empty handed.  it is pitch black. you are likely to be eaten by a grue. Actions:  ['east', 'north', 'west']
Qvalues:  [1.21, -4.03, -9.06]
   Action:  east
Reward:  0, Score 44, Done False
\\ \\ STATE 68 \\  gallery  you are empty handed.  gallery this is an art gallery. most of the paintings have been stolen by vandals with exceptional taste. the vandals left through either the north or west exits. Actions:  ['west', 'north']
Qvalues:  [0.38, 0.25]
   Action:  west
Reward:  0, Score 44, Done False
\\ \\ STATE 69 \\  you have moved into a dark place. it is pitch black. you are likely to be eaten by a grue.  you are empty handed.  it is pitch black. you are likely to be eaten by a grue. Actions:  ['east', 'north', 'west']
Qvalues:  [1.27, -3.95, -8.96]
   Action:  east
Reward:  0, Score 44, Done False
\\ \\ STATE 70 \\  gallery  you are empty handed.  gallery this is an art gallery. most of the paintings have been stolen by vandals with exceptional taste. the vandals left through either the north or west exits. Actions:  ['west', 'north']
Qvalues:  [0.46, 0.31]
   Action:  north
Reward:  0, Score 44, Done False
\\ \\ STATE 71 \\  you have moved into a dark place. it is pitch black. you are likely to be eaten by a grue.  you are empty handed.  it is pitch black. you are likely to be eaten by a grue. Actions:  ['south', 'north']
Qvalues:  [0.78, -3.93]
   Action:  south
Reward:  0, Score 44, Done False
\\ \\ STATE 72 \\  gallery  you are empty handed.  gallery this is an art gallery. most of the paintings have been stolen by vandals with exceptional taste. the vandals left through either the north or west exits. Actions:  ['west', 'north']
Qvalues:  [0.62, 0.45]
   Action:  west
Reward:  0, Score 44, Done False
\\ \\ STATE 73 \\  you have moved into a dark place. it is pitch black. you are likely to be eaten by a grue.  you are empty handed.  it is pitch black. you are likely to be eaten by a grue. Actions:  ['east', 'north', 'west']
Qvalues:  [1.32, -3.86, -8.82]
   Action:  east
Reward:  0, Score 44, Done False
\\ \\ STATE 74 \\  gallery  you are empty handed.  gallery this is an art gallery. most of the paintings have been stolen by vandals with exceptional taste. the vandals left through either the north or west exits. Actions:  ['west', 'north', 'wait']
Qvalues:  [0.75, 0.57, 0.39]
   Action:  north
Reward:  0, Score 44, Done False
\\ \\ STATE 75 \\  you have moved into a dark place. it is pitch black. you are likely to be eaten by a grue.  you are empty handed.  it is pitch black. you are likely to be eaten by a grue. Actions:  ['south', 'north']
Qvalues:  [0.87, -3.84]
   Action:  south
Reward:  0, Score 44, Done False
\\ \\ STATE 76 \\  gallery  you are empty handed.  gallery this is an art gallery. most of the paintings have been stolen by vandals with exceptional taste. the vandals left through either the north or west exits. Actions:  ['west', 'north']
Qvalues:  [0.93, 0.75]
   Action:  north
Reward:  0, Score 44, Done False
\\ \\ STATE 77 \\  you have moved into a dark place. it is pitch black. you are likely to be eaten by a grue.  you are empty handed.  it is pitch black. you are likely to be eaten by a grue. Actions:  ['south', 'north']
Qvalues:  [0.99, -3.68]
   Action:  south
Reward:  0, Score 44, Done False
\\ \\ STATE 78 \\  gallery  you are empty handed.  gallery this is an art gallery. most of the paintings have been stolen by vandals with exceptional taste. the vandals left through either the north or west exits. Actions:  ['west', 'north']
Qvalues:  [1.14, 0.97]
   Action:  west
Reward:  0, Score 44, Done False
\\ \\ STATE 79 \\  you have moved into a dark place. it is pitch black. you are likely to be eaten by a grue.  you are empty handed.  it is pitch black. you are likely to be eaten by a grue. Actions:  ['east', 'north', 'west']
Qvalues:  [1.51, -3.52, -8.51]
   Action:  east
Reward:  0, Score 44, Done False
\\ \\ STATE 80 \\  gallery  you are empty handed.  gallery this is an art gallery. most of the paintings have been stolen by vandals with exceptional taste. the vandals left through either the north or west exits. Actions:  ['west', 'north']
Qvalues:  [1.33, 1.18]
   Action:  west
Reward:  0, Score 44, Done False
\\ \\ STATE 81 \\  you have moved into a dark place. it is pitch black. you are likely to be eaten by a grue.  you are empty handed.  it is pitch black. you are likely to be eaten by a grue. Actions:  ['east', 'north', 'west']
Qvalues:  [1.66, -3.32, -8.33]
   Action:  east
Reward:  0, Score 44, Done False
\\ \\ STATE 82 \\  gallery  you are empty handed.  gallery this is an art gallery. most of the paintings have been stolen by vandals with exceptional taste. the vandals left through either the north or west exits. Actions:  ['west', 'north']
Qvalues:  [1.55, 1.4]
   Action:  north
Reward:  0, Score 44, Done False
\\ \\ STATE 83 \\  you have moved into a dark place. it is pitch black. you are likely to be eaten by a grue.  you are empty handed.  it is pitch black. you are likely to be eaten by a grue. Actions:  ['south', 'north']
Qvalues:  [1.41, -3.15]
   Action:  south
Reward:  0, Score 44, Done False
\\ \\ STATE 84 \\  gallery  you are empty handed.  gallery this is an art gallery. most of the paintings have been stolen by vandals with exceptional taste. the vandals left through either the north or west exits. Actions:  ['west', 'north']
Qvalues:  [1.65, 1.51]
   Action:  west
Reward:  0, Score 44, Done False
\\ \\ STATE 85 \\  you have moved into a dark place. it is pitch black. you are likely to be eaten by a grue.  you are empty handed.  it is pitch black. you are likely to be eaten by a grue. Actions:  ['east', 'north', 'west']
Qvalues:  [1.81, -3.09, -8.17]
   Action:  east
Reward:  0, Score 44, Done False
\\ \\ STATE 86 \\  gallery  you are empty handed.  gallery this is an art gallery. most of the paintings have been stolen by vandals with exceptional taste. the vandals left through either the north or west exits. Actions:  ['west', 'north']
Qvalues:  [1.66, 1.56]
   Action:  west
Reward:  0, Score 44, Done False
\\ \\ STATE 87 \\  you have moved into a dark place. it is pitch black. you are likely to be eaten by a grue.  you are empty handed.  it is pitch black. you are likely to be eaten by a grue. Actions:  ['east', 'north', 'west']
Qvalues:  [1.79, -3.06, -8.2]
   Action:  east
Reward:  0, Score 44, Done False
\\ \\ STATE 88 \\  gallery  you are empty handed.  gallery this is an art gallery. most of the paintings have been stolen by vandals with exceptional taste. the vandals left through either the north or west exits. Actions:  ['west', 'north']
Qvalues:  [1.6, 1.53]
   Action:  north
Reward:  0, Score 44, Done False
\\ \\ STATE 89 \\  you have moved into a dark place. it is pitch black. you are likely to be eaten by a grue.  you are empty handed.  it is pitch black. you are likely to be eaten by a grue. Actions:  ['south', 'north']
Qvalues:  [1.48, -3.07]
   Action:  south
Reward:  0, Score 44, Done False
\\ \\ STATE 90 \\  gallery  you are empty handed.  gallery this is an art gallery. most of the paintings have been stolen by vandals with exceptional taste. the vandals left through either the north or west exits. Actions:  ['west', 'north']
Qvalues:  [1.53, 1.5]
   Action:  north
Reward:  0, Score 44, Done False
\\ \\ STATE 91 \\  you have moved into a dark place. it is pitch black. you are likely to be eaten by a grue.  you are empty handed.  it is pitch black. you are likely to be eaten by a grue. Actions:  ['south', 'north']
Qvalues:  [1.46, -3.08]
   Action:  south
Reward:  0, Score 44, Done False
\\ \\ STATE 92 \\  gallery  you are empty handed.  gallery this is an art gallery. most of the paintings have been stolen by vandals with exceptional taste. the vandals left through either the north or west exits. Actions:  ['west', 'north']
Qvalues:  [1.45, 1.42]
   Action:  west
Reward:  0, Score 44, Done False
\\ \\ STATE 93 \\  you have moved into a dark place. it is pitch black. you are likely to be eaten by a grue.  you are empty handed.  it is pitch black. you are likely to be eaten by a grue. Actions:  ['east', 'north', 'west']
Qvalues:  [1.46, -3.16, -8.41]
   Action:  east
Reward:  0, Score 44, Done False
\\ \\ STATE 94 \\  gallery  you are empty handed.  gallery this is an art gallery. most of the paintings have been stolen by vandals with exceptional taste. the vandals left through either the north or west exits. Actions:  ['west', 'north']
Qvalues:  [1.38, 1.33]
   Action:  north
Reward:  0, Score 44, Done False
\\ \\ STATE 95 \\  you have moved into a dark place. it is pitch black. you are likely to be eaten by a grue.  you are empty handed.  it is pitch black. you are likely to be eaten by a grue. Actions:  ['south', 'north']
Qvalues:  [1.27, -3.26]
   Action:  south
Reward:  0, Score 44, Done False
\\ \\ STATE 96 \\  gallery  you are empty handed.  gallery this is an art gallery. most of the paintings have been stolen by vandals with exceptional taste. the vandals left through either the north or west exits. Actions:  ['west', 'north']
Qvalues:  [1.29, 1.22]
   Action:  west
Reward:  0, Score 44, Done False
\\ \\ STATE 97 \\  you have moved into a dark place. it is pitch black. you are likely to be eaten by a grue.  you are empty handed.  it is pitch black. you are likely to be eaten by a grue. Actions:  ['east', 'north', 'west']
Qvalues:  [1.18, -3.36, -8.54]
   Action:  east
Reward:  0, Score 44, Done False
\\ \\ STATE 98 \\  gallery  you are empty handed.  gallery this is an art gallery. most of the paintings have been stolen by vandals with exceptional taste. the vandals left through either the north or west exits. Actions:  ['west', 'north']
Qvalues:  [1.25, 1.13]
   Action:  north
Reward:  0, Score 44, Done False
\\ \\ STATE 99 \\  you have moved into a dark place. it is pitch black. you are likely to be eaten by a grue.  you are empty handed.  it is pitch black. you are likely to be eaten by a grue. Actions:  ['south', 'north']
Qvalues:  [0.99, -3.47]
   Action:  south
Reward:  0, Score 44, Done True
\\
\\ STATE 100 \\ gallery  you are empty handed.  gallery this is an art gallery. most of the paintings have been stolen by vandals with exceptional taste. the vandals left through either the north or west exits.
Actions: ['west', 'north']
Qvalues: [1.21, 1.04]
   Action: reset
Reward13458: 0, Score 0, Done False

\subsection{\texttt{Inhumane}}\label{appendix:Inhuman}

STATE 0 \\ You awaken in your tent, lying on the floor, with a hangover like you wouldn't believe. As you lever yourself up, you notice that there is complete silence instead of the usual morning complaints of the workers and Roboff's usual shouting. This can only mean one thing Roboff must have finally run off on you. Panicked, you look around and see your amulet lying on the floor. At least Roboff had the sense to not trigger off your alarm. Knowing him, he also had the sense to take the emergency radio. How the megakrip are you going to get out of this one? INHUMANE : An Infralogic Massacre. Copyright ( C ) by Infobom Inhumane and Infobom are trademarks of Sadistic Software, Ins. Revision 2 / Serial Number 31415926 / Inform v1405 Library 5 / 8 First time players should type " about ". In Your Tent You are in your tent in the encampment. The tent appears to have been completely cleaned out. It is very hot in here and getting hotter. The tent flap to the south is open. On the ground is the ancient jeweled amulet that you " borrowed " from Mr. Meshugge. You are holding nothing. In Your Tent You are in your tent in the encampment. The tent appears to have been completely cleaned out. It is very hot in here and getting hotter. The tent flap to the south is open. On the ground is the ancient jeweled amulet that you " borrowed " from Mr. Meshugge. Actions:  ['south', 'take amulet']
Qvalues:  [6.04, 4.66]
   Action:  south
Reward:  0, Score 0, Done False
\\ \\ STATE 1 \\ center of camp this is the center of your camp. the entire place seems to be completely deserted. you can see roboff's tent to the south and the storage tent west. east is open desert. you are holding nothing. center of camp this is the center of your camp. the entire place seems to be completely deserted. you can see roboff's tent to the south and the storage tent west. east is open desert. Actions:  ['east', 'north', 'south', 'west']
Qvalues:  [7.01, 5.93, 5.76, 5.3]
   Action:  south
Reward:  0, Score 0, Done False
\\ \\ STATE 2 \\ roboff's tent you are in francisco roboff's tent. the place is bare except for the emergency signal radio and roboff's toilet in one corner. there is a note taped to the tent wall. you are holding nothing. roboff's tent you are in francisco roboff's tent. the place is bare except for the emergency signal radio and roboff's toilet in one corner. there is a note taped to the tent wall. Actions:  ['north', 'take signal off radio']
Qvalues:  [6.3, 4.72]
   Action:  north
Reward:  0, Score 0, Done False
\\ \\ STATE 3 \\ center of campyou are holding nothing. center of camp this is the center of your camp. the entire place seems to be completely deserted. you can see roboff's tent to the south and the storage tent west. east is open desert. Actions:  ['east', 'north', 'south', 'west']
Qvalues:  [7.15, 5.82, 5.72, 5.61]
   Action:  north
Reward:  0, Score 0, Done False
\\ \\ STATE 4 \\ in your tent on the ground is the ancient jeweled amulet that you " borrowed " from mr. meshugge. you are holding nothing. in your tent you are in your tent in the encampment. the tent appears to have been completely cleaned out. it is very hot in here and getting hotter. the tent flap to the south is open. on the ground is the ancient jeweled amulet that you " borrowed " from mr. meshugge. Actions:  ['south', 'take amulet']
Qvalues:  [6.36, 5.16]
   Action:  south
Reward:  0, Score 0, Done False
\\ \\ STATE 5 \\ center of campyou are holding nothing. center of camp this is the center of your camp. the entire place seems to be completely deserted. you can see roboff's tent to the south and the storage tent west. east is open desert. Actions:  ['east', 'north', 'west', 'south']
Qvalues:  [7.14, 5.84, 5.66, 5.66]
   Action:  east
Reward:  0, Score 0, Done False
\\ \\ STATE 6 \\ you venture out onto the sands and rapidly lose sight of camp. in the desert all you can see in any direction is sand and sunlight. you are sweating like a pig. you are holding nothing. in the desert all you can see in any direction is sand and sunlight. you are sweating like a pig. Actions:  ['put all down', 'north', 'west']
Qvalues:  [7.71, 7.29, 6.3]
   Action:  put all down
Reward:  0, Score 0, Done False
\\ \\ STATE 7 \\ it falls and is lost forever in the sands. it is very dark. before you can do anything, something begins gnawing on you. you have died you are floating disembodied in absolute blackness. time passes at a strange, relative rate. after an infinite / infinitesmal time, you sense a presence near you. a voice speaks, resounding in your head as in a hollow gourd... " ah! another of those wimpy delvers come to steal my treasure. how boring. most of it was taken by that other guy who had the sense to avoid the nastier traps... but i am getting pretty tired of guarding what's left. i don't suppose that you have any chance of thinking your way through my traps, of course. but i'll let you try, and i won't even vaporize you when you try to leave. in fact, i'll make you a deal. if you prove to me that you are an incompetent harry sullivan by getting yourself killed by every single one of my deathtraps, after getting as far towards solving it as possible, then i'll give up and let you take the coffin. " that sound good? i thought so. " that hardly counts as being killed by a trap, you know. even you could have gotten farther into that situation. " you don't seem to have really gotten fooled by any traps yet. maybe there is hope for you. " round room this room is at the peak of the pyramid. it is perfectly conical. in the center is a tall stone obelisk that you could probably use to get back up to the fake toilet. there are four stairways leaving the room at the four compass points. there is a shovel lying here. you are holding : a torch round room this room is at the peak of the pyramid. it is perfectly conical. in the center is a tall stone obelisk that you could probably use to get back up to the fake toilet. there are four stairways leaving the room at the four compass points. there is a shovel lying here. Actions:  ['push button', 'north', 'east', 'west', 'south', 'take button off torch', 'take shovel', 'put torch down', 'examine obelisk', 'up']
Qvalues:  [8.67, 8.06, 8.02, 7.76, 7.63, 7.4, 6.97, 6.88, 6.55, 5.09]
   Action:  north
Reward:  0, Score 0, Done False
\\ \\ STATE 8 \\ darkness it is very dark. before you can do anything, something begins gnawing on you. you have died you are back in the infinite blackness. eventually the voice returns, senses you disgustedly, and speaks. " that hardly counts as being killed by a trap, you know. even you could have gotten farther into that situation. " you don't seem to have really gotten fooled by any traps yet. maybe there is hope for you. " round room this room is at the peak of the pyramid. it is perfectly conical. in the center is a tall stone obelisk that you could probably use to get back up to the fake toilet. there are four stairways leaving the room at the four compass points. there is a shovel lying here. you are holding : a torch round room this room is at the peak of the pyramid. it is perfectly conical. in the center is a tall stone obelisk that you could probably use to get back up to the fake toilet. there are four stairways leaving the room at the four compass points. there is a shovel lying here. Actions:  ['push button', 'east', 'north', 'west', 'south', 'take button off torch', 'take shovel', 'put torch down', 'examine obelisk', 'up']
Qvalues:  [8.76, 8.1, 7.99, 7.62, 7.55, 7.38, 6.93, 6.9, 6.49, 5.19]
   Action:  take button off torch
Reward:  0, Score 0, Done False
\\ \\ STATE 9 \\ removed. you are holding : a torch button a torch round room this room is at the peak of the pyramid. it is perfectly conical. in the center is a tall stone obelisk that you could probably use to get back up to the fake toilet. there are four stairways leaving the room at the four compass points. there is a shovel lying here. Actions:  ['push button', 'north', 'west', 'east', 'south', 'take shovel', 'put all down', 'put torch down', 'put button down', 'examine obelisk', 'up']
Qvalues:  [8.26, 8.02, 7.8, 7.78, 7.59, 6.96, 6.76, 6.72, 6.71, 6.49, 5.56]
   Action:  push button
Reward:  0, Score 0, Done False
\\ \\ STATE 10 \\ the torch flickers a few times and then glows steadily. you are holding : a torch button a torch ( providing light ) round room this room is at the peak of the pyramid. it is perfectly conical. in the center is a tall stone obelisk that you could probably use to get back up to the fake toilet. there are four stairways leaving the room at the four compass points. there is a shovel lying here. Actions:  ['east', 'push button', 'north', 'south', 'take shovel', 'put button down', 'examine obelisk', 'west', 'put torch down', 'put all down', 'up']
Qvalues:  [9.97, 8.69, 8.26, 8.14, 7.65, 7.56, 7.41, 7.09, 6.9, 6.68, 5.32]
   Action:  south
Reward:  0, Score 0, Done False
\\ \\ STATE 11 \\ corridor at alcoves the corridor goes north to the stairs and south from here. there are alcoves to the east and west. in the floor between the alcoves is a quartz window. you are holding : a torch button a torch ( providing light ) corridor at alcoves the corridor goes north to the stairs and south from here. there are alcoves to the east and west. in the floor between the alcoves is a quartz window. Actions:  ['south', 'west', 'push button', 'north', 'put button down', 'put all down', 'put torch down', 'east']
Qvalues:  [8.79, 8.53, 8.07, 7.82, 7.62, 7.56, 7.45, 7.41]
   Action:  south
Reward:  0, Score 0, Done False
\\ \\ STATE 12 \\ outside door you are at the end of the hall. to the south is a wooden door with a steel knob on it. the door is closed. you are holding : a torch button a torch ( providing light ) outside door you are at the end of the hall. to the south is a wooden door with a steel knob on it. the door is closed. Actions:  ['open door', 'push button', 'take brass off door', 'put button down', 'north', 'put torch down', 'put all down']
Qvalues:  [9.73, 8.84, 8.64, 8.39, 8.32, 8.16, 8.05]
   Action:  push button
Reward:  0, Score 0, Done False
\\ \\ STATE 13 \\ the torch flickers a few times and then glows steadily. you are holding : a torch button a torch ( providing light ) outside door you are at the end of the hall. to the south is a wooden door with a steel knob on it. the door is closed. Actions:  ['open door', 'push button', 'take brass off door', 'put button down', 'north', 'put all down', 'put torch down']
Qvalues:  [9.83, 8.92, 8.7, 8.67, 8.23, 8.15, 8.09]
   Action:  push button
Reward:  0, Score 0, Done False
\\ \\ STATE 14 \\ the torch flickers a few times and then glows steadily. you are holding : a torch button a torch ( providing light ) outside door you are at the end of the hall. to the south is a wooden door with a steel knob on it. the door is closed. Actions:  ['open door', 'push button', 'put button down', 'take brass off door', 'north', 'put all down', 'put torch down']
Qvalues:  [9.93, 9.04, 8.76, 8.76, 8.28, 8.19, 8.13]
   Action:  open door
Reward:  0, Score 0, Done False
\\ \\ STATE 15 \\ the door slams open. it is apparently on a very strong spring. if you were on the other side, you would have been slammed into the wall. as it is, the steel door knob slams into the inside wall. you are holding : a torch button a torch ( providing light ) outside door you are at the end of the hall. to the south is a wooden door with a steel knob on it. the door is open. Actions:  ['south', 'push button', 'take brass off door', 'put button down', 'close door', 'put all down', 'north', 'put torch down']
Qvalues:  [10.85, 9.79, 9.52, 9.4, 9.14, 8.61, 8.56, 8.52]
   Action:  south
Reward:  0, Score 0, Done False
\\ \\ STATE 16 \\ room with chest this room is empty except for a small chest at one end. the only exit is the door to the north. the door is open. you see something glimmering behind the door. you'll have to close the door to get it. the chest is closed and locked. you are holding : a torch button a torch ( providing light ) room with chest this room is empty except for a small chest at one end. the only exit is the door to the north. the door is open. you see something glimmering behind the door. you'll have to close the door to get it. the chest is closed and locked. Actions:  ['close door', 'push button', 'take brass off door', 'put torch down', 'north', 'put all down', 'put button down']
Qvalues:  [12.47, 11.55, 11.36, 10.31, 10.26, 10.19, 10.05]
   Action:  close door
Reward:  0, Score 0, Done False
\\ \\ STATE 17 \\ after much tugging, you manage to get the door shut. behind it is a brass needle that you might use to unlock the chest. you are holding : a torch button a torch ( providing light ) room with chest this room is empty except for a small chest at one end. the only exit is the door to the north. the door is closed. on the ground is a long brass needle. the chest is closed and locked. Actions:  ['open chest with needle', 'take brass', 'push button', 'put all down', 'put torch down', 'put button down']
Qvalues:  [13.9, 12.14, 12.14, 11.0, 10.83, 10.5]
   Action:  open chest with needle
Reward:  0, Score 0, Done False
\\ \\ STATE 18 \\ ( first taking the brass needle ) there is a click and the chest lid pops open. inside is a priceless dorkmid! as you stare at it, you notice the smell of coal gas leaking out. you are holding : a brass needle a torch button a torch ( providing light ) room with chest this room is empty except for a small chest at one end. the only exit is the door to the north. the door is closed. the chest is open. the chest contains : a priceless dorkmid Actions:  ['open door', 'put needle down', 'put needle in chest', 'push button', 'put button in chest', 'put torch in chest', 'put all in chest', 'put all down', 'put torch down', 'put button down', 'take dorkmid']
Qvalues:  [15.18, 14.95, 14.61, 13.77, 13.53, 13.52, 13.44, 12.72, 12.46, 12.15, 11.09]
   Action:  open door
Reward:  10, Score 10, Done False
\\ \\ STATE 19 \\ the door slams open. you barely avoid the swinging knob. instead, the steel knob hits the wall with a bang. the wall seems to be made of flint. you have died you are back in the infinite blackness. eventually the voice returns, senses you disgustedly, and speaks. " sparks in a room full of natural gas is not really a smart thing to do. " you've only gotten killed by one trap. not bad. " round room this room is at the peak of the pyramid. it is perfectly conical. in the center is a tall stone obelisk that you could probably use to get back up to the fake toilet. there are four stairways leaving the room at the four compass points. there is a shovel lying here. you are holding : a torch button a torch ( providing light ) round room this room is at the peak of the pyramid. it is perfectly conical. in the center is a tall stone obelisk that you could probably use to get back up to the fake toilet. there are four stairways leaving the room at the four compass points. there is a shovel lying here. Actions:  ['east', 'north', 'push button', 'south', 'take shovel', 'west', 'examine obelisk', 'put button down', 'put torch down', 'put all down', 'up']
Qvalues:  [11.75, 8.96, 8.88, 8.11, 7.56, 7.54, 7.53, 7.32, 6.79, 6.74, 6.41]
   Action:  south
Reward:  0, Score 10, Done False
\\ \\ STATE 20 \\ corridor at alcovesyou are holding : a torch button a torch ( providing light ) corridor at alcoves the corridor goes north to the stairs and south from here. there are alcoves to the east and west. in the floor between the alcoves is a quartz window. Actions:  ['south', 'west', 'push button', 'put button down', 'north', 'put all down', 'east', 'put torch down']
Qvalues:  [8.68, 8.19, 7.98, 7.51, 7.43, 7.36, 7.34, 7.22]
   Action:  put button down
Reward:  0, Score 10, Done False
\\ \\ STATE 21 \\ dropped. you are holding : a torch ( providing light ) corridor at alcoves the corridor goes north to the stairs and south from here. there are alcoves to the east and west. in the floor between the alcoves is a quartz window. you can see a torch button here. Actions:  ['south', 'take button', 'west', 'push button', 'put torch down', 'north', 'east']
Qvalues:  [7.64, 7.58, 7.48, 7.33, 6.72, 6.63, 6.35]
   Action:  west
Reward:  0, Score 10, Done False
\\ \\ STATE 22 \\ west alcove this alcove has a stone platform set in the floor. the platform shifts a little but does not move when you step on it. sticking out of the platform is an iron crank. you are holding : a torch ( providing light ) west alcove this alcove has a stone platform set in the floor. the platform shifts a little but does not move when you step on it. sticking out of the platform is an iron crank. Actions:  ['turn crank', 'east', 'put torch down']
Qvalues:  [8.75, 7.73, 6.44]
   Action:  east
Reward:  0, Score 10, Done False
\\ \\ STATE 23 \\ corridor at alcoves you can see a torch button here. you are holding : a torch ( providing light ) corridor at alcoves the corridor goes north to the stairs and south from here. there are alcoves to the east and west. in the floor between the alcoves is a quartz window. you can see a torch button here. Actions:  ['south', 'take button', 'west', 'push button', 'put torch down', 'north', 'east']
Qvalues:  [7.75, 7.69, 7.48, 7.46, 6.76, 6.7, 6.62]
   Action:  take button
Reward:  0, Score 10, Done False
\\ \\ STATE 24 \\ taken. you are holding : a torch button a torch ( providing light ) corridor at alcoves the corridor goes north to the stairs and south from here. there are alcoves to the east and west. in the floor between the alcoves is a quartz window. Actions:  ['south', 'west', 'push button', 'put button down', 'put all down', 'put torch down', 'north', 'east']
Qvalues:  [8.53, 7.97, 7.74, 7.28, 7.28, 7.24, 7.09, 6.74]
   Action:  west
Reward:  0, Score 10, Done False
\\ \\ STATE 25 \\ west alcoveyou are holding : a torch button a torch ( providing light ) west alcove this alcove has a stone platform set in the floor. the platform shifts a little but does not move when you step on it. sticking out of the platform is an iron crank. Actions:  ['turn crank', 'push button', 'east', 'put button down', 'put torch down', 'put all down']
Qvalues:  [9.2, 7.92, 7.65, 7.19, 6.75, 6.7]
   Action:  turn crank
Reward:  0, Score 10, Done False
\\ \\ STATE 26 \\ the platform slowly moves down a shaft below it. in the other alcove the other platform slowly moves up, counterbalancing the one you are on. on platform you are on the crank platform at the bottom of the shaft under the alcove. there is an opening to the east. you are holding : a torch button a torch ( providing light ) on platform you are on the crank platform at the bottom of the shaft under the alcove. there is an opening to the east. Actions:  ['push button', 'turn crank', 'put button down', 'put torch down', 'put all down', 'east']
Qvalues:  [10.23, 10.22, 9.02, 8.93, 8.7, 8.63]
   Action:  push button
Reward:  0, Score 10, Done False
\\ \\ STATE 27 \\ the torch flickers a few times and then glows steadily. you are holding : a torch button a torch ( providing light ) on platform you are on the crank platform at the bottom of the shaft under the alcove. there is an opening to the east. Actions:  ['turn crank', 'push button', 'put button down', 'east', 'put torch down', 'put all down']
Qvalues:  [11.21, 10.29, 9.09, 9.06, 8.8, 8.62]
   Action:  put all down
Reward:  0, Score 10, Done False
\\ \\ STATE 28 \\ torch button : dropped. torch : dropped. you are holding nothing. on platform you are on the crank platform at the bottom of the shaft under the alcove. there is an opening to the east. lying on the ground is a brass torch of the " self igniting " type. the ignite button is on the handle. you can also see a torch button here. Actions:  ['take all', 'take torch', 'take button', 'push button', 'turn crank', 'east']
Qvalues:  [9.66, 8.96, 8.88, 8.8, 8.21, 7.81]
   Action:  take all
Reward:  0, Score 10, Done False
\\ \\ STATE 29 \\ iron crank : a wimp like you will never get that done. torch : taken. torch button : taken. you are holding : a torch button a torch ( providing light ) on platform you are on the crank platform at the bottom of the shaft under the alcove. there is an opening to the east. Actions:  ['turn crank', 'push button', 'put button down', 'put torch down', 'put all down', 'east']
Qvalues:  [11.85, 10.14, 8.88, 8.85, 8.72, 8.24]
   Action:  east
Reward:  0, Score 10, Done False
\\ \\ STATE 30 \\ in tiny room this is a tiny room. the only exit is back west. you can see the quartz window directly above your head. there is a huge diamond lying on the ground! you are holding : a torch button a torch ( providing light ) in tiny room this is a tiny room. the only exit is back west. you can see the quartz window directly above your head. there is a huge diamond lying on the ground! Actions:  ['west', 'take diamond', 'push button', 'put torch down', 'put button down', 'put all down']
Qvalues:  [10.17, 9.35, 8.97, 8.14, 8.13, 8.12]
   Action:  west
Reward:  0, Score 10, Done False
\\ \\ STATE 31 \\ on platformyou are holding : a torch button a torch ( providing light ) on platform you are on the crank platform at the bottom of the shaft under the alcove. there is an opening to the east. Actions:  ['turn crank', 'push button', 'put button down', 'put torch down', 'put all down', 'east']
Qvalues:  [11.12, 10.08, 8.86, 8.74, 8.66, 7.84]
   Action:  turn crank
Reward:  10, Score 20, Done False
\\ \\ STATE 32 \\ the platform moves back up to the alcove level. in the other alcove, you see the other platform moving back down until both of the platforms are even. west alcove this alcove has a stone platform set in the floor. the platform shifts a little but does not move when you step on it. sticking out of the platform is an iron crank. as the other platform moves down from the hole in the ceiling, you see a small robot standing on it. when it gets down to the floor, it looks at you, says " ex trrr mnnn ate ", points an arm at you, and turns you into a small pile of monatomic dust. you have died you are back in the infinite blackness. eventually the voice returns, senses you disgustedly, and speaks. " giving nasties a free ride down to where you are seems a little stupid. " you've been killed by two traps. that isn't really so bad. " round room this room is at the peak of the pyramid. it is perfectly conical. in the center is a tall stone obelisk that you could probably use to get back up to the fake toilet. there are four stairways leaving the room at the four compass points. there is a shovel lying here. you are holding : a torch button a torch ( providing light ) round room this room is at the peak of the pyramid. it is perfectly conical. in the center is a tall stone obelisk that you could probably use to get back up to the fake toilet. there are four stairways leaving the room at the four compass points. there is a shovel lying here. Actions:  ['push button', 'north', 'south', 'take shovel', 'examine obelisk', 'put button down', 'west', 'put torch down', 'put all down', 'east', 'up']
Qvalues:  [8.24, 7.81, 7.57, 7.32, 7.08, 6.95, 6.66, 6.58, 6.47, 6.31, 5.27]
   Action:  west
Reward:  0, Score 20, Done False
\\ \\ STATE 33 \\ t intersection the east / west corridor branches to the south here. you are holding : a torch button a torch ( providing light ) t intersection the east / west corridor branches to the south here. Actions:  ['east', 'push button', 'west', 'put torch down', 'put all down', 'put button down', 'south']
Qvalues:  [7.36, 6.89, 6.85, 6.17, 6.08, 6.05, 5.85]
   Action:  east
Reward:  0, Score 20, Done False
\\ \\ STATE 34 \\ round room there is a shovel lying here. you are holding : a torch button a torch ( providing light ) round room this room is at the peak of the pyramid. it is perfectly conical. in the center is a tall stone obelisk that you could probably use to get back up to the fake toilet. there are four stairways leaving the room at the four compass points. there is a shovel lying here. Actions:  ['push button', 'north', 'south', 'take shovel', 'examine obelisk', 'east', 'put button down', 'west', 'put torch down', 'put all down', 'up']
Qvalues:  [8.32, 7.79, 7.61, 7.21, 7.12, 6.97, 6.93, 6.61, 6.56, 6.4, 5.75]
   Action:  put button down
Reward:  0, Score 20, Done False
\\ \\ STATE 35 \\ dropped. you are holding : a torch ( providing light ) round room this room is at the peak of the pyramid. it is perfectly conical. in the center is a tall stone obelisk that you could probably use to get back up to the fake toilet. there are four stairways leaving the room at the four compass points. there is a shovel lying here. you can also see a torch button here. Actions:  ['take button', 'north', 'push button', 'take all', 'take shovel', 'east', 'south', 'west', 'examine obelisk', 'put torch down', 'up']
Qvalues:  [7.53, 7.32, 7.3, 7.18, 7.07, 7.0, 6.82, 6.78, 6.67, 6.18, 5.21]
   Action:  north
Reward:  0, Score 20, Done False
\\ \\ STATE 36 \\ landing this is a landing on the stairway. you can go down to the north or up to the south. a passage leads east. you are holding : a torch ( providing light ) landing this is a landing on the stairway. you can go down to the north or up to the south. a passage leads east. Actions:  ['north', 'south', 'east', 'put torch down']
Qvalues:  [8.16, 6.7, 6.56, 6.28]
   Action:  east
Reward:  0, Score 20, Done False
\\ \\ STATE 37 \\ t intersection the hallway goes east and west here with a branch north. you are holding : a torch ( providing light ) t intersection the hallway goes east and west here with a branch north. Actions:  ['west', 'north', 'put torch down', 'east']
Qvalues:  [7.2, 6.93, 6.24, 6.01]
   Action:  put torch down
Reward:  0, Score 20, Done False
\\ \\ STATE 38 \\ dropped. you are holding nothing. t intersection the hallway goes east and west here with a branch north. lying on the ground is a brass torch of the " self igniting " type. the ignite button is on the handle. Actions:  ['west', 'north', 'east', 'take torch']
Qvalues:  [6.88, 6.7, 6.47, 6.28]
   Action:  east
Reward:  0, Score 20, Done False
\\ \\ STATE 39 \\ darkness it is very dark. before you can do anything, something begins gnawing on you. you have died you are back in the infinite blackness. eventually the voice returns, senses you disgustedly, and speaks. " that hardly counts as being killed by a trap, you know. even you could have gotten farther into that situation. " you've been killed by two traps. that isn't really so bad. " round room this room is at the peak of the pyramid. it is perfectly conical. in the center is a tall stone obelisk that you could probably use to get back up to the fake toilet. there are four stairways leaving the room at the four compass points. there is a shovel lying here. you can also see a torch button here. you are holding : a torch ( providing light ) round room this room is at the peak of the pyramid. it is perfectly conical. in the center is a tall stone obelisk that you could probably use to get back up to the fake toilet. there are four stairways leaving the room at the four compass points. there is a shovel lying here. you can also see a torch button here. Actions:  ['take button', 'north', 'push button', 'take all', 'take shovel', 'south', 'west', 'examine obelisk', 'put torch down', 'east', 'up']
Qvalues:  [7.77, 7.65, 7.54, 7.45, 7.14, 7.03, 6.69, 6.66, 6.31, 6.06, 5.33]
   Action:  north
Reward:  0, Score 20, Done False
\\ \\ STATE 40 \\ landingyou are holding : a torch ( providing light ) landing this is a landing on the stairway. you can go down to the north or up to the south. a passage leads east. Actions:  ['north', 'south', 'east', 'put torch down']
Qvalues:  [8.24, 6.67, 6.57, 6.31]
   Action:  north
Reward:  0, Score 20, Done False
\\ \\ STATE 41 \\ bottom of shaft you are at the bottom of a ladder that goes up a shaft from here. to the south is the stairway up. on the floor is a small coin bag made of leather. it appears to have something in it. you are holding : a torch ( providing light ) bottom of shaft you are at the bottom of a ladder that goes up a shaft from here. to the south is the stairway up. on the floor is a small coin bag made of leather. it appears to have something in it. Actions:  ['up', 'take coin-bag', 'south', 'put torch down']
Qvalues:  [9.37, 8.64, 7.42, 7.26]
   Action:  take coin-bag
Reward:  0, Score 20, Done False
\\ \\ STATE 42 \\ taken. you are holding : a small leather coin bag a torch ( providing light ) bottom of shaft you are at the bottom of a ladder that goes up a shaft from here. to the south is the stairway up. Actions:  ['up', 'put coin-bag down', 'south', 'put torch down', 'put all down']
Qvalues:  [9.71, 8.47, 7.53, 7.14, 7.05]
   Action:  up
Reward:  0, Score 20, Done False
\\ \\ STATE 43 \\ you climb up the ladder to a hole at the top of the shaft. exercise wheel room this room is laid out like an exercise wheel for rats except that it is 40 ft. across. you can turn the wheel by walking north or south. equally spaced around the rim are twelve coin holders. some have gold coins in them. the coin holder at your feet has a coin in it. the hole is at the bottom of the wheel. below it is the shaft that you entered this room by. you are holding : a small leather coin bag a torch ( providing light ) exercise wheel room this room is laid out like an exercise wheel for rats except that it is 40 ft. across. you can turn the wheel by walking north or south. equally spaced around the rim are twelve coin holders. some have gold coins in them. the coin holder at your feet has a coin in it. the hole is at the bottom of the wheel. below it is the shaft that you entered this room by. Actions:  ['south', 'put coin-bag down', 'put torch down', 'put all down', 'down', 'north']
Qvalues:  [10.62, 8.9, 8.48, 8.29, 8.23, 8.13]
   Action:  south
Reward:  0, Score 20, Done False
\\ \\ STATE 44 \\ the wheel rotates under your feet. you stay at the bottom, but the hole in the rim that you entered by has moved. the coin holder at your feet has a coin in it. the hole is slightly up the north side. you are holding : a small leather coin bag a torch ( providing light ) exercise wheel room this room is laid out like an exercise wheel for rats except that it is 40 ft. across. you can turn the wheel by walking north or south. equally spaced around the rim are twelve coin holders. some have gold coins in them. the coin holder at your feet has a coin in it. the hole is slightly up the north side. Actions:  ['south', 'put coin-bag down', 'north', 'put torch down', 'put all down']
Qvalues:  [11.58, 9.76, 9.28, 9.2, 9.07]
   Action:  put torch down
Reward:  0, Score 20, Done False
\\ \\ STATE 45 \\ dropped. you are holding : a small leather coin bag exercise wheel room this room is laid out like an exercise wheel for rats except that it is 40 ft. across. you can turn the wheel by walking north or south. equally spaced around the rim are twelve coin holders. some have gold coins in them. the coin holder at your feet has a coin in it. the hole is slightly up the north side. lying on the ground is a brass torch of the " self igniting " type. the ignite button is on the handle. Actions:  ['take brass', 'south', 'put coin-bag down', 'north']
Qvalues:  [10.36, 10.14, 9.24, 7.98]
   Action:  north
Reward:  0, Score 20, Done False
\\ \\ STATE 46 \\ the wheel rotates under your feet. you stay at the bottom, but the hole in the rim that you entered by has moved. the coin holder at your feet has a coin in it. the hole is at the bottom of the wheel. below it is the shaft that you entered this room by. you are holding : a small leather coin bag exercise wheel room this room is laid out like an exercise wheel for rats except that it is 40 ft. across. you can turn the wheel by walking north or south. equally spaced around the rim are twelve coin holders. some have gold coins in them. the coin holder at your feet has a coin in it. the hole is at the bottom of the wheel. below it is the shaft that you entered this room by. lying on the ground is a brass torch of the " self igniting " type. the ignite button is on the handle. Actions:  ['south', 'take torch', 'put coin-bag down', 'down', 'north']
Qvalues:  [8.84, 8.82, 8.14, 7.12, 7.07]
   Action:  put coin-bag down
Reward:  0, Score 20, Done False
\\ \\ STATE 47 \\ dropped. you are holding nothing. exercise wheel room this room is laid out like an exercise wheel for rats except that it is 40 ft. across. you can turn the wheel by walking north or south. equally spaced around the rim are twelve coin holders. some have gold coins in them. the coin holder at your feet has a coin in it. the hole is at the bottom of the wheel. below it is the shaft that you entered this room by. on the floor is a small coin bag made of leather. it appears to have something in it. lying on the ground is a brass torch of the " self igniting " type. the ignite button is on the handle. Actions:  ['take all', 'south', 'take torch', 'take coin-bag', 'take in floor', 'north']
Qvalues:  [9.69, 9.44, 9.1, 8.53, 8.18, 7.63]
   Action:  take all
Reward:  0, Score 20, Done False
\\ \\ STATE 48 \\ small leather coin bag : taken. torch : taken. coin holder : you pick up the coin and put it in your bag. you are holding : a torch ( providing light ) a small leather coin bag exercise wheel room this room is laid out like an exercise wheel for rats except that it is 40 ft. across. you can turn the wheel by walking north or south. equally spaced around the rim are twelve coin holders. some have gold coins in them. the coin holder at your feet is empty. the hole is at the bottom of the wheel. below it is the shaft that you entered this room by. Actions:  ['south', 'put coin-bag down', 'put torch down', 'put all down', 'north', 'down']
Qvalues:  [9.59, 8.74, 8.14, 8.11, 7.7, 7.53]
   Action:  north
Reward:  0, Score 20, Done False
\\ \\ STATE 49 \\ the wheel rotates under your feet. you stay at the bottom, but the hole in the rim that you entered by has moved. the coin holder at your feet has a coin in it. the hole is slightly up the south side. you are holding : a torch ( providing light ) a small leather coin bag exercise wheel room this room is laid out like an exercise wheel for rats except that it is 40 ft. across. you can turn the wheel by walking north or south. equally spaced around the rim are twelve coin holders. some have gold coins in them. the coin holder at your feet has a coin in it. the hole is slightly up the south side. Actions:  ['south', 'put coin-bag down', 'put all down', 'put torch down', 'north']
Qvalues:  [8.63, 8.06, 7.47, 7.36, 7.29]
   Action:  put coin-bag down
Reward:  0, Score 20, Done False
\\ \\ STATE 50 \\ dropped. you are holding : a torch ( providing light ) exercise wheel room this room is laid out like an exercise wheel for rats except that it is 40 ft. across. you can turn the wheel by walking north or south. equally spaced around the rim are twelve coin holders. some have gold coins in them. the coin holder at your feet has a coin in it. the hole is slightly up the south side. on the floor is a small coin bag made of leather. it appears to have something in it. Actions:  ['south', 'take coin-bag', 'north', 'put torch down']
Qvalues:  [9.01, 8.69, 8.03, 7.6]
   Action:  south
Reward:  0, Score 20, Done False
\\ \\ STATE 51 \\ the wheel rotates under your feet. you stay at the bottom, but the hole in the rim that you entered by has moved. the coin holder at your feet is empty. the hole is at the bottom of the wheel. below it is the shaft that you entered this room by. you are holding : a torch ( providing light ) exercise wheel room this room is laid out like an exercise wheel for rats except that it is 40 ft. across. you can turn the wheel by walking north or south. equally spaced around the rim are twelve coin holders. some have gold coins in them. the coin holder at your feet is empty. the hole is at the bottom of the wheel. below it is the shaft that you entered this room by. on the floor is a small coin bag made of leather. it appears to have something in it. Actions:  ['south', 'take coin-bag', 'put torch down', 'north', 'take in floor']
Qvalues:  [10.48, 10.08, 8.75, 8.58, 8.15]
   Action:  put torch down
Reward:  0, Score 20, Done False
\\ \\ STATE 52 \\ dropped. you are holding nothing. exercise wheel room this room is laid out like an exercise wheel for rats except that it is 40 ft. across. you can turn the wheel by walking north or south. equally spaced around the rim are twelve coin holders. some have gold coins in them. the coin holder at your feet is empty. the hole is at the bottom of the wheel. below it is the shaft that you entered this room by. lying on the ground is a brass torch of the " self igniting " type. the ignite button is on the handle. on the floor is a small coin bag made of leather. it appears to have something in it. Actions:  ['take all', 'south', 'take torch', 'take coin-bag', 'take in floor', 'north']
Qvalues:  [10.04, 9.65, 9.47, 8.81, 8.42, 8.17]
   Action:  take torch
Reward:  0, Score 20, Done False
\\ \\ STATE 53 \\ taken. you are holding : a torch ( providing light ) exercise wheel room this room is laid out like an exercise wheel for rats except that it is 40 ft. across. you can turn the wheel by walking north or south. equally spaced around the rim are twelve coin holders. some have gold coins in them. the coin holder at your feet is empty. the hole is at the bottom of the wheel. below it is the shaft that you entered this room by. on the floor is a small coin bag made of leather. it appears to have something in it. Actions:  ['south', 'take coin-bag', 'put torch down', 'north', 'take in floor']
Qvalues:  [10.47, 9.96, 8.81, 8.62, 8.6]
   Action:  south
Reward:  0, Score 20, Done False
\\ \\ STATE 54 \\ the wheel rotates under your feet. you stay at the bottom, but the hole in the rim that you entered by has moved. the coin holder at your feet has a coin in it. the hole is slightly up the north side. you are holding : a torch ( providing light ) exercise wheel room this room is laid out like an exercise wheel for rats except that it is 40 ft. across. you can turn the wheel by walking north or south. equally spaced around the rim are twelve coin holders. some have gold coins in them. the coin holder at your feet has a coin in it. the hole is slightly up the north side. on the floor is a small coin bag made of leather. it appears to have something in it. Actions:  ['south', 'take coin-bag', 'put torch down', 'north']
Qvalues:  [11.74, 11.16, 9.81, 9.42]
   Action:  take coin-bag
Reward:  0, Score 20, Done False
\\ \\ STATE 55 \\ taken. you are holding : a small leather coin bag a torch ( providing light ) exercise wheel room this room is laid out like an exercise wheel for rats except that it is 40 ft. across. you can turn the wheel by walking north or south. equally spaced around the rim are twelve coin holders. some have gold coins in them. the coin holder at your feet has a coin in it. the hole is slightly up the north side. Actions:  ['south', 'put coin-bag down', 'north', 'put torch down', 'put all down']
Qvalues:  [12.41, 10.59, 10.06, 9.98, 9.9]
   Action:  south
Reward:  0, Score 20, Done False
\\ \\ STATE 56 \\ the wheel rotates under your feet. you stay at the bottom, but the hole in the rim that you entered by has moved. the coin holder at your feet has a coin in it. the hole is less than half way up the north side. you are holding : a small leather coin bag a torch ( providing light ) exercise wheel room this room is laid out like an exercise wheel for rats except that it is 40 ft. across. you can turn the wheel by walking north or south. equally spaced around the rim are twelve coin holders. some have gold coins in them. the coin holder at your feet has a coin in it. the hole is less than half way up the north side. Actions:  ['south', 'put coin-bag down', 'put torch down', 'north', 'put all down']
Qvalues:  [13.58, 11.65, 10.99, 10.91, 10.85]
   Action:  put all down
Reward:  0, Score 20, Done False
\\ \\ STATE 57 \\ small leather coin bag : dropped. torch : dropped. you are holding nothing. exercise wheel room this room is laid out like an exercise wheel for rats except that it is 40 ft. across. you can turn the wheel by walking north or south. equally spaced around the rim are twelve coin holders. some have gold coins in them. the coin holder at your feet has a coin in it. the hole is less than half way up the north side. lying on the ground is a brass torch of the " self igniting " type. the ignite button is on the handle. on the floor is a small coin bag made of leather. it appears to have something in it. Actions:  ['take all', 'south', 'take torch', 'take coin-bag', 'north']
Qvalues:  [12.65, 12.31, 12.0, 11.39, 9.82]
   Action:  south
Reward:  0, Score 20, Done False
\\ \\ STATE 58 \\ the wheel rotates under your feet. you stay at the bottom, but the hole in the rim that you entered by has moved. the coin holder at your feet has a coin in it. the hole is half way up the north side. you are holding nothing. exercise wheel room this room is laid out like an exercise wheel for rats except that it is 40 ft. across. you can turn the wheel by walking north or south. equally spaced around the rim are twelve coin holders. some have gold coins in them. the coin holder at your feet has a coin in it. the hole is half way up the north side. lying on the ground is a brass torch of the " self igniting " type. the ignite button is on the handle. on the floor is a small coin bag made of leather. it appears to have something in it. Actions:  ['take all', 'south', 'take torch', 'take coin-bag', 'north']
Qvalues:  [14.04, 13.49, 13.35, 12.75, 10.79]
   Action:  take torch
Reward:  0, Score 20, Done False
\\ \\ STATE 59 \\ taken. you are holding : a torch ( providing light ) exercise wheel room this room is laid out like an exercise wheel for rats except that it is 40 ft. across. you can turn the wheel by walking north or south. equally spaced around the rim are twelve coin holders. some have gold coins in them. the coin holder at your feet has a coin in it. the hole is half way up the north side. on the floor is a small coin bag made of leather. it appears to have something in it. Actions:  ['south', 'take coin-bag', 'put torch down', 'north']
Qvalues:  [14.71, 14.12, 12.3, 11.95]
   Action:  south
Reward:  0, Score 20, Done False
\\ \\ STATE 60 \\ the wheel rotates under your feet. you stay at the bottom, but the hole in the rim that you entered by has moved. the coin holder at your feet has a coin in it. the hole is more than half way up the north side. you are holding : a torch ( providing light ) exercise wheel room this room is laid out like an exercise wheel for rats except that it is 40 ft. across. you can turn the wheel by walking north or south. equally spaced around the rim are twelve coin holders. some have gold coins in them. the coin holder at your feet has a coin in it. the hole is more than half way up the north side. on the floor is a small coin bag made of leather. it appears to have something in it. Actions:  ['south', 'take coin-bag', 'put torch down', 'north']
Qvalues:  [15.89, 15.25, 13.22, 13.01]
   Action:  south
Reward:  10, Score 30, Done False
\\ \\ STATE 61 \\ the wheel rotates under your feet. you stay at the bottom, but the hole in the rim that you entered by has moved. the coin holder at your feet has a coin in it. the hole is almost to the top on the north side. behind it is a hole from which water is pouring very fast. the water turns the wheel back to its original position, then pours down the hole. you are taken with it. you have died you are back in the infinite blackness. eventually the voice returns, senses you disgustedly, and speaks. " don't you know how lab rats eventually end up? " three deathtraps have zocked you so far. you're beginning to seem rather dumb. " round room this room is at the peak of the pyramid. it is perfectly conical. in the center is a tall stone obelisk that you could probably use to get back up to the fake toilet. there are four stairways leaving the room at the four compass points. there is a shovel lying here. you can also see a torch button here. you are holding : a torch ( providing light ) round room this room is at the peak of the pyramid. it is perfectly conical. in the center is a tall stone obelisk that you could probably use to get back up to the fake toilet. there are four stairways leaving the room at the four compass points. there is a shovel lying here. you can also see a torch button here. Actions:  ['take button', 'north', 'push button', 'take all', 'south', 'take shovel', 'west', 'examine obelisk', 'put torch down', 'east', 'up']
Qvalues:  [8.61, 8.49, 8.34, 8.29, 7.95, 7.88, 7.43, 7.37, 7.15, 7.11, 5.9]
   Action:  put torch down
Reward:  0, Score 30, Done False
\\ \\ STATE 62 \\ dropped. you are holding nothing. round room this room is at the peak of the pyramid. it is perfectly conical. in the center is a tall stone obelisk that you could probably use to get back up to the fake toilet. there are four stairways leaving the room at the four compass points. lying on the ground is a brass torch of the " self igniting " type. the ignite button is on the handle. there is a shovel lying here. you can also see a torch button here. Actions:  ['west', 'north', 'take all', 'push button', 'south', 'east', 'take button', 'take torch', 'take shovel', 'examine obelisk', 'up']
Qvalues:  [7.85, 7.68, 7.68, 7.68, 7.66, 7.62, 7.51, 7.44, 7.13, 6.9, 6.1]
   Action:  south
Reward:  0, Score 30, Done False
\\ \\ STATE 63 \\ darkness it is very dark. before you can do anything, something begins gnawing on you. you have died you are back in the infinite blackness. eventually the voice returns, senses you disgustedly, and speaks. " that hardly counts as being killed by a trap, you know. even you could have gotten farther into that situation. " three deathtraps have zocked you so far. you're beginning to seem rather dumb. " round room this room is at the peak of the pyramid. it is perfectly conical. in the center is a tall stone obelisk that you could probably use to get back up to the fake toilet. there are four stairways leaving the room at the four compass points. there is a shovel lying here. you can also see a torch button here. you are holding : a torch ( providing light ) round room this room is at the peak of the pyramid. it is perfectly conical. in the center is a tall stone obelisk that you could probably use to get back up to the fake toilet. there are four stairways leaving the room at the four compass points. there is a shovel lying here. you can also see a torch button here. Actions:  ['take button', 'north', 'push button', 'take all', 'south', 'take shovel', 'west', 'examine obelisk', 'put torch down', 'east', 'up']
Qvalues:  [8.62, 8.38, 8.34, 8.28, 7.94, 7.89, 7.54, 7.35, 7.17, 6.62, 5.78]
   Action:  take all
Reward:  0, Score 30, Done False
\\ \\ STATE 64 \\ torch button : taken. shovel : taken. stone obelisk : a wimp like you will never get that done. you are holding : a shovel a torch button a torch ( providing light ) round room this room is at the peak of the pyramid. it is perfectly conical. in the center is a tall stone obelisk that you could probably use to get back up to the fake toilet. there are four stairways leaving the room at the four compass points. Actions:  ['push button', 'north', 'south', 'put shovel down', 'east', 'put button down', 'examine obelisk', 'put torch down', 'west', 'put all down', 'up']
Qvalues:  [8.96, 8.86, 8.77, 8.23, 8.21, 7.93, 7.84, 7.54, 7.53, 7.5, 5.75]
   Action:  push button
Reward:  0, Score 30, Done False
\\ \\ STATE 65 \\ the torch flickers a few times and then glows steadily. you are holding : a shovel a torch button a torch ( providing light ) round room this room is at the peak of the pyramid. it is perfectly conical. in the center is a tall stone obelisk that you could probably use to get back up to the fake toilet. there are four stairways leaving the room at the four compass points. Actions:  ['east', 'push button', 'north', 'south', 'put shovel down', 'put button down', 'examine obelisk', 'west', 'put torch down', 'put all down', 'up']
Qvalues:  [10.04, 9.23, 9.06, 9.0, 8.39, 8.16, 8.09, 7.59, 7.55, 7.38, 6.29]
   Action:  east
Reward:  0, Score 30, Done False
\\ \\ STATE 66 \\ t intersection to the hall goes east from here. to the west is the stairway to the round room. a narrow passage leads south. you are holding : a shovel a torch button a torch ( providing light ) t intersection to the hall goes east from here. to the west is the stairway to the round room. a narrow passage leads south. Actions:  ['east', 'west', 'push button', 'south', 'put shovel down', 'put torch down', 'put button down', 'put all down']
Qvalues:  [13.19, 8.57, 8.34, 7.48, 7.11, 6.89, 6.88, 6.69]
   Action:  east
Reward:  0, Score 30, Done False
\\ \\ STATE 67 \\ corridor at doorway the hall goes east and west from here and there is a doorway to the north. you are holding : a shovel a torch button a torch ( providing light ) corridor at doorway the hall goes east and west from here and there is a doorway to the north. Actions:  ['north', 'east', 'west', 'push button', 'put shovel down', 'put torch down', 'put button down', 'put all down']
Qvalues:  [15.16, 11.38, 8.99, 8.52, 7.44, 7.26, 7.01, 6.99]
   Action:  north
Reward:  0, Score 30, Done False
\\ \\ STATE 68 \\ room full of gadgets this is the south end of a long room. the walls and floor and ceiling are covered with thousands of metal gadgets that you might want to examine more carefully. there is a path free of gadgets to the north end of the room. you are holding : a shovel a torch button a torch ( providing light ) room full of gadgets this is the south end of a long room. the walls and floor and ceiling are covered with thousands of metal gadgets that you might want to examine more carefully. there is a path free of gadgets to the north end of the room. Actions:  ['north', 'push button', 'south', 'put shovel down', 'put torch down', 'put button down', 'put all down']
Qvalues:  [16.71, 10.81, 9.31, 9.13, 8.81, 8.61, 8.58]
   Action:  north
Reward:  0, Score 30, Done False
\\ \\ STATE 69 \\ north gadget room the room stretches away back south. the metal gadgets are clustered even more densely here. sitting on the floor in a clear space is a cut crystal bowl. the bowl contains : a large perfect ruby a hundreds of quartz chipsyou are holding : a shovel a torch button a torch ( providing light ) north gadget room the room stretches away back south. the metal gadgets are clustered even more densely here. sitting on the floor in a clear space is a cut crystal bowl. the bowl contains : a large perfect ruby a hundreds of quartz chips Actions:  ['take chips', 'push button', 'put torch down', 'put all down', 'put shovel down', 'put button down', 'south']
Qvalues:  [13.84, 9.5, 8.06, 7.99, 7.98, 7.77, 7.26]
   Action:  take chips
Reward:  10, Score 40, Done False
\\ \\ STATE 70 \\ as you do, several of the quartz chips spill onto the floor. some hit metal gadgets, triggering them. darts fly. some darts hit more gadgets. more darts fly. after a few seconds of this, you have been punctured several million times. you have died you are back in the infinite blackness. eventually the voice returns, senses you disgustedly, and speaks. " when standing inside a mass of uranium, it is not usually a good idea to drop a neutron. " you've stumbled into four traps. you're not looking very good. " round room this room is at the peak of the pyramid. it is perfectly conical. in the center is a tall stone obelisk that you could probably use to get back up to the fake toilet. there are four stairways leaving the room at the four compass points. there is a shovel lying here. you are holding : a torch button a torch ( providing light ) round room this room is at the peak of the pyramid. it is perfectly conical. in the center is a tall stone obelisk that you could probably use to get back up to the fake toilet. there are four stairways leaving the room at the four compass points. there is a shovel lying here. Actions:  ['push button', 'south', 'north', 'take shovel', 'examine obelisk', 'put button down', 'put torch down', 'west', 'put all down', 'east', 'up']
Qvalues:  [9.13, 8.58, 8.54, 8.07, 7.74, 7.73, 7.45, 7.42, 7.36, 6.8, 5.76]
   Action:  push button
Reward:  0, Score 40, Done False
\\ \\ STATE 71 \\ the torch flickers a few times and then glows steadily. you are holding : a torch button a torch ( providing light ) round room this room is at the peak of the pyramid. it is perfectly conical. in the center is a tall stone obelisk that you could probably use to get back up to the fake toilet. there are four stairways leaving the room at the four compass points. there is a shovel lying here. Actions:  ['east', 'push button', 'north', 'south', 'take shovel', 'put button down', 'examine obelisk', 'west', 'put torch down', 'put all down', 'up']
Qvalues:  [10.62, 9.35, 8.78, 8.78, 8.0, 7.93, 7.87, 7.55, 7.35, 7.17, 6.39]
   Action:  push button
Reward:  0, Score 40, Done False
\\ \\ STATE 72 \\ the torch flickers a few times and then glows steadily. you are holding : a torch button a torch ( providing light ) round room this room is at the peak of the pyramid. it is perfectly conical. in the center is a tall stone obelisk that you could probably use to get back up to the fake toilet. there are four stairways leaving the room at the four compass points. there is a shovel lying here. Actions:  ['east', 'push button', 'south', 'north', 'put button down', 'take shovel', 'examine obelisk', 'west', 'put torch down', 'put all down', 'up']
Qvalues:  [10.63, 9.25, 8.68, 8.54, 7.96, 7.96, 7.71, 7.57, 7.32, 7.15, 6.03]
   Action:  east
Reward:  0, Score 40, Done False
\\ \\ STATE 73 \\ t intersectionyou are holding : a torch button a torch ( providing light ) t intersection to the hall goes east from here. to the west is the stairway to the round room. a narrow passage leads south. Actions:  ['west', 'push button', 'south', 'put torch down', 'put all down', 'put button down', 'east']
Qvalues:  [8.01, 7.65, 7.22, 7.01, 6.94, 6.93, 6.46]
   Action:  east
Reward:  0, Score 40, Done False
\\ \\ STATE 74 \\ corridor at doorwayyou are holding : a torch button a torch ( providing light ) corridor at doorway the hall goes east and west from here and there is a doorway to the north. Actions:  ['west', 'push button', 'north', 'put torch down', 'put all down', 'put button down', 'east']
Qvalues:  [7.61, 7.48, 7.24, 6.89, 6.78, 6.58, 5.01]
   Action:  west
Reward:  0, Score 40, Done False
\\ \\ STATE 75 \\ t intersectionyou are holding : a torch button a torch ( providing light ) t intersection to the hall goes east from here. to the west is the stairway to the round room. a narrow passage leads south. Actions:  ['west', 'push button', 'south', 'put torch down', 'put all down', 'put button down', 'east']
Qvalues:  [7.94, 7.53, 7.02, 6.93, 6.89, 6.83, 6.15]
   Action:  push button
Reward:  0, Score 40, Done False
\\ \\ STATE 76 \\ the torch flickers a few times and then glows steadily. you are holding : a torch button a torch ( providing light ) t intersection to the hall goes east from here. to the west is the stairway to the round room. a narrow passage leads south. Actions:  ['west', 'push button', 'east', 'south', 'put button down', 'put torch down', 'put all down']
Qvalues:  [8.16, 7.75, 7.7, 7.34, 7.01, 6.85, 6.73]
   Action:  west
Reward:  0, Score 40, Done False
\\ \\ STATE 77 \\ round room there is a shovel lying here. you are holding : a torch button a torch ( providing light ) round room this room is at the peak of the pyramid. it is perfectly conical. in the center is a tall stone obelisk that you could probably use to get back up to the fake toilet. there are four stairways leaving the room at the four compass points. there is a shovel lying here. Actions:  ['push button', 'south', 'north', 'east', 'put button down', 'take shovel', 'examine obelisk', 'put torch down', 'west', 'put all down', 'up']
Qvalues:  [9.42, 8.58, 8.33, 7.99, 7.82, 7.74, 7.61, 7.2, 7.18, 7.05, 5.89]
   Action:  west
Reward:  0, Score 40, Done False
\\ \\ STATE 78 \\ t intersectionyou are holding : a torch button a torch ( providing light ) t intersection the east / west corridor branches to the south here. Actions:  ['east', 'push button', 'west', 'put torch down', 'put all down', 'put button down', 'south']
Qvalues:  [7.66, 7.53, 7.46, 6.63, 6.6, 6.53, 6.31]
   Action:  put all down
Reward:  0, Score 40, Done False
\\ \\ STATE 79 \\ torch button : dropped. torch : dropped. you are holding nothing. t intersection the east / west corridor branches to the south here. lying on the ground is a brass torch of the " self igniting " type. the ignite button is on the handle. you can also see a torch button here. Actions:  ['west', 'east', 'push button', 'south', 'take all', 'take button', 'take torch']
Qvalues:  [7.47, 7.33, 7.29, 7.08, 7.01, 6.79, 6.66]
   Action:  take all
Reward:  0, Score 40, Done False
\\ \\ STATE 80 \\ torch : taken. torch button : taken. you are holding : a torch button a torch ( providing light ) t intersection the east / west corridor branches to the south here. Actions:  ['east', 'west', 'push button', 'put all down', 'put torch down', 'put button down', 'south']
Qvalues:  [8.2, 7.56, 7.44, 6.46, 6.42, 6.34, 6.2]
   Action:  east
Reward:  0, Score 40, Done False
\\ \\ STATE 81 \\ round room there is a shovel lying here. you are holding : a torch button a torch ( providing light ) round room this room is at the peak of the pyramid. it is perfectly conical. in the center is a tall stone obelisk that you could probably use to get back up to the fake toilet. there are four stairways leaving the room at the four compass points. there is a shovel lying here. Actions:  ['push button', 'south', 'north', 'east', 'put button down', 'take shovel', 'examine obelisk', 'put torch down', 'west', 'put all down', 'up']
Qvalues:  [9.17, 8.31, 8.05, 8.02, 7.6, 7.51, 7.33, 6.95, 6.94, 6.82, 5.48]
   Action:  push button
Reward:  0, Score 40, Done False
\\ \\ STATE 82 \\ the torch flickers a few times and then glows steadily. you are holding : a torch button a torch ( providing light ) round room this room is at the peak of the pyramid. it is perfectly conical. in the center is a tall stone obelisk that you could probably use to get back up to the fake toilet. there are four stairways leaving the room at the four compass points. there is a shovel lying here. Actions:  ['east', 'push button', 'south', 'north', 'put button down', 'take shovel', 'examine obelisk', 'west', 'put torch down', 'put all down', 'up']
Qvalues:  [10.47, 9.04, 8.33, 8.11, 7.78, 7.62, 7.35, 7.27, 7.01, 6.81, 5.67]
   Action:  east
Reward:  0, Score 40, Done False
\\ \\ STATE 83 \\ t intersectionyou are holding : a torch button a torch ( providing light ) t intersection to the hall goes east from here. to the west is the stairway to the round room. a narrow passage leads south. Actions:  ['west', 'push button', 'south', 'put torch down', 'put all down', 'put button down', 'east']
Qvalues:  [7.63, 7.29, 6.78, 6.6, 6.56, 6.53, 6.3]
   Action:  push button
Reward:  0, Score 40, Done False
\\ \\ STATE 84 \\ the torch flickers a few times and then glows steadily. you are holding : a torch button a torch ( providing light ) t intersection to the hall goes east from here. to the west is the stairway to the round room. a narrow passage leads south. Actions:  ['west', 'push button', 'east', 'south', 'put button down', 'put torch down', 'put all down']
Qvalues:  [7.92, 7.45, 7.27, 7.0, 6.68, 6.57, 6.44]
   Action:  push button
Reward:  0, Score 40, Done False
\\ \\ STATE 85 \\ the torch flickers a few times and then glows steadily. you are holding : a torch button a torch ( providing light ) t intersection to the hall goes east from here. to the west is the stairway to the round room. a narrow passage leads south. Actions:  ['west', 'push button', 'east', 'south', 'put button down', 'put torch down', 'put all down']
Qvalues:  [7.89, 7.4, 7.27, 6.96, 6.64, 6.55, 6.41]
   Action:  put all down
Reward:  0, Score 40, Done False
\\ \\ STATE 86 \\ torch button : dropped. torch : dropped. you are holding nothing. t intersection to the hall goes east from here. to the west is the stairway to the round room. a narrow passage leads south. lying on the ground is a brass torch of the " self igniting " type. the ignite button is on the handle. you can also see a torch button here. Actions:  ['east', 'west', 'push button', 'take all', 'south', 'take button', 'take torch']
Qvalues:  [7.4, 7.13, 7.1, 7.08, 7.05, 6.86, 6.75]
   Action:  take all
Reward:  0, Score 40, Done False
\\ \\ STATE 87 \\ torch : taken. torch button : taken. you are holding : a torch button a torch ( providing light ) t intersection to the hall goes east from here. to the west is the stairway to the round room. a narrow passage leads south. Actions:  ['west', 'push button', 'south', 'east', 'put button down', 'put torch down', 'put all down']
Qvalues:  [7.89, 7.18, 6.81, 6.68, 6.31, 6.29, 6.27]
   Action:  west
Reward:  0, Score 40, Done False
\\ \\ STATE 88 \\ round room there is a shovel lying here. you are holding : a torch button a torch ( providing light ) round room this room is at the peak of the pyramid. it is perfectly conical. in the center is a tall stone obelisk that you could probably use to get back up to the fake toilet. there are four stairways leaving the room at the four compass points. there is a shovel lying here. Actions:  ['push button', 'south', 'north', 'east', 'take shovel', 'put button down', 'examine obelisk', 'west', 'put torch down', 'put all down', 'up']
Qvalues:  [8.74, 7.99, 7.93, 7.42, 7.38, 7.27, 7.14, 6.84, 6.76, 6.65, 5.54]
   Action:  push button
Reward:  0, Score 40, Done False
\\ \\ STATE 89 \\ the torch flickers a few times and then glows steadily. you are holding : a torch button a torch ( providing light ) round room this room is at the peak of the pyramid. it is perfectly conical. in the center is a tall stone obelisk that you could probably use to get back up to the fake toilet. there are four stairways leaving the room at the four compass points. there is a shovel lying here. Actions:  ['east', 'push button', 'south', 'north', 'take shovel', 'put button down', 'examine obelisk', 'west', 'put torch down', 'put all down', 'up']
Qvalues:  [10.16, 8.88, 8.24, 8.2, 7.52, 7.49, 7.35, 7.07, 6.85, 6.63, 5.99]
   Action:  north
Reward:  0, Score 40, Done False
\\ \\ STATE 90 \\ landingyou are holding : a torch button a torch ( providing light ) landing this is a landing on the stairway. you can go down to the north or up to the south. a passage leads east. Actions:  ['north', 'push button', 'south', 'put button down', 'put torch down', 'east', 'put all down']
Qvalues:  [8.7, 8.17, 7.45, 7.15, 6.9, 6.81, 6.81]
   Action:  put all down
Reward:  0, Score 40, Done False
\\ \\ STATE 91 \\ torch button : dropped. torch : dropped. you are holding nothing. landing this is a landing on the stairway. you can go down to the north or up to the south. a passage leads east. lying on the ground is a brass torch of the " self igniting " type. the ignite button is on the handle. you can also see a torch button here. Actions:  ['take all', 'take torch', 'north', 'take button', 'east', 'push button', 'south']
Qvalues:  [7.96, 7.45, 7.44, 7.34, 7.3, 7.3, 7.09]
   Action:  north
Reward:  0, Score 40, Done False
\\ \\ STATE 92 \\ darkness it is very dark. before you can do anything, something begins gnawing on you. you have died you are back in the infinite blackness. eventually the voice returns, senses you disgustedly, and speaks. " that hardly counts as being killed by a trap, you know. even you could have gotten farther into that situation. " you've stumbled into four traps. you're not looking very good. " round room this room is at the peak of the pyramid. it is perfectly conical. in the center is a tall stone obelisk that you could probably use to get back up to the fake toilet. there are four stairways leaving the room at the four compass points. there is a shovel lying here. you are holding : a torch ( providing light ) round room this room is at the peak of the pyramid. it is perfectly conical. in the center is a tall stone obelisk that you could probably use to get back up to the fake toilet. there are four stairways leaving the room at the four compass points. there is a shovel lying here. Actions:  ['south', 'north', 'take shovel', 'examine obelisk', 'put torch down', 'west', 'east', 'up']
Qvalues:  [7.93, 7.62, 7.36, 7.07, 6.82, 6.67, 6.48, 4.8]
   Action:  put torch down
Reward:  0, Score 40, Done False
\\ \\ STATE 93 \\ dropped. you are holding nothing. round room this room is at the peak of the pyramid. it is perfectly conical. in the center is a tall stone obelisk that you could probably use to get back up to the fake toilet. there are four stairways leaving the room at the four compass points. lying on the ground is a brass torch of the " self igniting " type. the ignite button is on the handle. there is a shovel lying here. Actions:  ['west', 'south', 'east', 'take all', 'north', 'take brass', 'take shovel', 'examine obelisk', 'up']
Qvalues:  [7.63, 7.56, 7.54, 7.46, 7.46, 7.34, 6.94, 6.73, 6.22]
   Action:  examine obelisk
Reward:  0, Score 40, Done False
\\ \\ STATE 94 \\ the obelisk is a monument to the great rahn. the inscription starts : " for rahn hath given us the postulates and the theorems ; we shalt thank rahn. for rahn hath said'between any two points there shalt be but one line'; and he found it good. for rahn hath said'if two planes intersecteth, they shalt form but one straight line'; and he found it good. for rahn hath said... " it goes on like this for a long time. when you manage to break the trance, your mind is numbed and full of rahn's strange commandments. round room this room is a conic surface with exits at even multiples of 90 degrees. in the center is a regular square pyramid with a narrow vertex angle. lying on the ground is a brass torch of the " self igniting " type. the ignite button is on the handle. there is a shovel lying here. you are holding nothing. round room this room is a conic surface with exits at even multiples of 90 degrees. in the center is a regular square pyramid with a narrow vertex angle. lying on the ground is a brass torch of the " self igniting " type. the ignite button is on the handle. there is a shovel lying here. Actions:  ['south', 'take all', 'east', 'west', 'take brass', 'north', 'take shovel', 'examine obelisk', 'up']
Qvalues:  [7.34, 7.27, 7.13, 7.1, 7.08, 7.05, 6.59, 6.37, 5.73]
   Action:  south
Reward:  0, Score 40, Done False
\\ \\ STATE 95 \\ darkness it is very dark. before you can do anything, something begins gnawing on you. you have died you are back in the infinite blackness. eventually the voice returns, senses you disgustedly, and speaks. " that hardly counts as being killed by a trap, you know. even you could have gotten farther into that situation. " you've stumbled into four traps. you're not looking very good. " round room this room is a conic surface with exits at even multiples of 90 degrees. in the center is a regular square pyramid with a narrow vertex angle. there is a shovel lying here. you are holding : a torch ( providing light ) round room this room is a conic surface with exits at even multiples of 90 degrees. in the center is a regular square pyramid with a narrow vertex angle. there is a shovel lying here. Actions:  ['south', 'north', 'take shovel', 'examine obelisk', 'put torch down', 'west', 'east', 'up']
Qvalues:  [8.0, 7.7, 7.34, 7.26, 6.71, 6.65, 6.15, 4.76]
   Action:  examine obelisk
Reward:  0, Score 40, Done False
\\ \\ STATE 97 \\ the obelisk is a monument to the great rahn. the inscription starts : " for rahn hath given us the postulates and the theorems ; we shalt thank rahn. for rahn hath said'between any two points there shalt be but one line'; and he found it good. for rahn hath said'if two planes intersecteth, they shalt form but one straight line'; and he found it good. for rahn hath said... " it goes on like this for a long time. when you manage to break the trance, your mind is numbed and full of rahn's strange commandments. round room this room is a conic surface with exits at even multiples of 90 degrees. in the center is a regular square pyramid with a narrow vertex angle. there is a shovel lying here. you are holding : a torch ( providing light ) round room this room is a conic surface with exits at even multiples of 90 degrees. in the center is a regular square pyramid with a narrow vertex angle. there is a shovel lying here. Actions:  ['south', 'north', 'take shovel', 'examine obelisk', 'put torch down', 'west', 'east', 'up']
Qvalues:  [7.91, 7.7, 7.27, 7.16, 6.6, 6.57, 6.45, 5.07]
   Action:  put torch down
Reward:  0, Score 40, Done False
\\ \\ STATE 98 \\ dropped. you are holding nothing. round room this room is a conic surface with exits at even multiples of 90 degrees. in the center is a regular square pyramid with a narrow vertex angle. lying on the ground is a brass torch of the " self igniting " type. the ignite button is on the handle. there is a shovel lying here. Actions:  ['take all', 'south', 'west', 'take brass', 'east', 'north', 'take shovel', 'examine obelisk', 'up']
Qvalues:  [7.4, 7.34, 7.26, 7.24, 7.21, 7.19, 6.74, 6.51, 6.06]
   Action:  south
Reward:  0, Score 40, Done False
\\ \\ STATE 99 \\ darkness it is very dark. before you can do anything, something begins gnawing on you. you have died you are back in the infinite blackness. eventually the voice returns, senses you disgustedly, and speaks. " that hardly counts as being killed by a trap, you know. even you could have gotten farther into that situation. " you've stumbled into four traps. you're not looking very good. " round room this room is a conic surface with exits at even multiples of 90 degrees. in the center is a regular square pyramid with a narrow vertex angle. there is a shovel lying here. you are holding : a torch ( providing light ) round room this room is a conic surface with exits at even multiples of 90 degrees. in the center is a regular square pyramid with a narrow vertex angle. there is a shovel lying here. Actions:  ['south', 'north', 'take shovel', 'examine obelisk', 'put torch down', 'west', 'east', 'up']
Qvalues:  [8.02, 7.67, 7.34, 7.23, 6.75, 6.66, 6.23, 4.74]
   Action:  west
Reward:  0, Score 40, Done False
\\ \\ STATE 100 \\ t intersectionyou are holding : a torch ( providing light ) t intersection the east / west corridor branches to the south here. Actions:  ['west', 'east', 'put torch down', 'south']
Qvalues:  [7.21, 6.7, 6.47, 5.87]
   Action:  east
Reward:  0, Score 40, Done True
\\ \\ STATE 101 \\ round room there is a shovel lying here. you are holding : a torch ( providing light ) round room this room is a conic surface with exits at even multiples of 90 degrees. in the center is a regular square pyramid with a narrow vertex angle. there is a shovel lying here. Actions:  ['south', 'north', 'take shovel', 'examine obelisk', 'put torch down', 'west', 'east', 'up']
Qvalues:  [7.82, 7.45, 7.12, 6.99, 6.59, 6.5, 6.18, 4.91]
   Action:  reset
Reward:  0, Score 0, Done False